\documentclass[preprint,12pt,authoryear]{elsarticle}

\usepackage[a4paper, margin=2.5cm]{geometry}

\usepackage{graphicx} 
\usepackage{array}
\usepackage{booktabs}
\usepackage{hyperref}
\usepackage{multirow}
\usepackage{subcaption}
\usepackage{float}
\usepackage{placeins}

\begin{document}

\begin{frontmatter}
\title{Math anxiety and associative knowledge structure are entwined in psychology students but not in Large Language Models like GPT-3.5 and GPT-4o}

\author[dipsco]{Luciana Ciringione\texorpdfstring{\corref{equal}}{}}

\author[cognosco]{Emma Franchino\texorpdfstring{\corref{equal}}{}}

\author[dipsco]{Simone Reigl}

\author[dipsco]{Isaia D'Onofrio}

\author[dipsco]{Anna Serbati}

\author[monaco1,monaco2]{Oleksandra Poquet}

\author[australia]{Florence Gabriel}

\author[cognosco]{Massimo Stella\texorpdfstring{\corref{cor2}}{}}

\affiliation[dipsco]{
            organization={Department of Psychology and Cognitive Science, University of Trento},
            addressline={Corso Bettini, 31},
            city={Rovereto},
            postcode={38068},
            state={TN},
            country={Italy}}

\affiliation[cognosco]{
            organization={CogNosco Lab, Department of Psychology and Cognitive Science, University of Trento},
            addressline={Corso Bettini, 31}, 
            city={Rovereto},
            postcode={38068}, 
            state={TN},
            country={Italy}}
            
\affiliation[monaco1]{
            organization={Technical University of Munich},
            addressline={Arcisstraße, 21}, 
            city={München},
            postcode={80333}, 
            state={BY},
            country={Germany}}

\affiliation[monaco2]{
            organization={Munich Data Science Institute},
            addressline={Walther-Von-Dyck Str. 1},
            city={Garching bei München},
            postcode={85748},
            state={BY},
            country={Germany}}
            
\affiliation[australia]{
            organization={Centre for Change and Complexity in Learning, University of South Australia},
            addressline={City West Campus, GPO Box 2471},
            city={Adelaide},
            postcode={5001},
            state={SA},
            country={Australia}}

\cortext[equal]{These authors equally contributed.}
\cortext[corr]{Correspondence e-mail: massimo.stella-1@unitn.it}

\begin{abstract}
    Math anxiety poses significant challenges for university psychology students, affecting their career choices and overall well-being. This study employs a framework based on behavioural forma mentis networks (i.e. cognitive models that map how individuals structure their associative knowledge and emotional perceptions of concepts) to explore individual and group differences in the perception and association of concepts related to math and anxiety. We conducted 4 experiments involving psychology undergraduates from 2 samples ($n_1 = 70$, $n_2 = 57$) compared against GPT-simulated students (GPT-3.5: $n_3 = 300$; GPT-4o: $n_4 = 300$). Experiments 1, 2, and 3 employ individual-level network features to predict psychometric scores for math anxiety and its facets (observational, social and evaluational) from the Math Anxiety Scale. Experiment 4 focuses on group-level perceptions extracted from human students, GPT-3.5 and GPT-4o’s networks. Results indicate that, in students, positive valence ratings and higher network degree for "anxiety", together with negative ratings for "math", can predict higher total and evaluative math anxiety. In contrast, these models do not work on GPT-based data because of differences in simulated networks and psychometric scores compared to humans. These results were also reconciled with differences found in the ways that high/low subgroups of simulated and real students framed semantically and emotionally STEM concepts. High math-anxiety students collectively framed "anxiety" in an emotionally polarising way, absent in the negative perception of low math-anxiety students. "Science" was rated positively, but contrasted against the negative perception of "math". These findings underscore the importance of understanding concept perception and associations in managing students’ math anxiety. 
\end{abstract}


\end{frontmatter}

\section{Introduction} \label{Introduction}
Math anxiety is characterised by feelings of apprehension and heightened physiological responses when individuals encounter math-related situations \citep{Hembree1990, Richardson1972}. This includes exposure to evaluative contexts involving mathematical ability (e.g. taking a math exam), the need to handle numbers in everyday scenarios (e.g. splitting bills), and the passive observation of math-related activities (e.g. watching someone solve a problem \cite{Hunt2011}). Although suffering from math anxiety may not meet the diagnostic criteria of the Diagnostic and Statistical Manual of Mental Disorders \citep{apa2022dsm} for an anxiety disorder, evidence shows that individuals who suffer from math anxiety experience a significant impairment in their well-being, as well as in their academic success \citep{ashcraft2002math, Wu2014, Chang2016, Gabriel2020, caviola2022math}. Furthermore, math anxiety should be distinguished from anxiety related to other STEM (Science, Technological, Engineering, and Mathematics) subjects \citep{Paechter2017}, or from general test anxiety \citep{Zeidner2007, cassady2002cognitive}.

Math anxiety manifests in both cognitive and emotional dimensions: individuals experience impairment in working memory \citep{beilock2007stereotype, Macher2012}, and report feelings of considerable tension and worry. Reducing the maximum load of working memory means that math anxiety can drastically reduce the ability for individuals to load and manipulate symbols, competencies and knowledge, negatively impacting their math performance \citep{stella2022network}. Besides this, as a consequence of math anxiety, individuals adopt safety-seeking behaviours, such as actions that could help them avoid feared stimuli or outcomes \citep{Cuming2009, Gangemi2012}. Since its early days \citep{mowrer1960learning}, psychology has defined avoidance as a behavioural response to external stimuli that produce fear and anxiety, aimed at reducing the experience of fear through negative reinforcement, which can be intrinsically rewarding. Internal stimuli, such as thoughts and emotions, contribute to a prolonged anxious response that prevents the extinction of the anxiety and reinforces the emotion itself \citep{Hayes1996}. Although this principle originates from research on general anxiety disorders, it is also relevant to math anxiety. 

In the context of math anxiety, avoidance behaviour manifests as a reluctance to engage with math tasks, which prevents cognitive reprocessing and deepening of mathematical understanding. By avoiding math-related tasks, students fail to strengthen, extend, and nuance their mathematical concepts through continuous processing, leading to a cycle of avoidance that hinders their mathematical development \citep{Hembree1990}.

However, when anxiety leads an individual to experiential avoidance, preventing constant exposure to anxiety-inducing information, mental representations remain static, hindering the development of cognitive complexity (i.e. the creation of fewer associations with the target of avoidance) and strengthening the negative emotional aspect (i.e. to perceiving the target of avoidance as more negative than it is). This triggers various cognitive biases, common in both individuals who suffer from anxiety and those who do not \citep{vandenhout2014behavior}. For example, the "behaviour as information" bias leads individuals to believe that "If I avoid, then there is a danger", prompting safety-seeking behaviours that hinder further exposure to the feared stimulus, and create a cascading effect \citep{Gangemi2012}. Another type of bias is emotional reasoning, through which individuals tend to use their emotions as significant information about external events, even when the emotion is not generated by the situation being evaluated \citep{arntz1995if, Gangemi2012}. Math anxiety can also be transmitted and reinforce gender stereotypes, as demonstrated by the work of \citet{beilock2010female}, which highlighted that female teachers’ math anxiety can reinforce gender stereotypes in female students, but not in male students. Nowadays, experiential avoidance \citep{Hayes1996} is the primary target of standard interventions for anxiety disorders, and it is considered both a vulnerability and maintenance factor in models of functioning for a variety of anxiety disorders \citep{kashdan2006experiential, dymond2019overcoming, vandenhout2014behavior}.
Experiential avoidance can also have negative repercussions on sustaining math anxiety \citep{beilock2010female}, but how could it be measured and captured in students dealing with math learning?

Methods from cognitive network science can provide a deeper understanding of students’ thought processes \citep{Doczi2019}, since mental representations of knowledge are part of a much more complex system, known as the mental lexicon. The mental lexicon can be defined as a repository of knowledge suited for the acquisition, processing, and use of all information that can be expressed via language \citep{Doczi2019}. A key aspect of the mental lexicon is associative knowledge, a cognitive subsystem that interfaces with semantic and autobiographical memory, mapping similarities between conceptual representations, and driving recall, as well as the processing of conceptual information \citep{collins1975spreading, laiacona1993perceptual, stella2024cognitive}. Numerous cognitive processes, such as word learning \citep{hills2009longitudinal, stella2017multiplex}, creativity \citep{kenett2019semantic, stella2019viability}, and domain-specific competence levels \citep{tyumeneva2017distinctive, siew2020applications}, among many others, can all be investigated by representing associative knowledge as a network of concepts. Networks of concepts can then be further analysed quantitatively using graph theory and network analytical techniques \citep{siew2019network}.

Behavioural forma mentis networks (BFMNs) are complex networks modelling how individuals structure their associative knowledge about the external world \citep{stella2019viability, stella2020forma}. BFMNs owe their name to their nature as proxies of someone’s mindset (\cite{smeets2006effect}; "forma mentis" stands for mindset in Latin), captured in terms of associating ideas/concepts and attributing valence to them. The strength of BFMNs lies in their ability to represent knowledge structures via memory recall patterns, also known as free associations \citep{DeDeyne2013}, and going beyond the necessity of defining specific associative constraints. Free associations can be represented as multidimensional links of memory recall patterns between ideas, combining semantic, syntactic, phonological, but also visual, experiential or simply co-occurring features of conceptual associations in someone’s cognitive memory \citep{Citraro2023}. BFMNs use free associations to structure how individuals’ or groups’ associative knowledge is shaped around key target ideas (e.g. cues that elicit someone’s memory and start the creation of links between cues and responses). On top of this cognitive dimension, BFMNs also encapsulate an affective perspective, according to whether individuals/groups perceive their recalled ideas positively, negatively or neutrally valenced. Consequently, affective perceptions in terms of conceptual valence can alter the quality of free association structures of conceptual associations. For example, positive concepts might tend to link with each other, while a negative concept might be contrasted with other positive ideas. BFMNs have been used in the past to assess negative attitudes towards science among high school students and researchers \citep{stella2019viability}, and can also be employed at different times to assess changes in individual or group-level mindsets over time \citep{stella2020forma}.

BFMNs have also been applied in the emerging field of machine psychology (i.e. a field lying at the intersection of psychology and computer science) to investigate the mindsets of Large Language Models (LLMs; \cite{abramski2023cognitive}). As artificial intelligences (AIs) learn conceptual associations from vast amounts of human texts, LLMs can mirror alterations in their perception of specific concepts or in their ways to associate ideas while answering questions \citep{sasson2023mirror}, analogously to humans. In previous research, BFMNs showed that OpenAI’s ChatGPT systems GPT-3, GPT-3.5 and GPT-4 all framed math in negative terms, providing stereotypical associations (e.g. "math" - "boring") reminiscent of those produced by high school students affected by math anxiety \citep{abramski2023cognitive}. The mechanisms through which LLMs like OpenAI’s ChatGPT systems can reproduce negative attitudes resembling human ones are not yet clear \citep{anoop2022bias}, nor is it understood how interaction with the models themselves can reinforce anxious mechanisms and stereotypes regarding math topics. What is more evident is that LLMs have specific mindset representations and associative knowledge, which can be accessed and investigated through tasks based on linguistic instructions \citep{sasson2023mirror}. This means that LLMs can also be used as digital twins (i.e. simulations of real-world counterparts that can be tweaked and tuned to explore alternative scenarios or compare empirical real-world data against computer simulations; \cite{singh2021digital}).

The current study aims to investigate the relationship between math anxiety and the structure of individuals’ mindsets (as captured by BFMNs) in three samples of undergraduate students and their digital twins (i.e. students simulated by LLMs). We captured math anxiety by means of a psychometric questionnaire, i.e. Math Anxiety Scale (MAS; \cite{Hunt2011, franchino2025network}), and built BFMNs according to the same protocol introduced in \citet{stella2019viability}. Furthermore, simulated psychology students — generated using ChatGPT’s GPT-3.5 and GPT-4o — were used to verify the validity of the observed relationship between psychometric scores and BFMN structure. If the responses of simulated participants differed from those of human participants, this would suggest that the observed differences were not merely due to the wording of the task instructions, but rather to psychological mechanisms present in humans and absent in machines.

This study serves as an exploratory investigation of: 
\begin{enumerate}
    \item Individual differences in perceiving math anxiety. How do students (with high or low levels of math anxiety) frame, structure, and perceive "math" and "anxiety" in relationship with other concepts? 
    
    The individual analysis is grounded in the ability to reconstruct individual mindsets as collections of associations and perceptions between hundreds of concepts. 
    \item In group-level distinctions between collective mindsets. How do clusters of students with either high or low levels of math anxiety perceive and associate ideas related to STEM subjects? 
\end{enumerate}
We compare human results against LLM-generated simulations and discuss the repercussions of this quantitative approach in view of past and current literature in psychology and educational fields. 

We will present four experiments to answer the just-mentioned research questions; the first two experiments include only human participants (i.e. two different samples of psychology undergraduates), while the last two experiments involve both human students and GPT-3.5 and GPT-4o simulated students. Below, we discuss the methodology commonly used for the first three experiments.

\section{Methods} \label{Methods}
\subsection{Materials}
The first three experiments share the same methodology in terms of analysis and procedure, while they differ in the participant sample and materials used. 

In all the participant samples, each participant (either human or GPT-simulated) had to perform the following tasks:
\begin{itemize}
    \item \textit{Completion of the MAS-IT questionnaire.} The Mathematics Anxiety Scale–IT is the Italian translated scale of the MAS-UK questionnaire \citep{Hunt2011} and has been validated in an Italian undergraduate population to assess math anxiety \citep{franchino2025network}. The factors originally identified by the MAS-UK, which are considered in the current study, are the following:
    \begin{enumerate}
        \item Evaluation MA (Math Anxiety), which encompasses anxiety associated with the evaluation of mathematical ability, often in formal academic settings (e.g. taking a math exam, answering a question in front of a class).
        \item Everyday/Social MA, which includes anxiety experienced in everyday situations involving math, often with social implications (e.g. calculating change, splitting bills, remembering phone numbers).
        \item Passive Observation MA, which refers to anxiety triggered by the passive observation of math-related activities, even without direct participation (e.g. watching someone solve a problem, listening to a math class) \citep{Hunt2011}.
    \end{enumerate}
    Participants were asked to rate their anxiety level on a 5-point Likert scale, ranging from 1 ("not anxious at all") to 5 ("very anxious"), facilitating a comprehensive, yet approximate quantification of math anxiety levels as expressed by individuals. All the original and translated items of the MAS questionnaire can be found in Table \ref{tab:mas_it_uk_items}.
    \item \textit{Performance of a free association task.} Different cue words were selected as stimuli, to which participants had to associate the first three words that came to their mind when reading them. Participants were instructed to perform the free association task as quickly as possible to preserve the spontaneity and effectiveness of the free association method \citep{DeDeyne2013}. For the same reason, if a participant was unable to provide one or more associations, they were allowed to leave those responses blank.
    \item \textit{Valence attribution task.} For all the words provided in the free association task and the cue words (i.e. if a participant did not leave any association blank, the total was of 160 words), participants had to rate their emotional valence, using a Likert scale ranging from 1 ("very negative") to 5 ("very positive"), with neutrality being represented by a score of 3. For this task, no timing instruction was given.
\end{itemize}
For the human participant samples, each participant sat in front of a computer, where they were first asked to complete an informed consent form outlining key information about privacy and data protection. The entire experiment lasted approximately 40 minutes.

As for the GPT simulated students, the model was prompted to impersonate an undergraduate psychology student, and consequently, it generated the answers to the different tasks, coherently with its personification. All the details of this process will be described in Section \nameref{Methods - Experiment 3}.

\subsection{Analysis} \label{Analysis}
\subsubsection{Behavioural forma mentis networks} \label{Forma mentis networks}
\paragraph{Network construction}
Each participant read cues, provided responses and rated, in terms of valence, both cues and responses. The resulting data, encapsulated in a data frame, was used to build a BFMN for each participant. Links were built between cues and responses (responses were not linked to each other), following the past protocol established by \citet{stella2019viability}, and inspired by past works with free association data \citep{DeDeyne2013}. Links were built as undirected, indicating a memory recall event involving both the cue and the response in the same way. 

Numerical valence ratings were transformed into categorical "positive", "negative" and "neutral" valence labels (corresponding respectively to the values: 1, -1 and 0) according to the protocol on individual-level BFMNs established by \citet{stella2020forma}. For each participant, the lower (Q1) and upper (Q3) quartiles were computed on the distribution of all valence ratings attributed to cues and responses. Concepts possessing more than one rating (i.e. responses being repeated twice or more) were attributed a median rating. Concepts with a rating strictly lower than Q1 or equal to the minimum value in the data were assigned a negative valence label. Concepts with a rating strictly higher than Q3 or equal to the maximum value in the data were assigned a positive valence label. A neutral valence label was assigned to the remaining concepts. This approach extracted valence labels, adapting to individuals’ affective biases; e.g. an individual rating concepts with '3s' to express negative, rather than neutral, attitudes could have a distribution skewed towards higher ratings, and this would be captured by quartiles. This process led to the creation of one forma mentis network per respondent. 

\paragraph{Network features: Degree}
BFMNs are cognitive networks where nodes represent concepts and links indicate memory recall patterns. Cues and responses are treated equally as nodes. Every node can have its own "degree" (also called degree centrality), a local network measure corresponding to the number of links a node has \citep{siew2019network}. The higher the degree is, the more connections a given node is involved in. In BFMNs, the degree indicates the semantic richness of memory recalls attributed to a given concept (e.g. a node with degree 5 has been involved in at least 5 different memory recalls; it has been associated with 5 different concepts \citep{stella2020forma}). Degree centrality considers only connections in the immediacy of a target node, neglecting the rest of the network structure. 

\paragraph{Network features: Closeness centrality}
While node degree maps the local connections a concept engages in, cognitive networks can have more global features, going beyond local connectivity and engaging the whole network structure at the same time \citep{stella2020forma}. One could consider how many connections separate any two target nodes, i.e. how many memory recalls have to be traversed to go from one concept to the other in a network of free associations. Shortest path length is the smallest number of connections separating any two concepts. If there is no way to reach one node from another, then nodes are said to be disconnected, and their shortest path length is assumed to be infinity. One could consider the average shortest path length between one node and all its connected neighbours as a measure of how close a node is to the others. 

Closeness centrality (\ref{closeness centrality}) is defined as the inverse of the average shortest path length between node ($i$) and all other ($N$) nodes connected to it in the network \citep{siew2019network}. In networks of free associations, the closeness centrality of concepts was found to influence early word learning \citep{hills2009longitudinal}. The formula of closeness centrality is the following:
\begin{equation}
    C_i = \frac{N - 1}{\sum_{j} d(n_i, n_j)}, \label{closeness centrality}
\end{equation}
where $i$ refers to one node and $j$ to another, $N$ corresponds to the number of nodes in the graph, and $k(n_i, n_j)$ to the shortest path length connecting node $i$ to node $j$.

\subsubsection{Correlation and Prediction of Math Anxiety}
We use BFMNs to capture the mindset of every student. We aim to investigate how local structure (degree centrality) and affective perceptions (valence ratings) in someone’s mindset can differ according to different levels of math anxiety. We thus focus on the concepts of "math" and "anxiety" to extract BFMN measures of these concepts, to perform a regression analysis of math anxiety levels.

To examine the interrelationship among network measures and math anxiety factors, Pearson's correlation coefficients were calculated, and a correlogram was built. Specifically, we considered the MAS-UK factors (see \nameref{Methods} Section): Evaluation MA, Everyday/Social MA, and Passive Observation MA; and their sum, referred to as Total MA. As for network measures, we took into account the degree, closeness centrality (see \nameref{Forma mentis networks} Section above) and valence of the nodes "anxiety" and "math".

Moreover, we performed four independent Ordinary Least Squares (OLS) linear regression models to investigate whether network features can predict levels of math anxiety. As predictors, we included either three (i.e. the degree and valence of the node "anxiety" and the valence of the node "math") or four (i.e. the degree and valence of both the "anxiety" and "math" nodes) network features. The choice of sometimes excluding the degree of "math" as a predictor was based on the observation that in the Pearson's correlation matrices of both human samples from \nameref{Experiment 1} and \nameref{Experiment 2}, the degree of "math" was not significantly correlated with any of the MA factors. To determine whether the model with three or four predictors provided a better fit, we computed the Akaike Information Criterion (AIC) for both models and selected the one with the lower AIC value.  As for the dependent variable of the model, we considered the 3 factors of the MAS-IT and the Total MA. In the end, we calculated four linear regressions: one for each MA factor and the Total MA. All the variables (both the predictors and the dependent variable) were standardised.

\section{Experiment 1} \label{Experiment 1}
\subsection{Introduction}
This study explores undergraduate psychology students’ mental representations of mathematical concepts through a free association task, focusing on the affective tone of the associations. In this task, participants are given a cue word and are asked to provide the first three words that come to mind in response.

Research has highlighted the significant role of cognitive networks in shaping perceptions of STEM subjects. For example, \citet{stella2019viability} found that subjects as "math" and "physics" were frequently associated with negative emotions, contributing to a "negative emotional aura" and reinforcing a simplistic and unfavourable view of these fields. This finding aligns with the broader literature on math anxiety and its impact on cognitive processes. Moreover, \citet{beilock2010female} demonstrated that math anxiety can impair cognitive functioning and problem-solving abilities, emphasising the detrimental effects of negative emotional states on mathematical performance. This is supported by \citet{Chang2016}, who further elucidated how math anxiety disrupts working memory and increases avoidance behaviour in math-related tasks. 

Additionally, \citet{caviola2022math} provided insights into how math anxiety affects students’ performance and attitudes, finding that high levels of math anxiety are associated with poorer performance and more negative associations with mathematical tasks. Their work underscores the need to explore how these negative associations are formed and maintained within cognitive networks.

While these studies have significantly advanced our understanding of math anxiety and its consequences, there remains a need for further exploration into how cognitive networks, specifically related to math, are influenced by varying levels of anxiety. This experiment aims to investigate the characteristics of cognitive networks in a sample of undergraduate psychology students by exploring how perceptions and associations related to math vary in relation to levels of math anxiety. By employing a free association task to map cognitive networks, this research uncovers individual and group differences in the perception and association of mathematical ideas. The goal is to deepen our understanding of the psychological mechanisms underlying math anxiety and its cognitive and emotional implications, thereby contributing valuable insights into the interplay between anxiety, cognition, and academic performance.

\subsection{Participants and Materials} \label{Materials - Experiment 1}
75 undergraduate students from the Department of Psychology and Cognitive Science at the University of Trento (Italy) were recruited through convenience sampling using the department's social media channels and word of mouth. All participants were adults and native Italian speakers. The protocol was approved by the Human Research Ethics Committee of the University of South Australia (ID: 205244) and complied with the Declaration of Helsinki. Participants signed an informed consent and did not receive any compensation for their participation.

We cleaned the data, removing all the participants who did not provide any associations for the cue words "math" and "anxiety", which were fundamental for the network analysis. We also excluded participants with more than 1/3 of the data missing. In the end, only 70 participants out of the initial 75 were considered for the analysis.

Each participant followed the procedure described in the \nameref{Methods} Section. In particular, for Experiment 1, we selected 40 words as stimuli (i.e. cue words), which can be clustered into 4 groups, consisting of words related to:
    \begin{enumerate}
        \item STEM and Academic Disciplines
        \item Psychology and Mental Health
        \item Academic Evaluation
        \item Personal Attitude
    \end{enumerate}
The 40 cue words of the current experiment are reported in Table \ref{tab:cue_words_exp_1}. As for the analyses performed to obtain the results of Experiment 1, we refer to the \nameref{Analysis} Section.

\subsection{Results}
In Table \ref{tab:descriptive_stats_network} we report the statistics of the main network features described in Section \nameref{Forma mentis networks}. There we see that the degree of "math" had a mean of 6.53 (SD = 1.21), while the degree of "anxiety" had a mean of 6.29 (SD = 1.91), meaning that both "math" and "anxiety" are strongly connected to other words in students' BFMNs. The slightly higher degree for "math" suggests that it is linked to a wider range of concepts, while the strong connections to "anxiety" may reflect how emotionally important or present it is in students' minds. The closeness centrality of the word "math" had a mean of 0.09 (SD = 0.05), while the closeness centrality of the word "anxiety" had a mean of 0.08 (SD = 0.06) (cf. Table \ref{tab:descriptive_stats_network}). Both "math" and "anxiety" are equally central within the students' BFMNs, meaning that these two concepts were just as easily reachable from other ideas in the network, reflecting their overall importance or accessibility in students' thinking.

Furthermore, in Table \ref{tab:descriptive_stats_MA}, one can find the descriptive statistics of the three math anxiety factors and the total math anxiety.

\subsubsection{Correlations Analysis}
\begin{figure}[!htbp]
    \centering
    \includegraphics[width=1\linewidth]{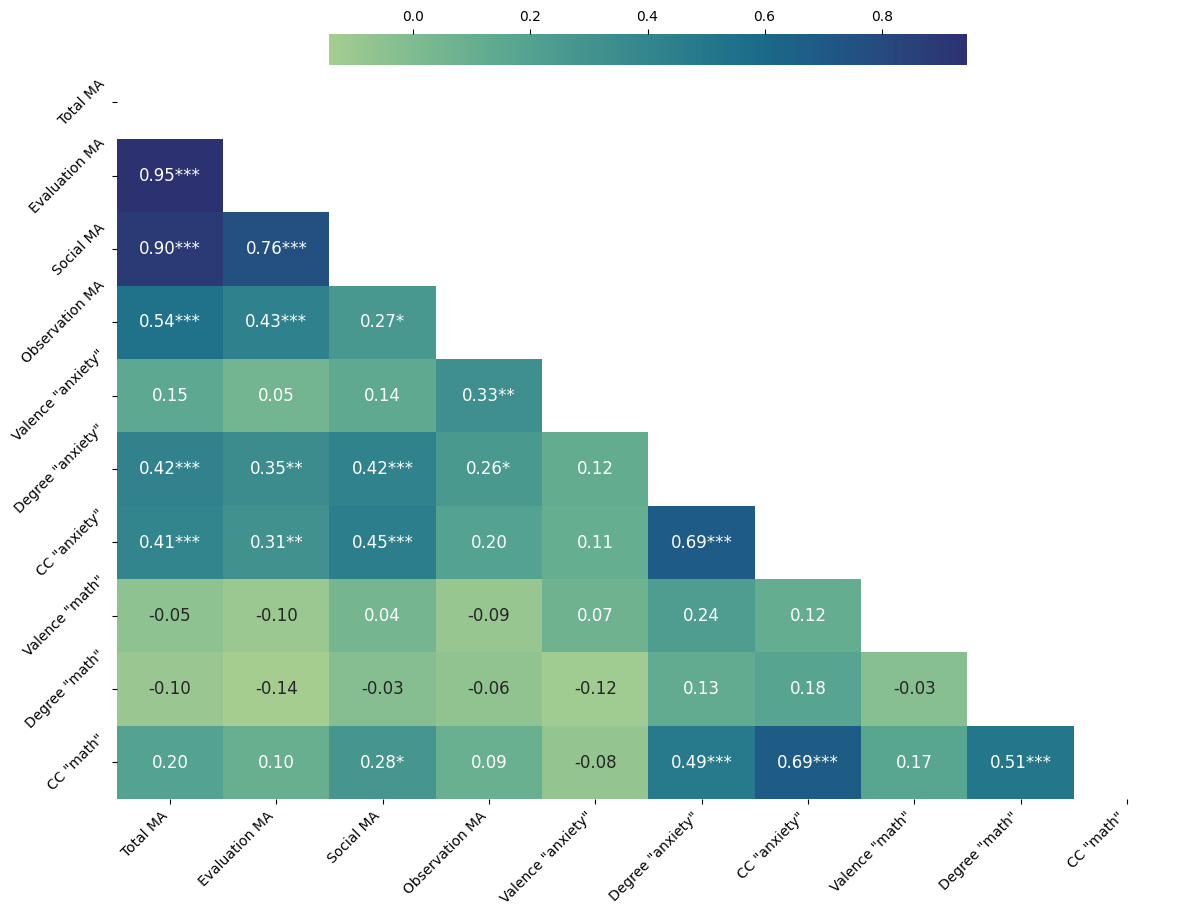}
    \caption{Correlogram Experiment 1. The valence, degree and closeness centrality (CC) for the nodes "anxiety" and "math" are reported, along with the factors of MA: Evaluation, Everyday/Social (Social) and Passive Observation (Observation), together with the Total MA.}
    \label{fig:correlogram_exp_1}
\end{figure}

The correlograms in Figure \ref{fig:correlogram_exp_1} examine the interrelationships among the math anxiety and network measures. Bivariate relationships between variables were inspected through zero-order Pearson's correlations, with significance level fixed to $\alpha$ = .05 ($N$ = 70).
Looking at Figure \ref{fig:correlogram_exp_1}, one can observe a persistent trend in the significant correlations that were found for each math anxiety factor and its totality. Specifically, across all math anxiety dimensions, significant correlations emerged with the network features of the node "anxiety". Total, Evaluation, and Everyday/Social MA were all significantly positively correlated to its degree ($r = .42$, $p < .001$; $r = .35$, $p < .01$; $r = .42$, $p < .001$, respectively) and closeness centrality ($r = .41$, $p < .001$; $r = .31$, $p < .01$; $r = .45$, $p < .001$, respectively). On the other hand, the Passive Observation factor showed a slightly different pattern, correlating positively with the degree ($r = .26$, $p < .05$), as well as with the valence ($r = .33$, $p < .01$) of "anxiety". Notably, no significant correlations were found between any MA factor, nor Total Ma and the network features of "math".

These results indicate that the number of associations attributed to "math" and its valence do not convey information correlating with math anxiety. Conversely, the valence ratings attributed to the word "anxiety" and the degree of this node showed a stronger correlation with different factors of math anxiety, suggesting that the emotional meaning and the number of connections made with this node relate closely to math anxiety.

Furthermore, closeness centrality and degree are highly correlated with each other and, in a regression model, might cause issues relative to multicollinearity. For these reasons, we selected only valence ratings and the network degree of "anxiety" and "math" as predictor variables for an OLS linear regression model, aiming at determining math anxiety scores.

\subsubsection{Linear Regressions Analysis}
We performed four independent OLS linear regression models to investigate how network characteristics relate to math anxiety, as described in the \nameref{Methods} Section.

In Tables from \ref{tab:lin_reg_total_MA} to \ref{tab:lin_reg_obs_MA} in the line "Experiment 1 Students", all the linear regression models' results are reported (one table for each MA factor and for Total MA). For Total and Evaluation MA, we included four predictors (AICs = $-36.814$ and $-23.538$, respectively). The regression models of these MA dimensions explained 23.8\% ($R^2 = .238$) and 19.9\% ($R^2 = .199$) of the variance, with the degree of "anxiety" being the only significant predictor ($\beta = .542$, $p < .001$; $\beta = .414$, $p < .001$).

In contrast, for the Everyday/Social and Passive Observation factors, the degree of "math" was excluded (AICs = $-45.751$ and $1.970$, respectively). The Everyday/Social model accounted for 19.2\% of the variance ($R^2 = .192$), again showing a significant effect of the degree of "anxiety" ($\beta = .740$, $p < .001$). The Passive Observation model explained 18.5\% of the variance ($R^2 = .185$) and, in addition to the degree of "anxiety" ($\beta = .390$, $p = .026$), the valence of "anxiety" also emerged as a significant predictor ($\beta = .452$, $p = .007$). 

Altogether, these results indicate that the structural prominence of "anxiety" within the network—reflected by its degree—consistently predicts math anxiety levels across its different dimensions, while the emotional valence of "anxiety" additionally contributes to Passive Observation MA.

\subsection{Discussion}
The results obtained from this sample of psychology undergraduates reveal a consistent relationship between the network characteristics (i.e. degree and closeness centrality) of "anxiety" and levels of math anxiety. Across correlation and regression analyses, the degree of "anxiety" emerged as the most reliable predictor, indicating that students who more frequently associated "anxiety" with other concepts reported higher math anxiety across its different dimensions. This finding suggests that the cognitive salience of anxiety within one’s semantic network — its structural prominence and accessibility — may reflect elevated math anxiety.

Additionally, the emotional valence of "anxiety" contributed uniquely to Passive Observation MA, highlighting how affective perceptions influence more passive or observational facets of math-related anxiety. Interestingly, the network features of "math" itself, including its degree and valence, did not significantly relate to math anxiety in this sample. This absence of association could suggest either a general conceptual distancing from "math" or that the emotional meaning of anxiety plays a more central role in shaping math anxiety than the semantic richness of "math".

Overall, these findings suggest that math anxiety is rooted not only in emotional responses to mathematics but also in the affective and structural organisation of semantic memory. A more interconnected and negatively valenced representation of "anxiety" appears to mirror stronger math-related fears, consistent with mechanisms such as rumination, which depletes working memory and heightens math anxiety \citep{lefevre2005mathematical, stella2022network}. This aligns with evidence that the semantic richness of negative concepts sustains maladaptive emotional states and that emotional reasoning—concluding affect rather than logic \citep{Gangemi2021}—may reinforce the negative encoding of math-related experiences.

\section{Experiment 2} \label{Experiment 2}
\subsection{Introduction}
In this experiment, the same paradigm as in \nameref{Experiment 1} was employed. Therefore, cue words and free associations were evaluated with affective valence scores to investigate associative memory of concepts relative to STEM subjects and math anxiety. However, the task differs in the actual cue words used to test the associative knowledge structure of the current student sample. The cues in this experiment were chosen to be more closely related to educational settings and self-regulated learning because of the rich, relevant literature highlighting how learning environments and cognitive strategies can display rich interplay with math anxiety. In fact, teachers can greatly influence their students and can even become crucial actors spreading math anxiety (cf. \cite{beilock2010female}). Even parents can pass math anxiety along with their interactions with children \citep{stella2022network}. However, math anxiety can be countered by building a variety of cognitive and meta-cognitive strategies, including self-regulated learning \citep{Gabriel2020}, one’s ability to understand, improve and master their own learning pace and environment by setting goals, monitoring progress and reinforcing weaknesses. Stronger self-regulated learning shows a beneficial effect against math anxiety \citep{Gabriel2020}.

We provided cues related to actors in educational settings and self-regulated learning to differentiate the overall setting of the BFMN free association and valence rating tasks of \nameref{Experiment 1}. In our reasoning, \nameref{Experiment 2} serves as a straightforward replication of the results in \nameref{Experiment 1} but within a different sample of undergraduate psychology students and a different set of starting cue words, always including "math" and "anxiety".

\subsection{Participants and Materials}
We recruited 80 psychology students via convenience sampling. All participants were adults and native Italian speakers. None of the participants in the present experiment took part in \nameref{Experiment 1}. As in \nameref{Experiment 1}, we excluded all the participants who left all the associations for the cue words "math" and "anxiety" blank, as well as the ones with more than 1/3 of the data missing. Following this procedure, only 57 participants were considered for the analyses.

The current experiment employed the same set of procedures reported in the \nameref{Methods} Section, ensuring that any observed differences in outcomes between the two experiments can be confidently attributed to the manipulated variables rather than variations in measurement tools or procedures. The cue words selected for the current experiment differ from those in \nameref{Experiment 1} and reflect these specific categories:
\begin{enumerate}
    \item STEM disciplines
    \item Specific actors, places and actions of education (e.g. teacher, class)
    \item Aspects of self-regulated learning and learning motivation (e.g. plan, delay, commitment)
    \item Key aspects of math anxiety (e.g. blackboard, equation)
\end{enumerate}
All the cue words used in this experiment can be found in Table \ref{tab:cue_words_exp_2}.

The analyses employed in the current experiment are the same as \nameref{Experiment 1} and are detailed in the \nameref{Methods} Section.

\subsection{Results}
The results for the main network measures (i.e. closeness centrality, degree, valence) can be found in Table \ref{tab:descriptive_stats_network}. There, we can see that the node "math" had a mean degree of 7.53 (SD = 2.05), and a mean closeness centrality of 0.05 (SD = 0.05). The node "anxiety" had a mean degree of 7.19 (SD = 2.89), and a mean closeness centrality of 0.09 (SD = 0.06). These statistics suggest that "math" and "anxiety" are relatively well-connected nodes in the network, with "math" showing slightly higher connectivity and centrality, indicating it may play a more central role in the semantic structure.

Table \ref{tab:descriptive_stats_MA} shows the descriptive statistics of math anxiety (specifically for each of the three MA factors and the Total MA) for the current student sample.

\subsubsection{Correlations Analysis}
\begin{figure}[!htbp]
    \centering
    \includegraphics[width=1\linewidth]{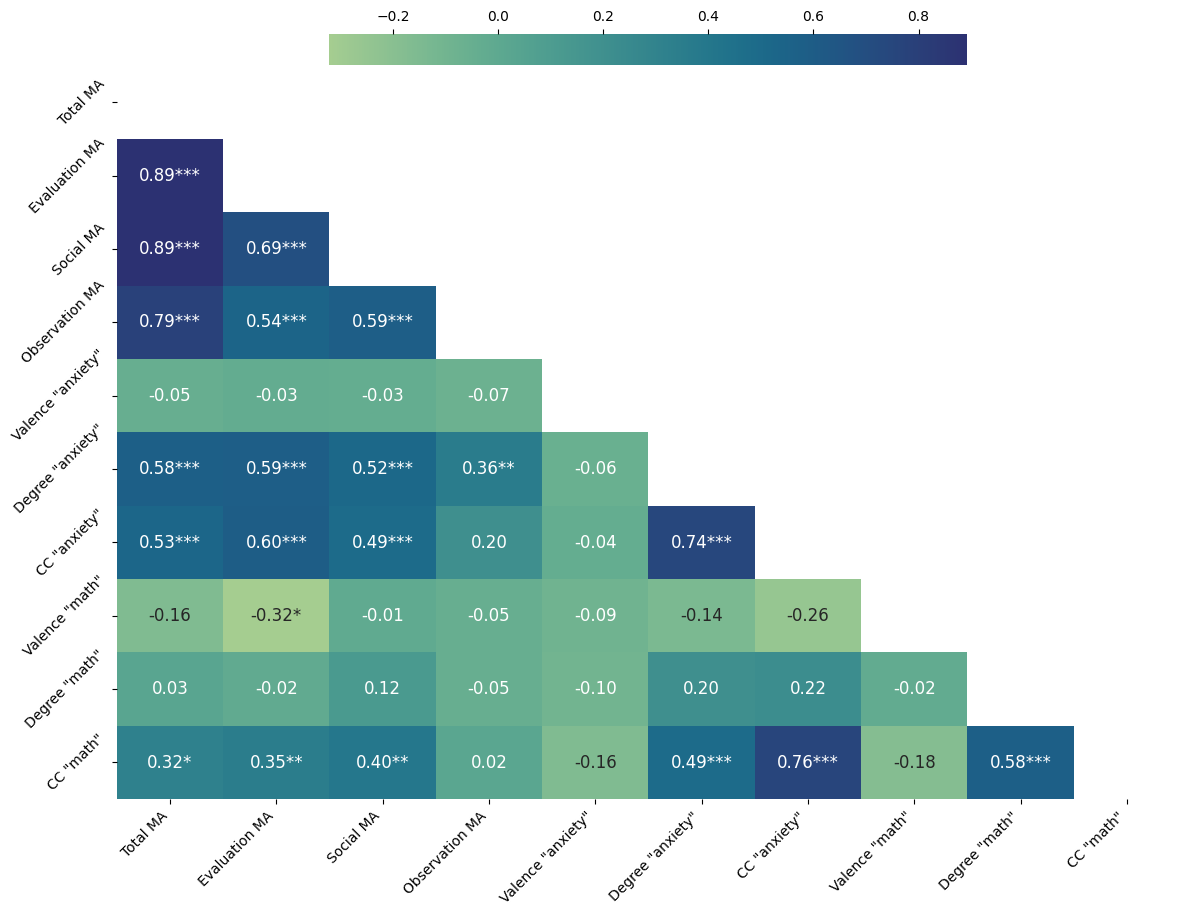}
    \caption{Correlogram Experiment 2. The valence, degree and closeness centrality (CC) for the nodes "anxiety" and "math" are reported, along with the factors of MA: Evaluation, Everyday/Social (Social) and Passive Observation (Observation), together with the total MA.}
    \label{fig:correlogram_exp_2}
\end{figure}

The correlogram in Figure \ref{fig:correlogram_exp_2} examines the interrelationships among the MA factors and Total MA (obtained from the MAS-IT questionnaire filling) with local and affective network characteristics for both the words "math" and "anxiety" across the \nameref{Experiment 2} population. 
From Figure \ref{fig:correlogram_exp_2}, one can observe that, as in \nameref{Experiment 1}, here too, some consistent patterns emerged across all math anxiety dimensions. Specifically, all the MA dimensions showed significant positive correlations with degree of "anxiety" (Total MA: $r = .58$, $p < .001$; Evaluation MA: $r = .59$, $p < .001$; Social MA: $r = .52$, $p < .001$; Passive Observation: $r = .36$, $p < .01$). Additionally, Total, Evaluation and Social present a significant positive correlation with the closeness centrality of the node "anxiety" (Total MA: $r = .53$, $p < .001$; Evaluation MA: $r = .60$, $p < .001$; Social MA: $r = .49$, $p < .001$).

As for the node "math", we can notice that all the dimensions that presented significant correlations with the closeness centrality of "anxiety", also presented them with the closeness centrality of "math" (Total MA: $r = .32$, $p < .05$; Evaluation MA: $r = .35$, $p < .01$; Social MA: $r = .40$, $p < .01$). Furthermore, the valence of "math" also negatively correlates with the Everyday/Social MA factor ($r = -.32$, $p < .05$).

Overall, these results mirror those observed in \nameref{Experiment 1}, indicating that higher levels of math anxiety —across all factors and its totality— are associated with a greater number of conceptual links to "anxiety" and its emotional valence. Moreover, the negative relationship between Evaluation MA and the valence of "math" suggests that students experiencing evaluation-related anxiety tend to perceive "math" more negatively.

\subsubsection{Linear Regressions Analysis}
Based on the correlational results, we analysed four independent OLS linear regression models to investigate how network characteristics relate to Total MA and the three relative factors, as described in the \nameref{Methods} Section. All the results of the linear regression models are reported in Tables \ref{tab:lin_reg_total_MA}-\ref{tab:lin_reg_obs_MA} (one for each MA dimension), in the line "Experiment 2 Students". Aligning with \nameref{Experiment 1}, a consistent pattern across all math anxiety factors and Total MA was found, with the degree of "anxiety" emerging as a significant predictor of MA levels. 

For Total and Evaluation MA, models with three predictors provided the best fit (AICs = $-21.532$ and $-28.252$, respectively), explaining 34.7\% ($R^2 = .347$) and 40.5\% ($R^2 = .405$) of the variance. In both cases, the degree of "anxiety" was a significant positive predictor ($\beta = .642$, $p < .001$; $\beta = .617$, $p < .001$), while for Evaluation MA, the valence of "math" also contributed significantly ($\beta = -.144$, $p = .027$), indicating that a more negative perception of "math" is associated with higher evaluation-related anxiety.

On the other hand, for the Everyday/Social and Passive Observation MA models, we considered four predictors (AICs = $9.618$ and $-4.665$, respectively) and accounted for the 27.2\% ($R^2 = .272$) and 13.1\% ($R^2 = .131$) of the variance, respectively. Both of these MA factors strengthen the results of the previously described MA dimension, showing a significant effect of the degree of "anxiety" (Everyday/Social MA: $\beta = .740$, $p < .001$;  Passive Observation MA: $\beta = .403$, $p = .008$).

These findings align with the ones of the first experiment, and underscore the central role of "anxiety"'s network connectivity in predicting different forms of math anxiety, whilst negative emotional evaluations of "math" appear to heighten specifically evaluation-related anxiety.

\subsection{Discussion}
The results of \nameref{Experiment 2} largely replicate findings of \nameref{Experiment 1}, even though they were obtained from a different sample and with BFMNs that present different contextual cues surrounding "math" and "anxiety". They reinforce the strong association between the structural prominence of "anxiety" within participants’ semantic networks and math anxiety levels. Across correlation and regression analyses, the degree of "anxiety" consistently emerged as a significant predictor, indicating that students who form richer associative links with "anxiety" tend to experience higher math anxiety. This supports the notion that math anxiety reflects not only an emotional response to mathematical tasks but also a more deeply rooted cognitive-emotional schema, where anxiety is conceptually central and easily activated.

Moreover, the valence of "math" was specifically related to Evaluation MA, with more negative emotional appraisals of "math" corresponding to stronger evaluation-related anxiety. This pattern suggests that evaluative contexts—such as tests or performance comparisons—may elicit affective reactions grounded in the negative emotional meaning attributed to mathematics itself. In contrast, other math anxiety dimensions were primarily linked to the structural rather than affective characteristics of "anxiety", underscoring the multifaceted nature of math anxiety as both a cognitive and affective construct.

As in \nameref{Experiment 1}, these results point to the intertwined contribution of semantic structure and emotional valence in understanding the mechanisms underlying math anxiety.

\section{Experiment 3} \label{Experiment 3}
\subsection{Introduction}
The advent of LLMs has markedly expanded the potential for simulating aspects of human cognition in research settings. LLMs are AIs trained on extensive datasets to produce human-like written content \citep{kasneci2023chatgpt}. LLMs are also capable of interacting with humans through conversations, thus they can understand and act upon instructions to psychological experiments that can be fully explained in words \citep{abramski2023cognitive}. Such capabilities make LLMs highly valuable for generating synthetic data emulating human responses, offering substantial potential to transform research methodologies in various disciplines, including psychology \citep{binz2023using}.

Given the valence and degree patterns in previous studies, we decided to further investigate math anxiety via BFMNs in a third sample of simulated human participants. This study employs ChatGPT, more specifically GPT-3.5 and GPT-4o, to emulate the responses of university psychology students, allowing us to explore human versus simulated cognitive patterns relative to math anxiety. \nameref{Experiment 3} thus follows the growing literature on machine psychology \citep{abramski2023cognitive, binz2023using}, using psychometric tools employed in the previous experiment (e.g. BFMNs, questionnaires) to investigate the mindsets of AIs as cognitive agents.

\subsection{Methods} \label{Methods - Experiment 3}
GPT simulated students follow the methodology described in the \nameref{Methods} Section and adopted in \nameref{Experiment 1} and \nameref{Experiment 2} for both the tasks they had to perform, and the statistical analysis conducted on the collected data. The main difference from the previous experiments lies in the simulation of GPT participants and the collection of their responses. These procedures will be detailed below.

\subsubsection{Prompt Engineering} \label{Prompt Engineering}
In this study, we aimed to replicate the original methodology of BFMNs established by \citet{stella2019viability}, but within a prompt engineering framework using LLMs. To achieve this, we developed a simulation leveraging Python as a programming language and the OpenAI API to interface with the GPT-3.5-turbo-0125 and GPT-4o language models. This simulation acted as a virtual laboratory, allowing us to systematically generate and analyse synthetic data that closely resembled the responses of human participants in the original study \citep{stella2019viability}.

Simulations were based on a prompt that necessitated several revisions and refinements to achieve an optimal balance between consistency and variability in the generated responses. Initial versions of the unpolished prompt frequently yielded responses that lacked the complexity and heterogeneity of human-generated data (e.g. incomplete responses or responses repeated across all questions). Subsequent iterations emphasising clarity, conciseness, and specificity allowed for more faithfully human-mirrored responses, avoiding structurally ambiguous, incomplete or incoherent outputs.

Both GPT-3.5 and GPT-4o can impersonate individuals and modify their responses based on personification \citep{kasneci2023chatgpt}. The role that GPT-3.5 and GPT-4o play for each simulated free association task, valence attribution task, and MAS-IT questionnaire is based on a generated profile. Such a profile is pivotal for enhancing the ecological validity and generalisation of the simulation study, as observed in real student samples. 

To embed variety in GPTs’ responses, we developed a prompting framework that assigned several human socio-demographic and educational features to the model's personification, before task instructions. These features were not used for regression analysis, but only to bolster more variety in the responses of simulated users. The adopted features were based on relevant educational literature about math anxiety \citep{stella2022network}, and they included:
\begin{itemize}
    \item Gender: this binary feature ranged only between "male" and "female", in light of past gender studies showing how male and female students can react differently to math anxiety \citep{beilock2010female};
    \item Age: this qualitative variable allows for random selection of age within realistic ranges for last year's high school and university students (18-25 y.o.). This is crucial because attitudes towards STEM subjects can evolve as students progress through their education and mature.
    \item Education level: this variable distinguishes students from either the onset of university (i.e. last year of high school) or the three years of BSc enrolment in university. This approach acknowledges that students’ perceptions and attitudes towards STEM subjects may evolve as they progress through their academic journey.
    \item Socioeconomic conditions: they were incorporated as a factor by randomly assigning students to different categories: low (in Italian: "basse"), medium-low (in Italian: "medio-basse"), medium (in Italian: "medie"), medium-high (in Italian: "medio-alte"), or high (in Italian: "alte"). This recognition reflects the potential influence of socioeconomic factors on educational opportunities, resources, and attitudes towards learning. 
\end{itemize}
The above attributes were incorporated into a single prompt, guiding GPT to respond as a student with a particular profile. The personification prompt used was the following: "\textit{Sei un}\texttt{\{gender\}} \textit{student}\texttt{\{gender\}} \textit{italian}\texttt{\{gender\}} \textit{di} \texttt{\{age\}} \textit{anni. Sei iscritt}\texttt{\{gender\}} \textit{al} \texttt{\{year\}} \textit{anno di} \texttt{\{education\}}\textit{. Sei cresciut}\texttt{\{gender\}} \textit{e vivi in condizioni socio-economiche} \texttt{\{socioeconomic\}}\textit{. Pertanto, ricorda che le risposte da fornire nel compito devono essere originali, creative e coerenti con le tue caratteristiche uniche."\footnote{English translation: "\textit{You are an}\texttt{\{gender\}} \textit{student}\texttt{\{gender\}} \textit{of Italian nationality}\texttt{\{gender\}} \textit{aged} \texttt{\{age\}} \textit{. You are enrolled in the} \texttt{\{year\}} \textit{year of} \texttt{\{education\}}\textit{. You grew up and live in} \texttt{\{socioeconomic\}} \textit{socio-economic conditions. Therefore, remember that the responses you provide in the task should be original, creative, and consistent with your unique characteristics."}}}

Additional prompting in the instructions for the psychometric task also emphasised the generation of "original, creative, and coherent" responses, aligning with each student’s unique profile. This method ensures that the simulated responses represent a diverse range of real student experiences, thereby improving the ecological validity and generalisation of the study’s findings. 

\subsubsection{Participants and Materials}
Using this prompt, we simulated a total of 300 students for each GPT model. Thus, we had 300 GPT-3.5 generated students and 300 GPT-4o generated students. We increased the sample size to roughly twice the number of human subjects to compensate for the lack of variability of AI-generated responses compared to human free associations (see \cite{abramski2023cognitive}). 

To maintain methodological consistency and facilitate meaningful comparisons with the findings of \nameref{Experiment 1}, the current experiment employed the same set of materials. This involved using the identical 40 cue words as stimuli (cf. Table \ref{tab:cue_words_exp_1}), valence rating scales, and the MAS-It questionnaire (cf. Table \ref{tab:mas_it_uk_items}), as detailed in \nameref{Methods} Section. This approach guarantees that differences in results between the two experiments are due only to the manipulated variables, and not to differences in measurement methods or procedures.

\subsubsection{Procedure} \label{Procedure}
The following steps were followed to obtain individual BFMNs and psychometric scores from simulated students:
\begin{itemize}
    \item \textit{Creating a random student profile.} Using the main prompt, the model is instructed to impersonate a realistic student profile (cf Section \nameref{Prompt Engineering}).
    \item \textit{Performing tasks.} Simulated students performed the procedures described in \nameref{Methods} Section. They completed the MAS-IT questionnaire, the free association task (with cue words used for \nameref{Experiment 1}) and the valence task, as human participants did in \nameref{Experiment 1} and \nameref{Experiment 2}.
    \item \textit{Collection of data.} The prompt emphasises a structured output format for consistent data collection, by directing the simulated student to provide their responses as a \textit{.txt} file, with each response separated by a comma (',').
\end{itemize}
For the statistical analyses conducted on the collected data, we refer to the \nameref{Analysis} Section.

\subsection{Results for GPT-3.5}
Descriptive statistics for the main network features are reported in Table \ref{tab:descriptive_stats_network}. For GPT-3.5 simulated students, the node "math" had a mean network degree of 5.16 (SD = 0.38) and a mean closeness centrality of 0.03 (SD = 0.01). The node "anxiety" had a mean degree of 5.53 (SD = 0.59) and a mean closeness centrality of 0.03 (SD = 0.02). Overall, both "math" and "anxiety" display relatively low network degrees and closeness centralities, indicating limited connectivity and influence within the GPT-3.5 semantic networks.

Furthermore, Table \ref{tab:descriptive_stats_MA} reports the descriptive statistics of math anxiety for the total math anxiety and its three relative factors.

\subsubsection{Correlations Analysis}
\begin{figure}[!htbp]
    \centering
    \includegraphics[width=1\linewidth]{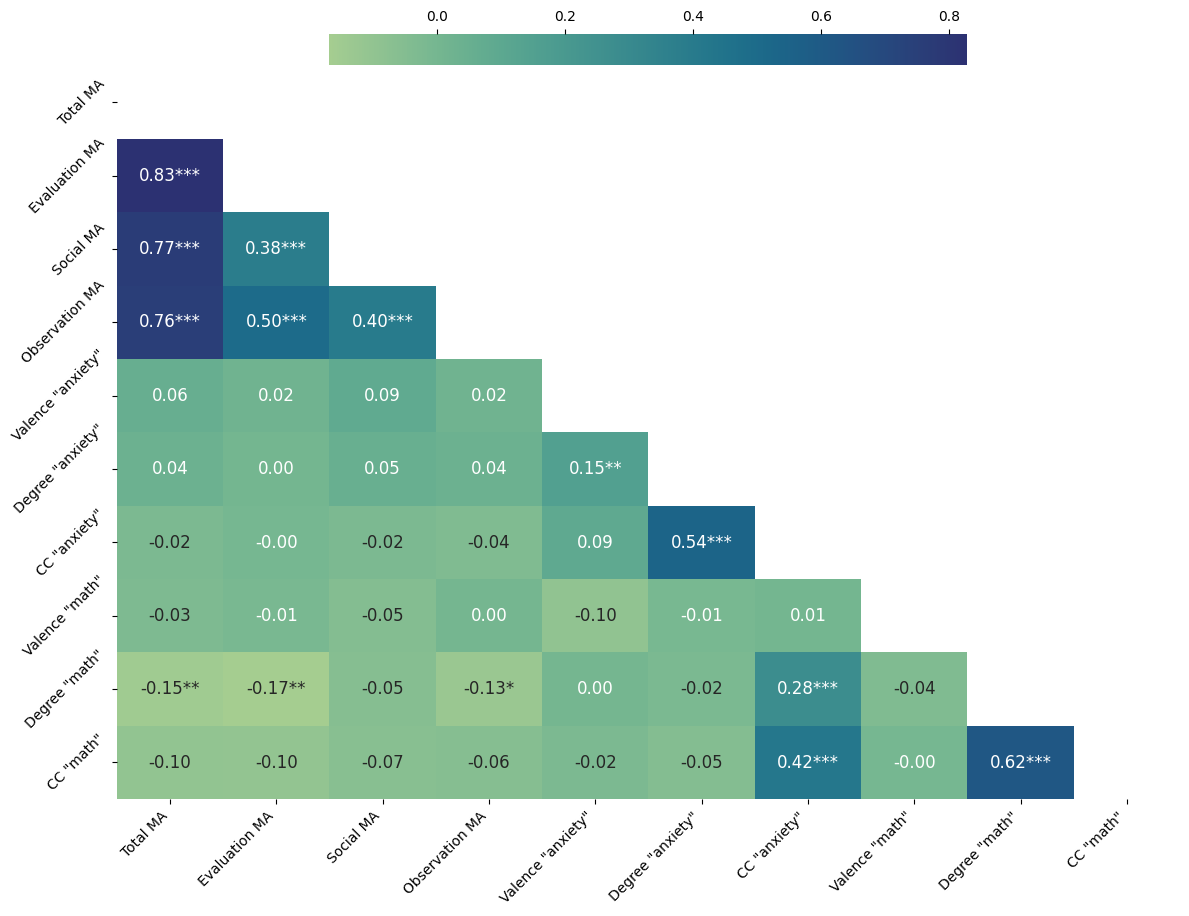}
    \caption{Correlogram Experiment 3 - GPT-3.5. The valence, degree and closeness centrality (CC) for the nodes "anxiety" and "math" are reported, along with the factors of MA: Evaluation, Everyday/Social (Social) and Passive Observation (Observation), together with the total MA.}
    \label{fig:correlogram_gpt_3_5}
\end{figure}

The correlogram in Figure \ref{fig:correlogram_gpt_3_5} examines the relationships among the three factors of the MAS-IT questionnaire and Total MA, with local and affective network characteristics (i.e. the degree, valence and closeness centrality) for the concepts "math" and "anxiety" across GPT-3.5 simulated students.

In contrast to the findings from \nameref{Experiment 1} and \nameref{Experiment 2}, the present results revealed that higher levels of math anxiety were generally associated with lower network connectivity of the concept "math". In particular, Total MA showed a significant negative correlation with the degree of "math" ($r = -.15$, $p < .01$), a pattern mirrored by Evaluation MA ($r = -.17$, $p < .01$) and Passive Observation MA ($r = -.13$, $p < .05$). In contrast, no significant associations were found between Everyday/Social MA and any of the network features of "math" or "anxiety". 

From these significant correlations, we can grasp that GPT-3.5 simulated students exhibiting higher math anxiety tend to associate the concept of "math" with fewer related concepts, reflecting a possible narrowing or reduction of conceptual links to "math".

\subsubsection{Linear Regressions Analysis}
The above patterns indicate that GPT-3.5 simulated students present fewer correlations between psychometric scores and BFMN structure compared to human students. We then analysed four independent linear regression models to investigate how network characteristics of "anxiety" and "math" relate to Total MA, Evaluation MA, Everyday/Social MA and Passive Observation MA. 

The results of the linear regression models are reported in Tables from \ref{tab:lin_reg_total_MA} to \ref{tab:lin_reg_obs_MA} (in line "GPT-3.5 Students"). 
For what concerns the predictors that we took into account, we included all four predictors for the Evaluation and Passive Observation MA models (AICs = $-368.951$, $-254.986$, $-269.904$, respectively), which respectively explained the 2.7\% ($R^2 = .027$), 2.9\% ($R^2 = .029$) and 1.7\% ($R^2 = .017$) of the variance. On the other hand, the Everyday/Social MA model included only three predictors (AIC = $-272.712$) and accounted for only 1.0\% of the variance ($R^2 = .010$). 

From the models' results, we can observe a consistent negative association between the degree of "math" and math anxiety levels, aligning with the correlation results. In particular, the significance of the predictor Degree "math" has been found for: Total ($\beta = -.104$, $p < .001$), Evaluation ($\beta = -.142$, $p = .004$) and Passive Observation ($\beta = -.103$, $p = .030$) MA models; while, as for Everyday/Social MA model, none of the predictors showed significant effects on any of the MA dimensions. 

Overall, these findings suggest that within GPT-3.5’s simulated associative network, the reduced number of conceptual links to "math" could be a significant predictor for higher levels of math anxiety.

\subsection{Results for GPT-4o}
In Table \ref{tab:descriptive_stats_network}, the descriptive statistics of network features are reported. We can observe that for GPT-4o simulated students, the node "math" presented a mean degree of 5.35 (SD = 0.50) and a mean closeness centrality of 0.03 (SD = 0.01). The node "anxiety" exhibited a mean degree of 5.63 (SD = 0.54) and a mean closeness centrality of 0.04 (SD = 0.01). Compared to GPT-3.5, GPT-4o simulated students show higher network degrees for both nodes, suggesting that "math" and "anxiety" are more densely interconnected within their semantic networks, whilst closeness centrality remains largely comparable across models.

The descriptive statistics for the undergraduate students generated by GPT-4o, relative to the three factors and their total math anxiety, are in Table \ref{tab:descriptive_stats_MA}.

\subsubsection{Correlations Analysis}
\begin{figure}[!htb]
    \centering
    \includegraphics[width=1\linewidth]{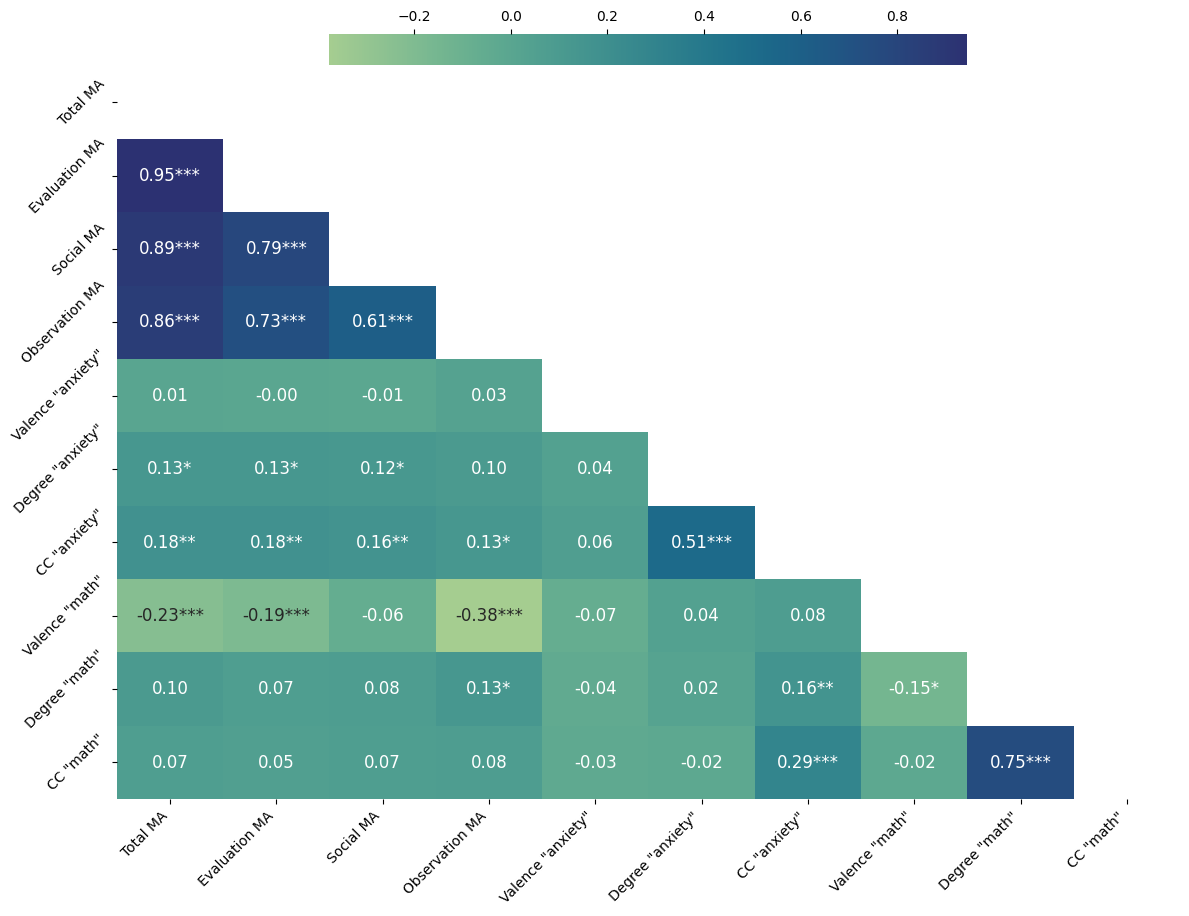}
    \caption{Correlogram Experiment 3 - GPT-4o. The valence, degree and closeness centrality (CC) for the nodes "anxiety" and "math" are reported, along with the factors of MA: Evaluation, Everyday/Social (Social) and Passive Observation (Observation), together with the total MA.}
    \label{fig:correlogram_gpt_4o}
\end{figure}

In figure \ref{fig:correlogram_gpt_4o}, one can find the results of the correlations between the MA factor and BFMNs measure for what concerns GPT-4o generated students. 

Firstly, one can note that the significant correlations found for the GPT-4o simulated students are quite different from those of GPT-3.5. Here, in fact, significant negative correlations are found for the valence assigned to the node of "math", as well as with the degree and closeness centrality of the node "anxiety". 

Specifically, Total ($r = -.23$, $p < .001$), Evaluation ($r = -.19$, $p < .001$), and Observation ($r = -.38$, $p < .001$) MA present significant negative correlations with the valence of "math". Always considering this node, we can find only one significantly positive correlation with the degree of "math" for the Passive Observation MA ($r = .13$, $p < .05$).

For what concerns "anxiety", we found significant correlations between its degree and Total ($r = .13$, $p < .05$), Evaluation ($r = .13$, $p < .05$) and Everyday/Social ($r = .12$, $p < .05$) MA; as well as with its closeness centrality and all the dimensions of MA (Total MA: $r = .18$, $p < .01$; Evaluation MA: $r = .18$, $p < .01$; Everyday/Social: $r = .16$, $p < .01$; Passive Observation: $r = .13$, $p < .05$).

Collectively, these results suggest that GPT-4o’s simulated students with higher math anxiety tended to establish broader associative connections to "anxiety" — making it a more central concept in their semantic networks — while concurrently attributing more negative emotional valence to "math". Unlike GPT-3.5, GPT-4o also reflected an increased integration of "anxiety" in the associative structure, pointing to a richer yet more affectively charged representation of math anxiety.

\subsubsection{Linear Regressions Analysis}
As previously explained for GPT-3.5 simulated students, also for GPT-4o ones, we conducted four independent OLS linear regressions to obtain the predictive power of the network features of "anxiety" and "math" on the three factors of MA and Total MA. All the results can be found in Tables \ref{tab:lin_reg_total_MA} - \ref{tab:lin_reg_obs_MA} (one for Total MA and three for each MA factor), in line "GPT-4o Students".

For GPT-4o, all linear regression models excluded the degree of "math" and included only three predictors (AICs = $-175.069$, $-173.910$, $-102.599$, $-165.770$, respectively). Our findings showed that for the total MA and all its factors, the degree of "anxiety" resulted in a significant predictor together with the valence of "math", with the only exception of Social MA, where only the degree of "anxiety" is significant. As for the model fit, we found a generally low explanatory power, ranging from $R^2 = .020$ to $R^2 = .154$, with Passive Observation MA showing the strongest fit.

Specifically, the Total MA model accounted for 7.2\% of the variance ($R^2 = .072$), with significant contributions from both the degree of "anxiety" ($\beta = .096$, $p < .05$) and the valence of "math" ($\beta = -.099$, $p < .001$). Evaluation MA followed a similar pattern, explaining 5.4\% of the variance ($R^2 = .054$) and showing significance for the degree of "anxiety" ($\beta = .091$, $p < .05$) and the valence of "math" ($\beta = -.082$, $p < .001$). Everyday/Social MA showed a weaker fit ($R^2 = .020$), with only the degree of "anxiety" ($\beta = .096$, $p < .05$) emerging as a significant predictor. The Passive Observation MA model displayed an explanatory power of 15\% ($R^2 = .154$), with significant effects for both the degree of "anxiety" ($\beta = .085$, $p < .05$) and the valence of "math" ($\beta = -.171$, $p < .001$).

\subsection{Discussion}
Across both GPT models, the relationship between math anxiety and associative network features diverged notably from that observed in human participants, in \nameref{Experiment 1} and \nameref{Experiment 2}. For GPT-3.5, higher math anxiety levels corresponded to fewer conceptual associations with "math", suggesting a simplified or avoidant semantic representation of the concept. Conversely, GPT-4o displayed more human-like patterns, where higher math anxiety was associated with greater connectivity of "anxiety" and more negative emotional evaluations of "math". These patterns indicate that GPT-4o developed a richer, more affectively coherent representation of anxiety-related content, whereas GPT-3.5’s associations remained structurally sparse and affectively flat. Despite this qualitative improvement, both models showed minimal explanatory power in the regression analyses—ranging from 1\% to 2.9\% of explained variance (except GPT-4o Passive Observation factor with an explanatory power of 15.4\%)—compared to the substantially higher values observed in human data, which range from 13.1\% to 40.5\%.

This discrepancy highlights a fundamental limitation in using LLMs to model human affective–cognitive constructs. Human math anxiety reflects an interplay between emotional experience, autobiographical memory, and individual differences \citep{stella2022network, lefevre2005mathematical}, whereas GPTs operate solely through linguistic co-occurrence patterns learned from text. As noted by \citet{binz2023using} and \citet{kasneci2023chatgpt}, LLMs can reproduce affective language but lack genuine emotional grounding or self-referential cognition. Consequently, their associative networks mimic surface-level correlations in human discourse rather than the underlying emotional dynamics driving psychometric variability. The improved sensitivity of GPT-4o compared to GPT-3.5 likely reflects enhanced contextual modelling and exposure to more affect-rich data, yet it still falls short of the experiential grounding that characterises human cognitive–emotional representations of math anxiety.

To gain a deeper understanding of the differences between human and simulated students, a group-level analysis could provide further insights, as we will explore in the following \nameref{Experiment 4}. This analysis may reveal additional layers of disparity between human and LLM cognitive patterns, enhancing our understanding of how these models interact with psychometric assessments and contributing to the ongoing discussion on the applicability of LLMs in human-centric evaluations.

\section{Experiment 4} \label{Experiment 4}
\subsection{Introduction}
\nameref{Experiment 3} highlighted differences between human and LLM data. We further explore this aspect by having a closer look at the psychometric scores of real and simulated participants. We also explore BFMNs and their structure, but in an aggregated way, merging BFMNs from students displaying analogously low or high levels of total math anxiety. The main reason for this approach is to better assess, with a more interpretable perspective, how the structure of recalls differed in their nature and valence ratings between humans and LLMs.

To this aim, we rely on literature from cognitive science. Semantic frame theory \citep{fillmore2006frame} posits that meaning attributed to a given concept can be reconstructed by considering its associations. For instance, if "anxiety" is rated positively, it might spawn positive memory recall patterns such as "motivation" or "defence". Reading these connections would inform about the contextual associations semantically framing the target word (i.e. "anxiety"). This means that additional insights on the network structure of "anxiety", "math" and other concepts can be traced by considering their associations. However, the theory states that this process of meaning reconstruction is possible only in the presence of enough contextual knowledge (e.g. associates; \cite{fillmore2006frame}). In individual BFMNs, the degree of math or anxiety totalled only a handful of connections, making it daunting to perform human coding and extrapolate how concepts were framed. This challenge has already been noticed in past investigations with BFMNs and mindset growth \citep{stella2020forma}. Hence, we operationalise semantic frames as network neighbourhoods of target concepts in BFMNs, which presented more connections than individual ones, i.e. BFMNs obtained by aggregating free associations and valence ratings across groups of individuals with similar features. 

We use the human data from \nameref{Experiment 1} and the GPT-simulated data from \nameref{Experiment 3} (collected using the methodology described in the \nameref{Procedure} Section) to build a total of 6 group-level BFMNs, expressing the ways of associating and perceiving ideas for: 
\begin{enumerate}
    \item Human students with total MA scores lower than the median.
    \item GPT-3.5 simulated students with total MA scores lower than the median.
    \item GPT-4o simulated students with total MA scores lower than the median.
    \item Human students with total MA scores higher than the median.
    \item GPT-3.5 simulated students with a total MA higher than the median.
    \item GPT-4o simulated students with higher-than-median total MA.
\end{enumerate} 
We exclude from the comparison simulated or real students with total math anxiety scores equal to the median, which is assumed as a neutrality bound for this experiment.

Considering group-level rather than individual-level BFMNs can lead to significantly richer and more nuanced semantic frames \citep{stella2021mapping, stella2022network}. This is due to group-level BFMNs aggregating together all edge lists and valence rating scores from individuals falling within the same group. We investigate group-level semantic frames for target concepts such as "anxiety" and "math" to better understand the results from the previous experiments.

\subsection{Methods}
\subsubsection{Participants} \label{Participants}
The sample consists of three groups: the group of human students, composed of 72 students from \nameref{Experiment 1}; the students simulated with GPT-3.5 (300 participants) and the ones simulated with GPT-4o (300 participants), the same used for \nameref{Experiment 3}. Additional information can be found in the following Sections: \nameref{Methods} and \nameref{Methods - Experiment 3}.

Students were split into subgroups based on the median of the Total Math Anxiety (obtained by the MAS-IT filling). Participants who presented a level of Total MA equal to the median were excluded from the analysis; students with levels of Total MA lower than the median were placed in the "low anxiety" group, while students with a level of Total MA higher than the median were placed in the "high anxiety" group. In Table \ref{tab:groups_participants}, the reader can find the number of students of each subgroup for the three different groups of participants.
\begin{table}[!htbp]
    \centering
    \begin{tabular}{lccc}
    \toprule
        Group & Low Anxiety & High Anxiety &
        Excluded \\
    \midrule
        Experiment 1 Students & 33 & 33 & 4 \\
        GPT-3.5 Students & 133 & 142 & 25 \\
        GPT-4o Students & 135 & 149 & 16 \\
    \bottomrule
    \end{tabular}
    \caption{Number of students for each anxiety-level subgroup, based on the levels of MA compared to the median of Total MA (i.e, "low anxiety", "high anxiety") and the ones excluded from such division (i.e. with levels of MA equal to the Total MA).}
    \label{tab:groups_participants}
\end{table}

\subsubsection{Network construction}
We built group-level BFMNs by aggregating the edge lists of all individuals for each group. Aggregation occurred in a way that preserved the repetition of individual associations, leading to a weighted BFMN where nodes (links) indicate all concepts (associations) ever read or recalled (performed) by at least one individual in the group. In the group-level BFMN, links are undirected and weighted by the frequency of repetition of the same associations across individuals (e.g. if 3 participants made the association "math" - "boring", then the edge would be weighted 3). 

Similarly, valence ratings for each concept were combined across all individuals in the group. This means that instead of assigning a single rating to each concept, we collected the ratings of different participants. As a result, each concept in the network has a list of emotional ratings (on a scale from 1 to 5), allowing for richer group-level analysis.

Valence labels (i.e. 1 for positive; 0 for neutral and -1 for negative) were assigned by verifying whether lists of scores assigned by participants in a given group to a specific concept were lower, equivalent, or higher compared to all scores provided by the same group for all remaining concepts, as performed in past investigations with BFMNs \citep{stella2019viability, stella2021mapping, abramski2023cognitive}. This comparison was operationalised as a non-parametric test (using the Kruskal-Wallis test) comparing the list of scores attributed by a given group to a target concept with the list of scores attributed by the same group to all other concepts. For instance, students might assign the following scores \{4, 5, 3, 5, 6, 10\} to "science". This set would then be compared to the set \{3, 4, 1, 3, 5, 4, ..., 4\} of all other scores assigned to all other concepts, such as "math", "physics", and so on, via a two-tailed Kruskal-Wallis test producing a test statistic and a $p$-value. Given the relatively small sample size of the human group, we fixed a significance level of $p < .1$ for this study, compatible with past relevant experiments \citep{stella2019viability}.

Since the Kruskal-Wallis test cannot be applied to samples with fewer than three elements, words with fewer than three scores assigned to them were considered neutral. For all the other words: if their mean score was lower (higher) than the mean scores of all the other concepts, and the Kruskal-Wallis test resulted significant ($p < .1$), a negative (positive) valence label (i.e. value -1 (1)) was assigned to such concept. On the other hand, concepts for which no statistically significant difference was found ($p > .1$) were attributed a neutral valence label (i.e. value of 0). The neutral valence label was also assigned if the test was indeed significant, but the two mean scores (i.e. the mean of the concept itself and the mean of all the other concepts) were equal. 

\subsubsection{Semantic frames}
Group-level BFMNs were built for all the subgroups (see \nameref{Participants} Section) independently. Then, semantic frames for key target concepts were identified as the network neighbourhoods of those concepts (i.e, all associations attributed to the target). Semantic frames were visualised in Python using code from the EmoAtlas package \citep{semeraro2024emoatlas}. The library uses hierarchical edge bundling to visualise nodes on a circular embedding, placing the nodes more interconnected with each other closer on the circle. Following the same colour scheme of past studies \citep{stella2021mapping}, we visualised negatively (positively, neutrally) perceived words (i.e. concepts with negative (positive, neutral) valence ratings) in blue (cyan, black). Font size was increased for words with a higher closeness centrality in the semantic frame \citep{semeraro2024emoatlas}. We also highlighted associations involving a positive (negative, neutral) concept in cyan (red, black). Contrastive associations linking positive and negative concepts were highlighted in purple. For each semantic frame, we counted the occurrence of positive, negative and contrastive associations to identify the mode of the valence attributed to a given semantic frame. Thanks to the interplay of connectivity and valence labels in BFMNs, it is possible for positive concepts to be framed in negative frames (i.e. semantic frames in which most of the associations are perceived negatively).

Last but not least, we computed the local clustering coefficient of each semantic frame, i.e. the fraction of links between neighbours of a target node \citep{newman2018networks}. The local clustering coefficient ranges from 0, meaning that the semantic frame features only links between the target concept and its associations, to 1, where the semantic frame features associations that are all linked with each other and with the target concept.

\subsection{Results}
In this Section, we will report the semantic frame of "math" and "anxiety", which are the central concepts of our study, but also the ones of related words found to be significant for the treated topic. Specifically, we will explore the semantic frames of two other STEM subjects: "science" and "statistics", which are expected to be perceived quite differently, and the concept of "therapist", since it can be considered a positive key actor in the management/overcoming of math anxiety.

Thanks to our participant subgroups, we will be able to investigate the differences between real and simulated students, as well as participants with high or low anxiety levels. We will consider the valence of the semantic frames perceived by the subgroups (i.e. the valence mode of the semantic frame) and its semantic content.
\paragraph{Math}
\begin{figure}[!htbp]
\centering
\begin{subfigure}[t]{0.48\textwidth}
    \includegraphics[width=\textwidth]{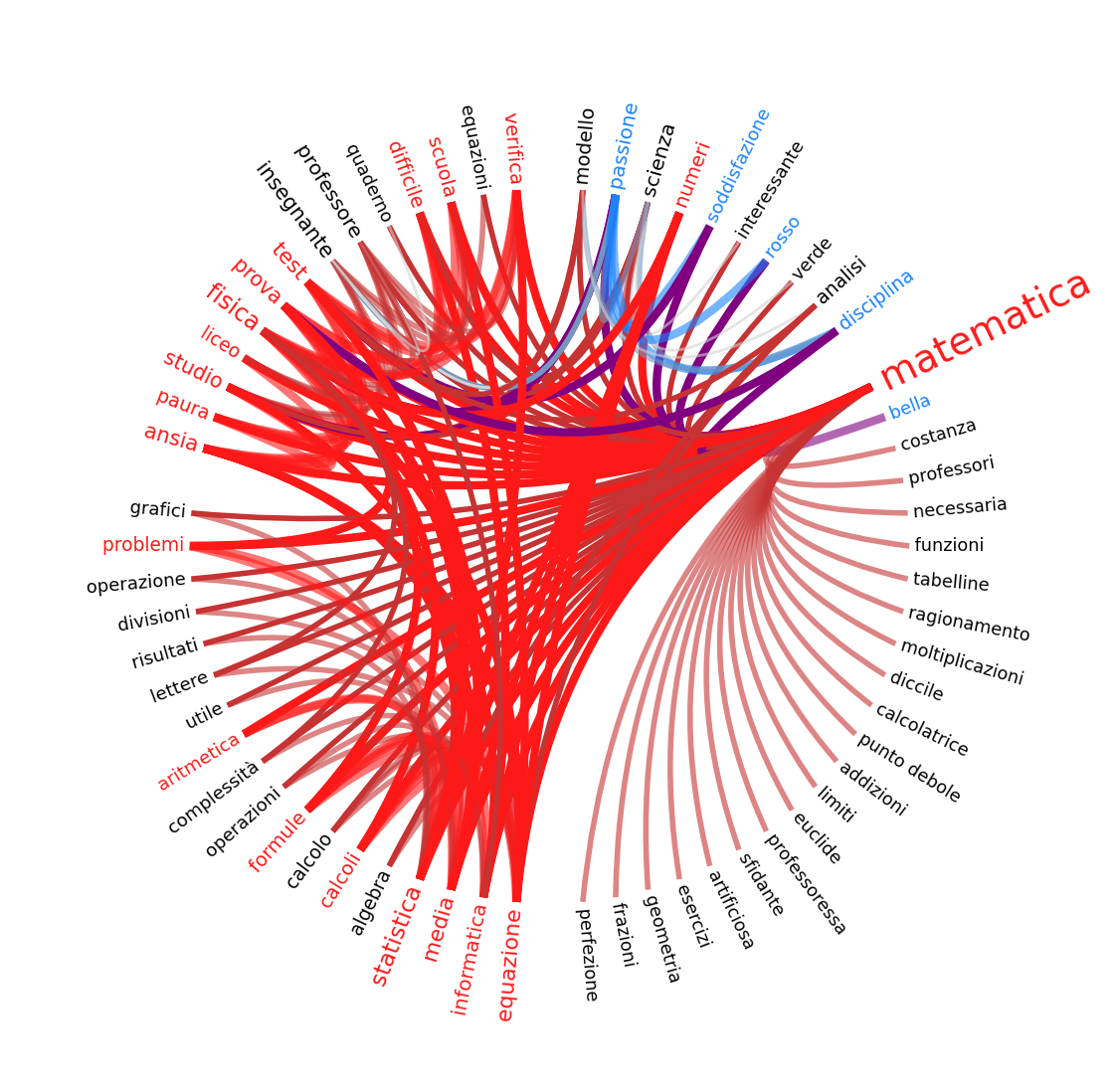}
    \caption{High anxiety human students}
    \label{high_anx_students_math}
\end{subfigure}
\begin{subfigure}[t]{0.48\textwidth}
    \includegraphics[width=\textwidth]{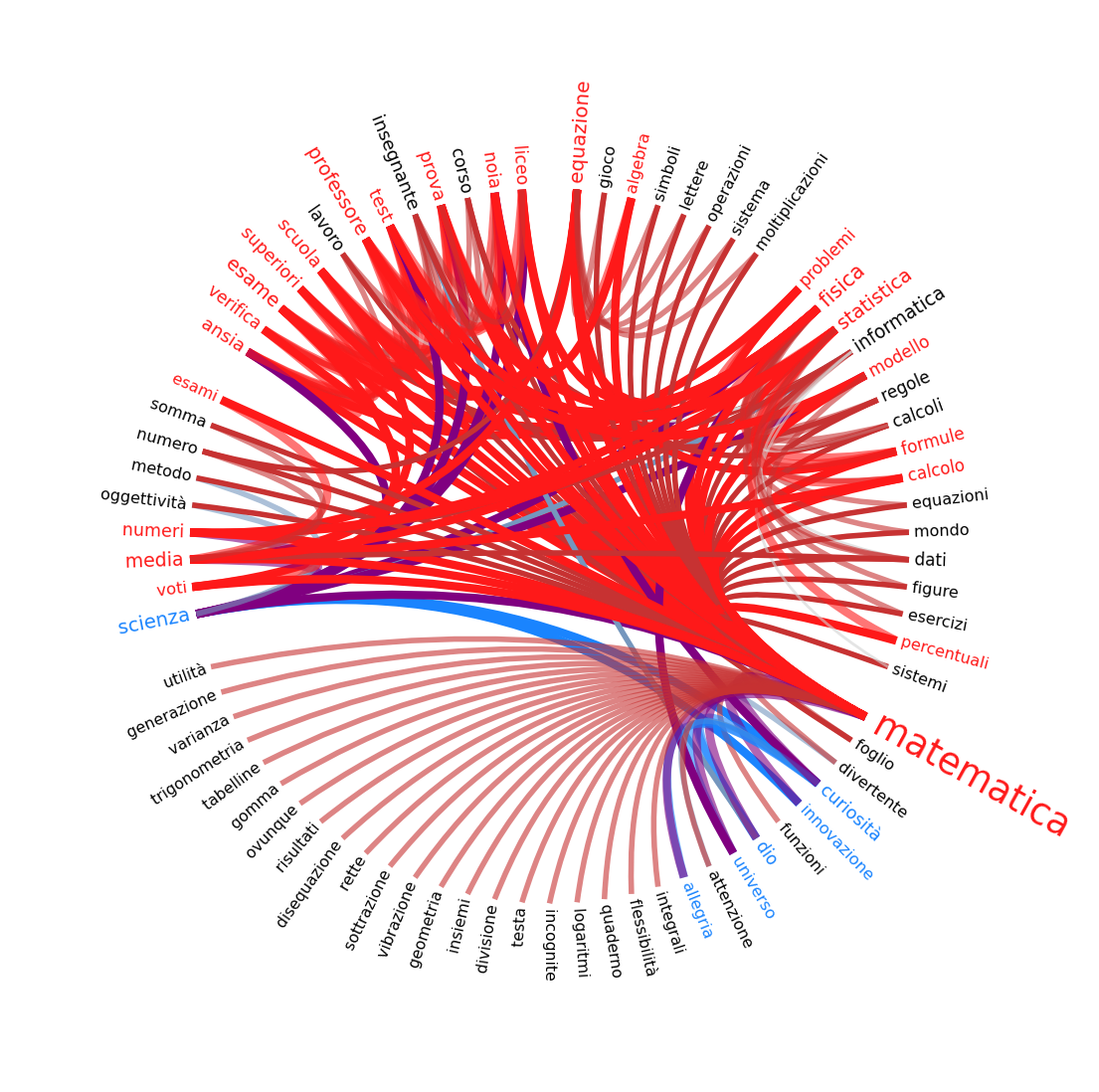}
    \caption{Low anxiety human students}
    \label{low_anx_students_math}
\end{subfigure}
\begin{subfigure}[t]{0.44\textwidth}
    \includegraphics[width=\textwidth]{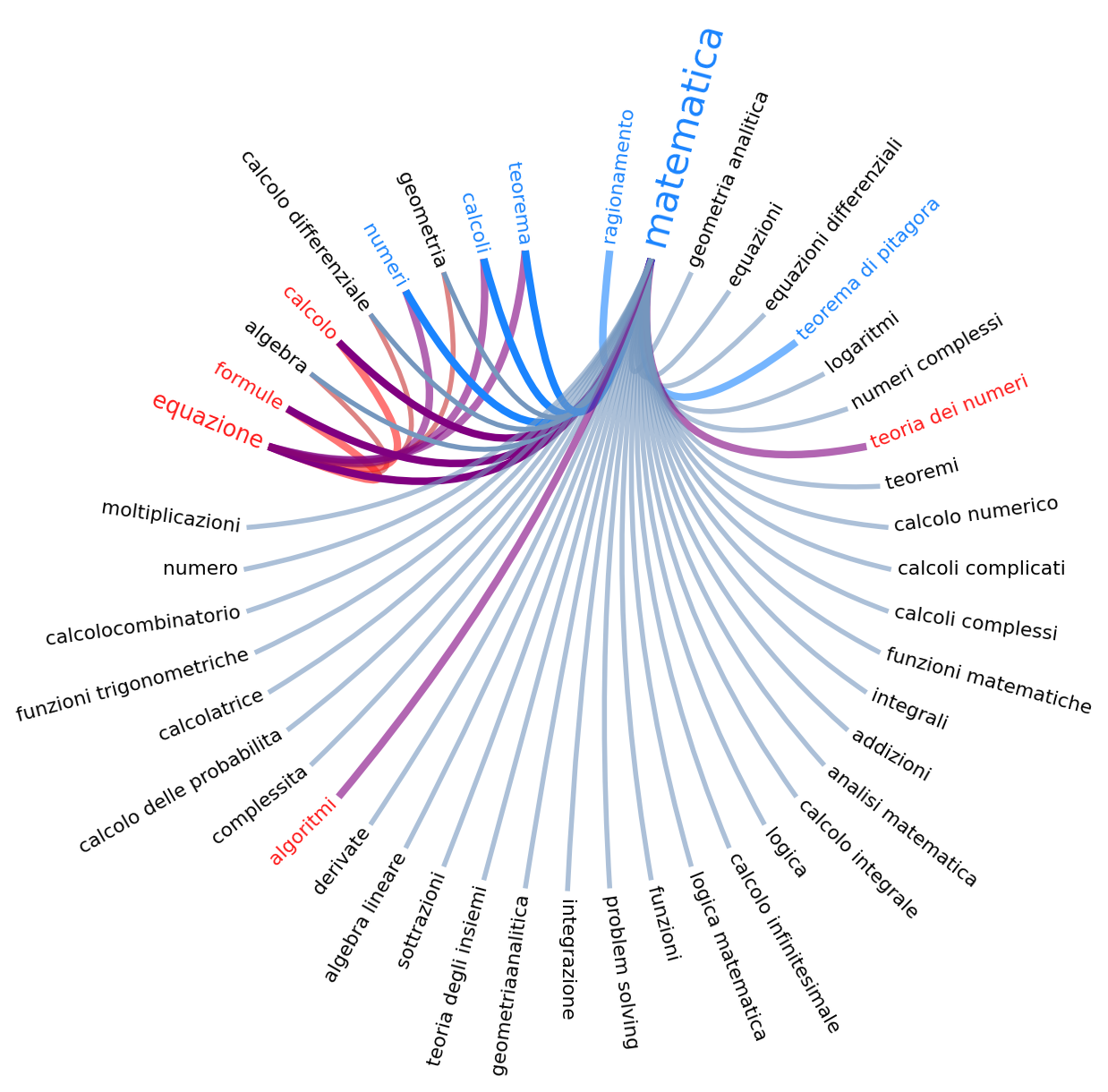}
    \caption{High anxiety GPT-3.5 students}
    \label{high_anx_GPT3.5_math}
\end{subfigure}
\begin{subfigure}[t]{0.46\textwidth}
    \includegraphics[width=\textwidth]{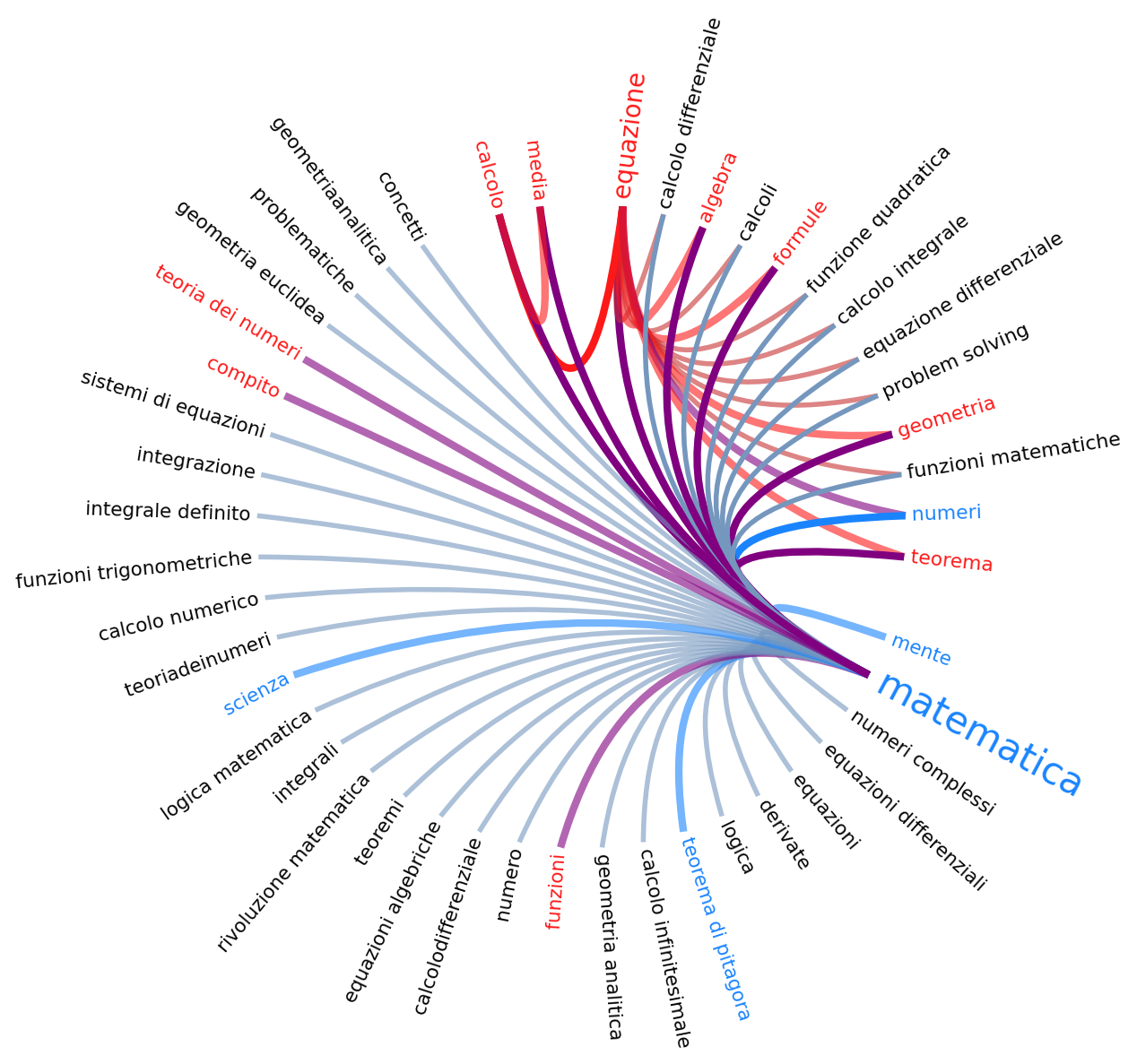}
    \caption{Low anxiety GPT-3.5 students}
    \label{low_anx_GPT3.5_math}
\end{subfigure}
\begin{subfigure}[t]{0.43\textwidth}
    \includegraphics[width=\textwidth]{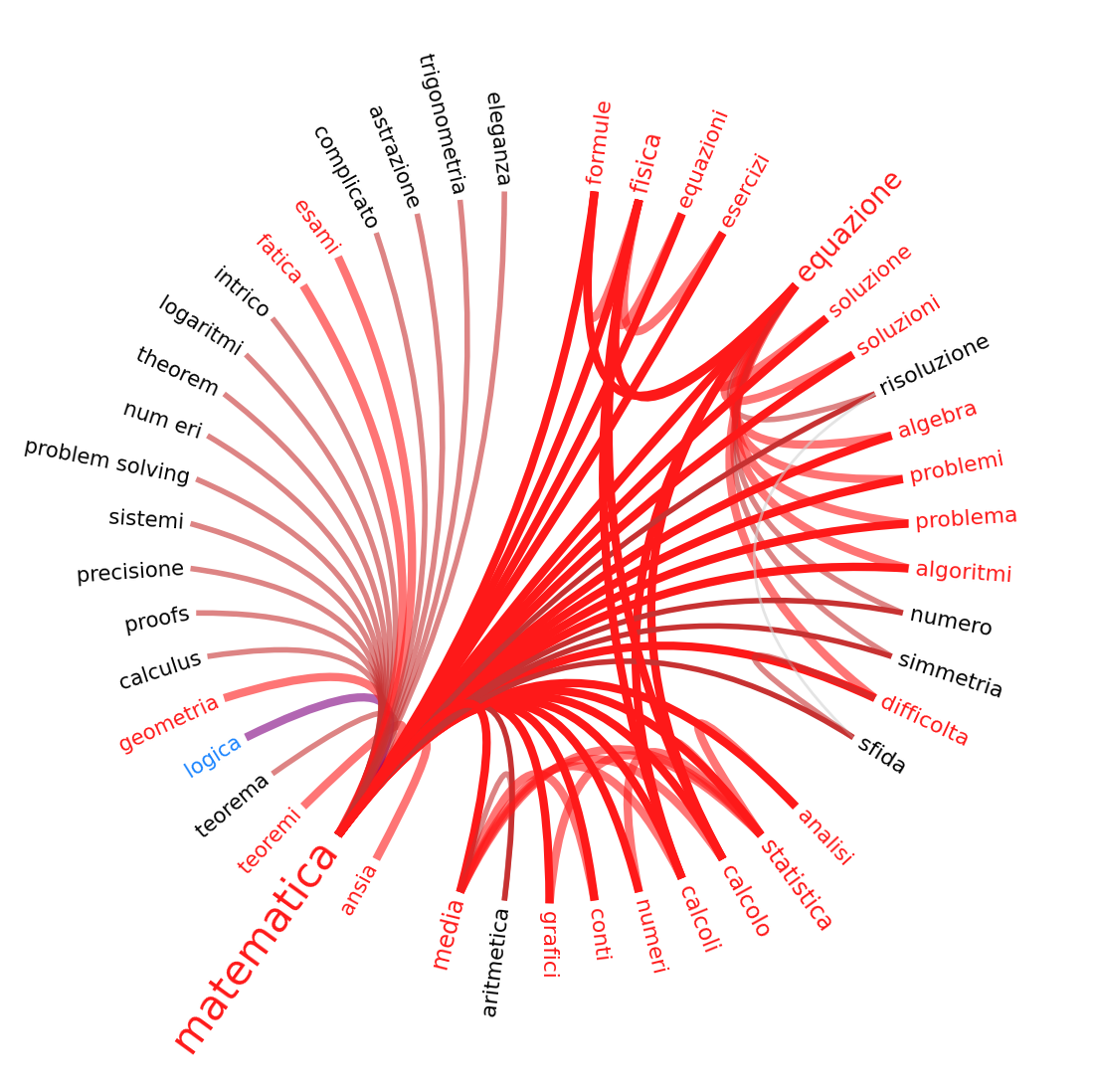}
    \caption{High anxiety GPT-4o students}
    \label{high_anx_gpt-4o_math}
\end{subfigure}
\begin{subfigure}[t]{0.43\textwidth}
    \includegraphics[width=\textwidth]{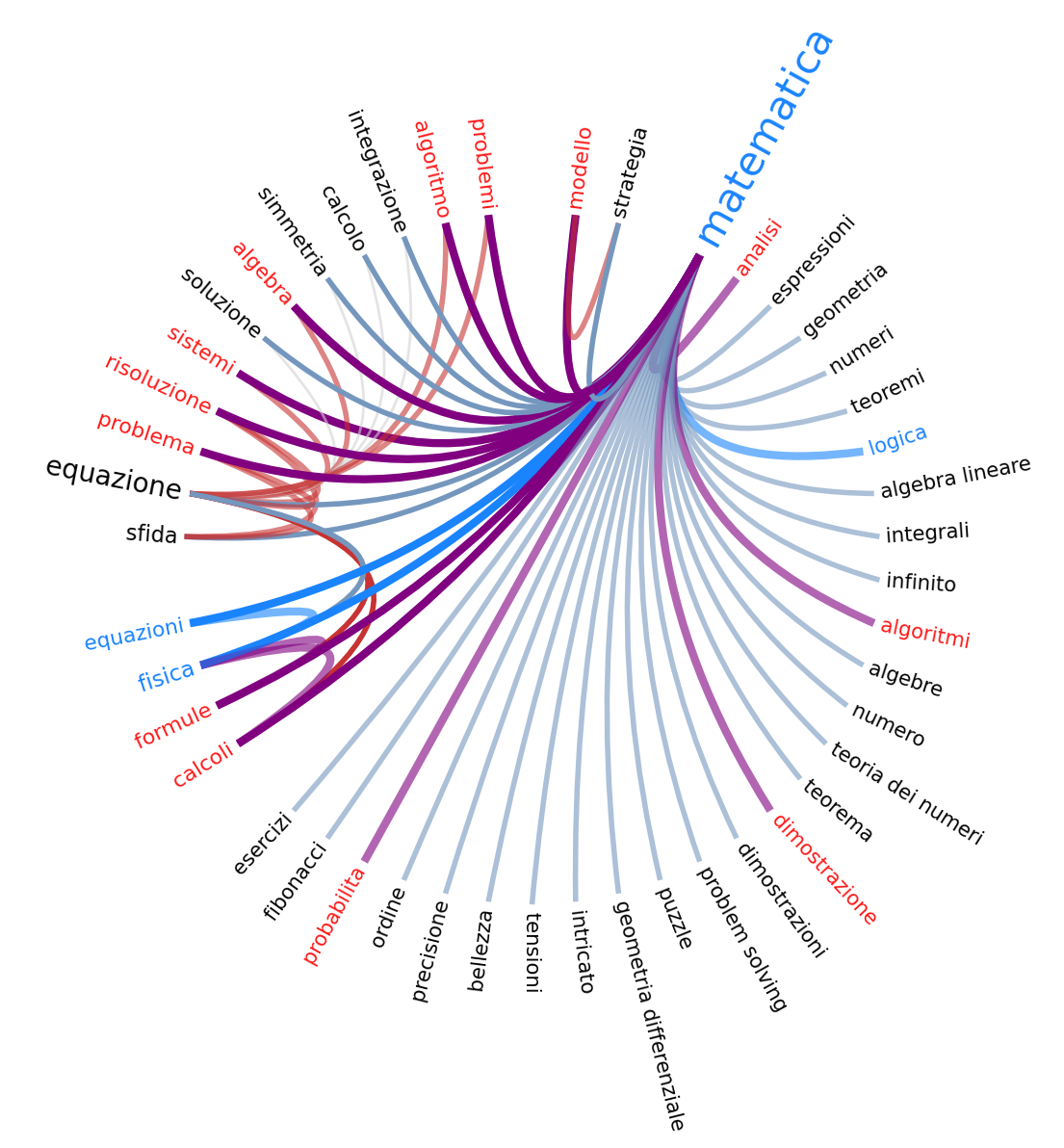}
    \caption{Low anxiety GPT-4o students}
    \label{low_anx_gpt-4o_math}
\end{subfigure}
    \caption{Semantic frames for the node "mathematics".}
    \label{fig:frame_maths}
\end{figure}

In Figure \ref{fig:frame_maths}, we can observe the semantic frames for the concept of "math", for each subgroup. One can notice that, independently of their math anxiety level, for human students (\ref{high_anx_students_math}; \ref{low_anx_students_math}) "math" is perceived as a negative concept. For both anxiety subgroups, "math" is surrounded mainly by negative or neutral words. The semantic content of these associations reveals that students recalled "math" mostly with other negative ideas relative to domain knowledge (e.g. "equations", "formulas"), but also to tests (e.g. "exam", "grade") and other disciplines (e.g. "statistics", "informatics"). Interestingly, this same pattern can be found in GPT-4o students, but only in the high-anxiety subgroup (\ref{high_anx_gpt-4o_math}). Here too, "math" results in a negative concept and the negatively perceived associations are related to specific math terms (e.g. "equations"), other disciplines (e.g. "physics"), and tests (e.g. "exam").

Differently, for what concerns both subgroups of GPT-3.5 students (\ref{low_anx_GPT3.5_math}, \ref{high_anx_GPT3.5_math}) and low-anxiety GPT-4o students (\ref{low_anx_gpt-4o_math}), "math" is considered a positive concept associated with mainly neutral words. These words were considerably more related to advanced domain knowledge than in the human samples (e.g. "differential equations", "number theory", "differential geometry"), which are absent in the subgroups of psychology students. Finally, even if the concept of "math" is considered positive, these simulated student subgroups still produced the same negative associations found in humans (e.g. "equation", "problem").

Comparing human students and GPT-generated students, we can notice that the former produced semantic frames with higher complexity. This can be observed in the representation of the semantic frame (Figure \ref{fig:frame_maths}) and confirmed by the statistical results of the local clustering coefficient ($C$) and the degree ($k$). In fact, the semantic frame of the high-anxiety human students was 7 times more clustered ($C = .056$) compared to its GPT-3.5 ($C = .008$) high-anxiety group, and almost twice as clustered as its GPT-4o ($C = 0.030$) counterpart. Similarly, the degree of "mathematics" in high-anxiety humans ($k = 63$) was higher than the degree for both GPT-3.5 and GPT-4o subgroups ($k = 45$ for both models). These measures indicate a richer and more complex conceptual representation in the high-anxiety subgroup of human students ($N = 33$) compared to the larger subgroups of simulated digital twins ($N = 142$ for GPT-3.5 and $N = 149$ for GPT-4o).

These findings indicate that: (i) independently of math anxiety level, most of the human sample exhibited a negative perception of "math"; (ii) simulated students did not mirror human perceptions: all GPT-3.5 simulated students framed "math" positively, while GPT-4o simulated students perceived it negatively or positively depending on their math anxiety level; (iii) GPT students showed lower complexity compared to human students.

\begin{figure}[!htbp]
\centering
\begin{subfigure}[b]{0.48\textwidth}
    \includegraphics[width=\textwidth]{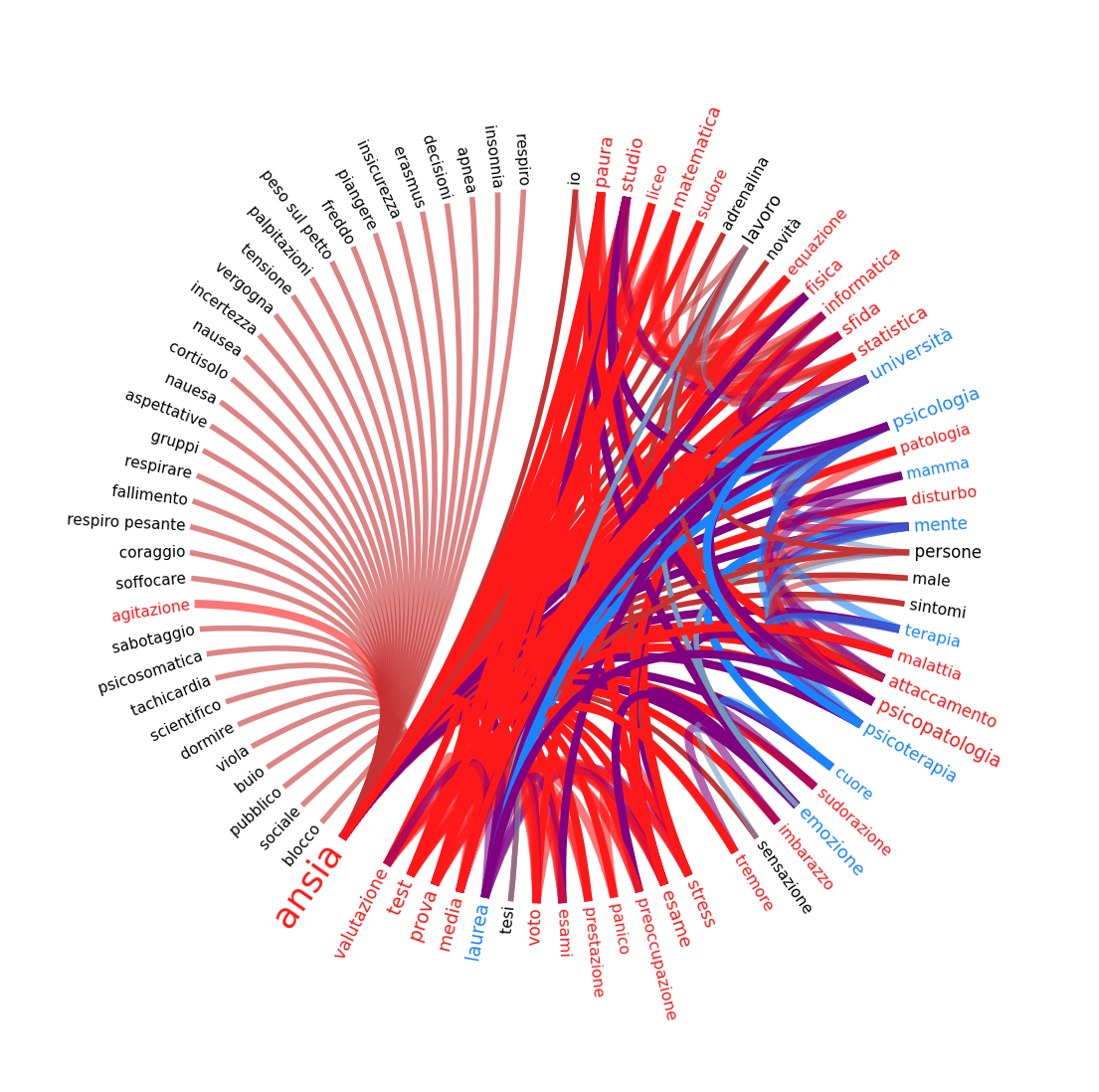}
    \caption{High anxiety students}
    \label{high_anx_students_anxiety}
\end{subfigure}
\begin{subfigure}[b]{0.48\textwidth}
    \includegraphics[width=\textwidth]{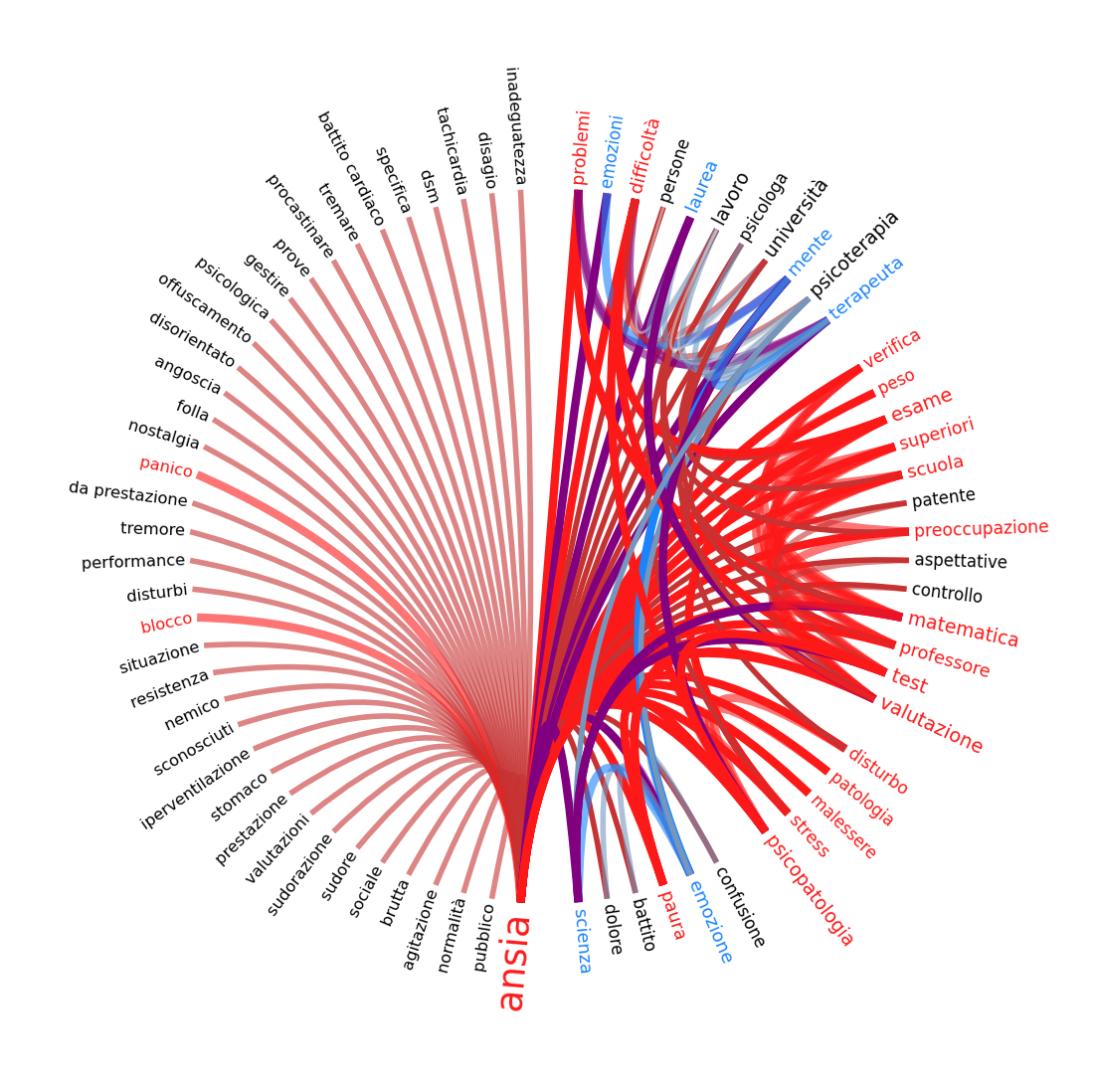}
    \caption{Low anxiety students}
    \label{low_anx_students_anxiety}
\end{subfigure}
\begin{subfigure}[b]{0.46\textwidth}
    \includegraphics[width=\textwidth]{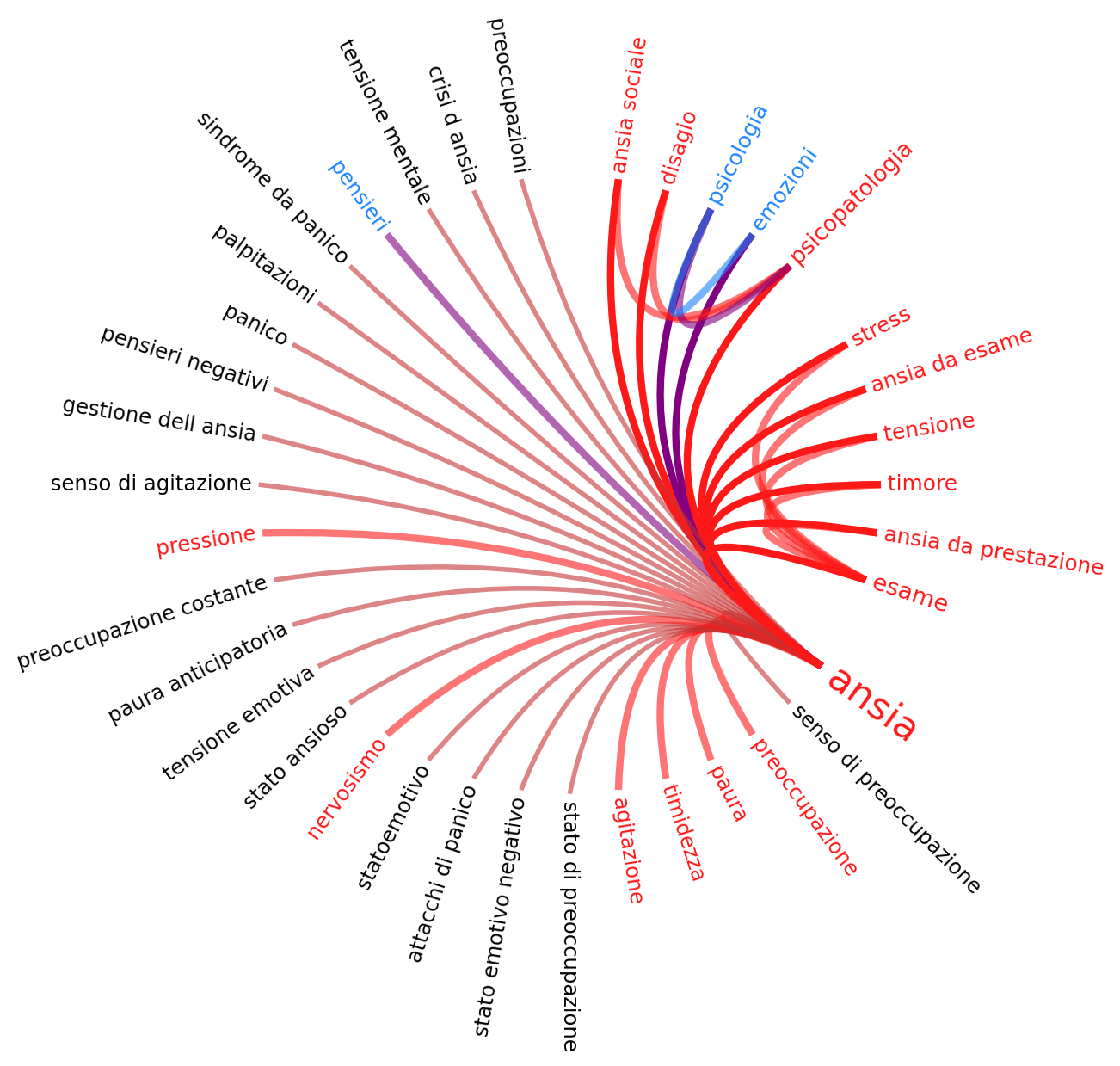}
    \caption{High anxiety GPT-3.5 students}
    \label{high_anx_GPT3.5_anxiety}
\end{subfigure}
\begin{subfigure}[b]{0.44\textwidth}
    \includegraphics[width=\textwidth]{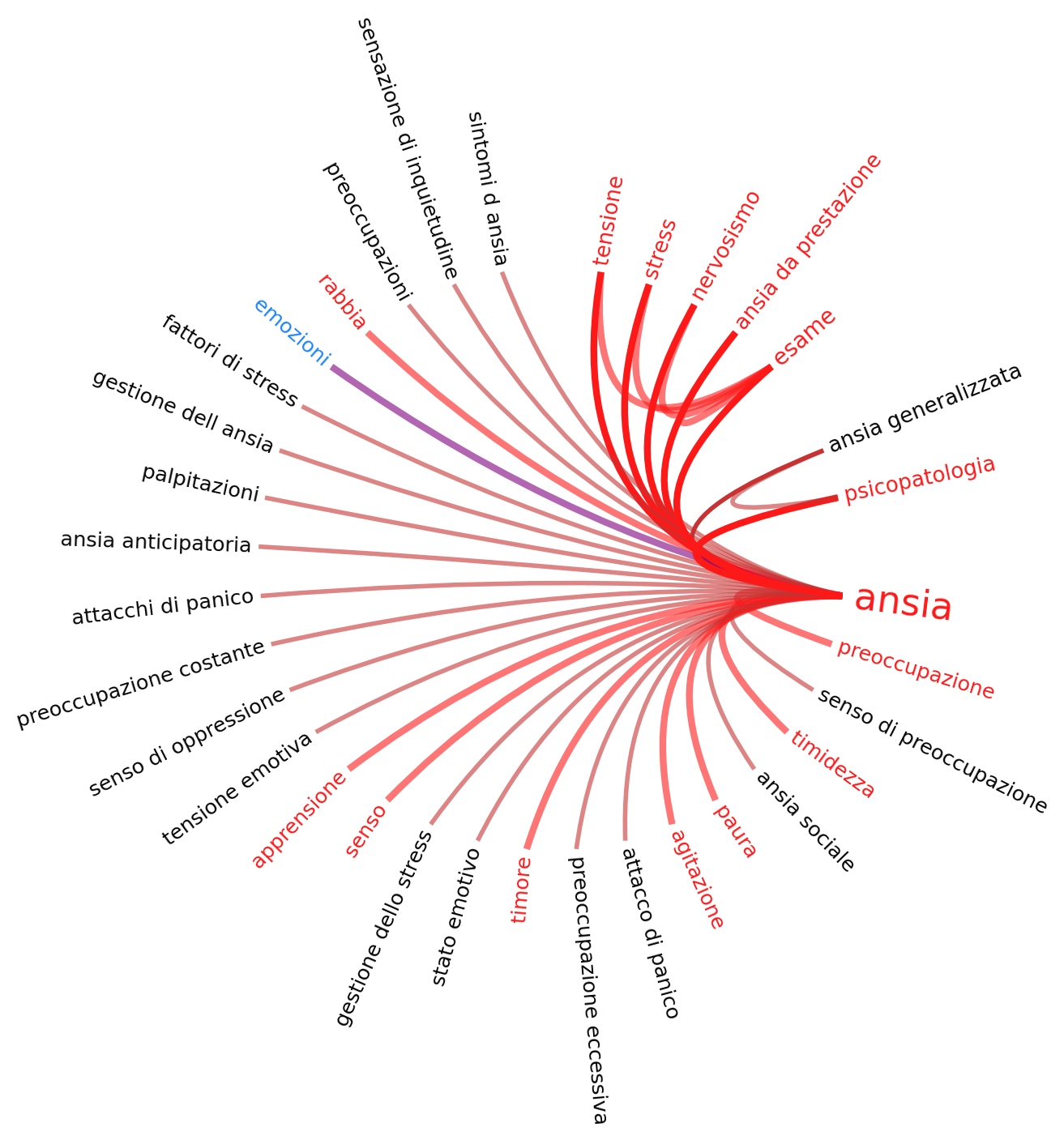}
    \caption{Low anxiety GPT-3.5 students}
    \label{low_anx_GPT-3.5_anxiety}
\end{subfigure}
\begin{subfigure}[b]{0.42\textwidth}
    \includegraphics[width=\textwidth]{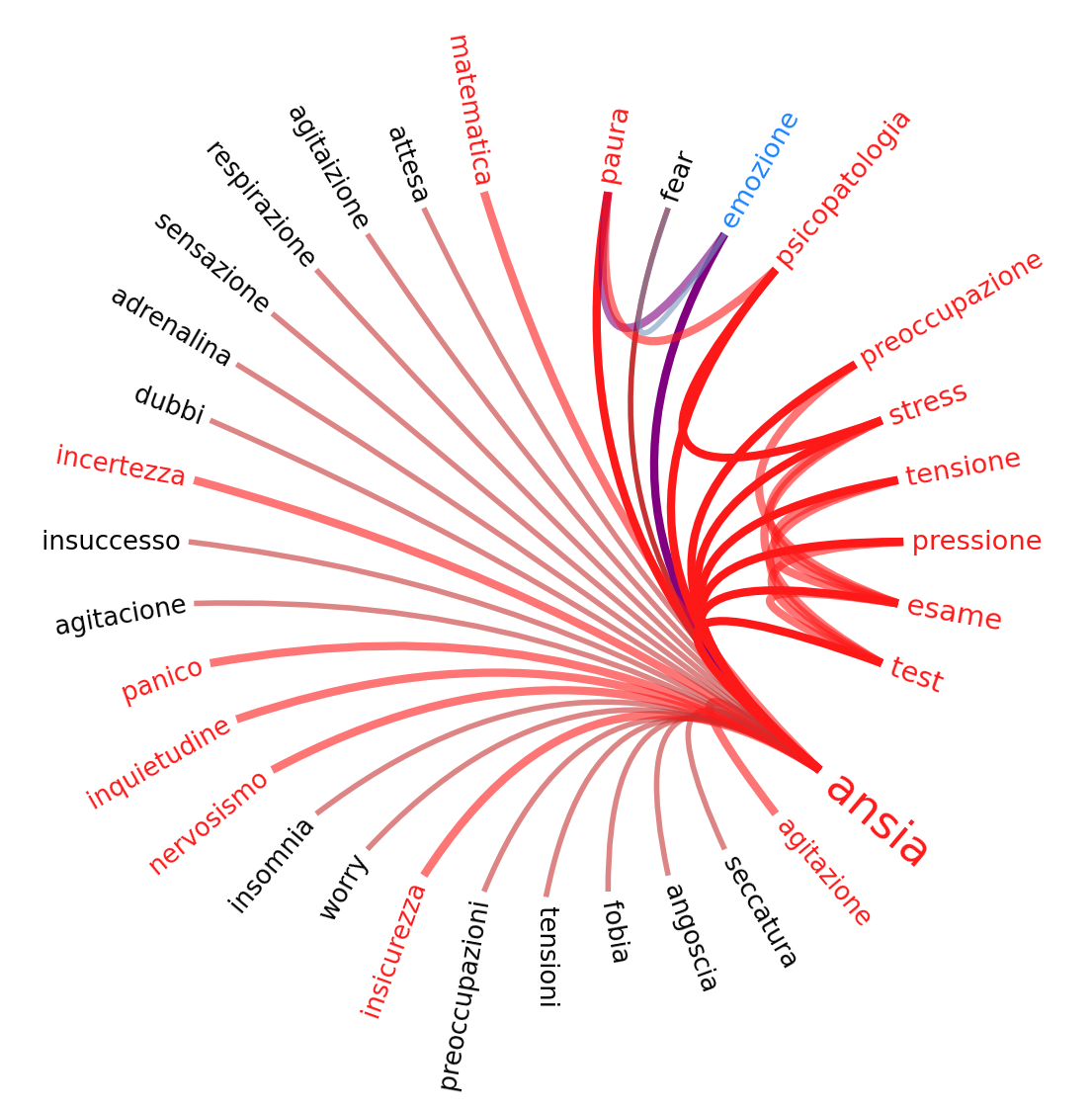}
    \caption{High anxiety GPT-4o students}
    \label{high_anx_gpt-4o_anxiety}
\end{subfigure}
\begin{subfigure}[b]{0.42\textwidth}
    \includegraphics[width=\textwidth]{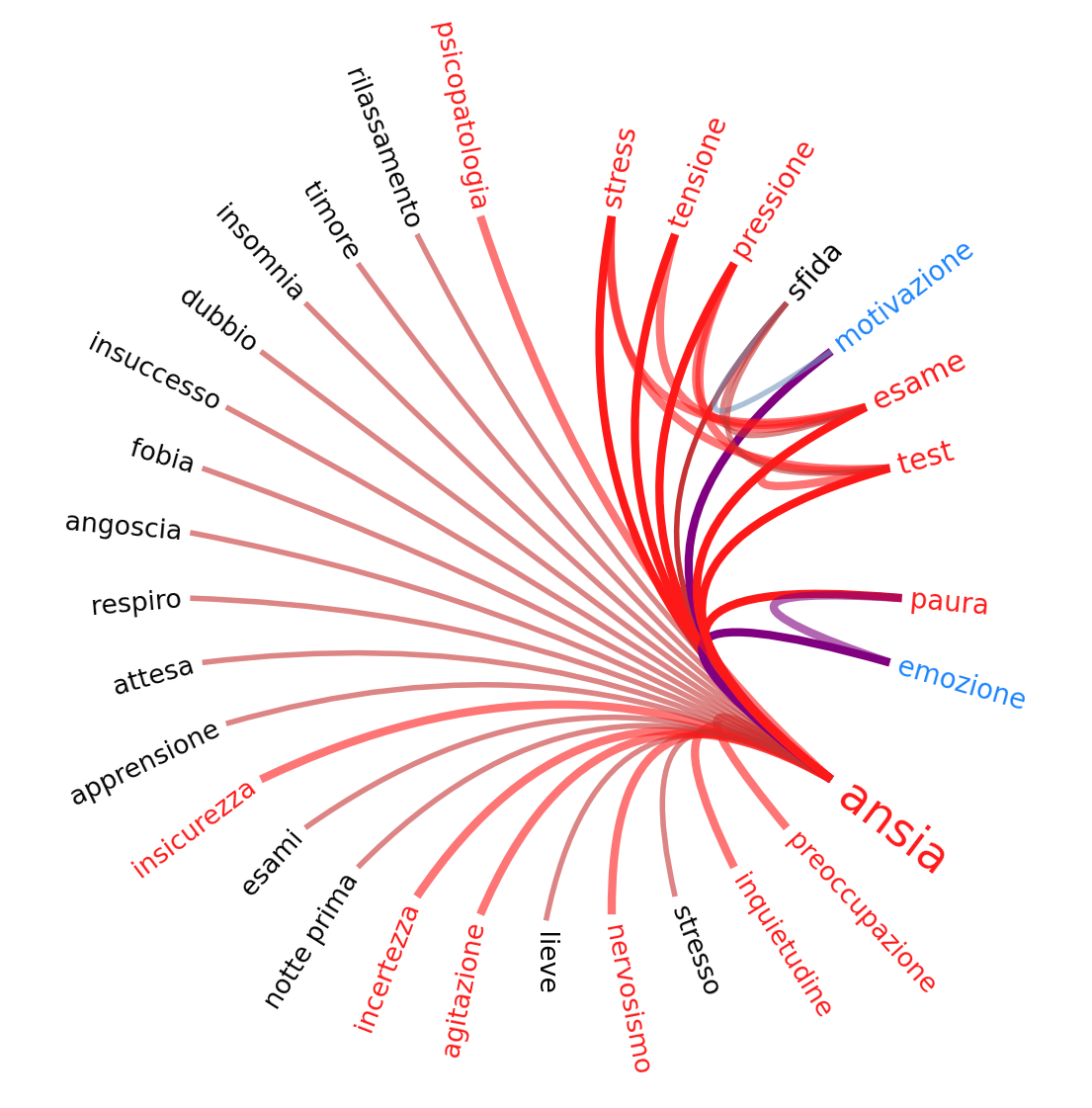}
    \caption{Low anxiety GPT-4o students}
    \label{low_anx_gpt-4o_anxiety}
\end{subfigure}
    \caption{Semantic frames for the node "anxiety".}
    \label{fig:frame_anxiety}
\end{figure}

\paragraph{Anxiety}
In the semantic frames for the concept of "anxiety" (Figure \ref{fig:frame_anxiety}), the difference in valence perception between humans and LLM-generated students found for the concept of "mathematics" is not present. Here, all the subgroups identify "anxiety" as a negative concept, surrounded mostly by negative or neutral associates. However, considering human individuals, we found that "anxiety" was framed along 31 negative associates in the high-anxiety group (\ref{low_anx_students_anxiety}) and with 21 negative associates in the low-anxiety group (\ref{high_anx_students_anxiety}). This difference, together with the presence of more negative connections among negative concepts (i.e. not only in "anxiety") in the high-anxiety case, both suggest that "anxiety" is framed in different ways across higher and lower levels of Total MA.

Reading the semantic content of the negative associations reveals that in the mindset of high-anxiety human students, "anxiety" is strongly tied with concepts relative to evaluations (e.g. "exams", "test") or evoking worry (e.g. "fear", "panic",  "embarrassment"). Some of these associations are also persistent in the low-anxiety group, but the associates themselves are less clustered with each other. In comparison, GPT-3.5 students identify "anxiety" along vague emotional contexts, mentioning more of its symptoms (e.g. "tension", "nervousness", "uncertainty", "pressure"), rather than the same math-related situational associations provided by human participants. Contrarily, in GPT-4o simulated students, other than words linked with anxiety causes/symptoms (e.g. "panic", "fear"), we can find words related to the academic context (e.g. "test", "exams") and, in the high-anxiety group, also the node "mathematics". This supports the finding of a negatively perceived semantic frame of "mathematics" in the GPT-4o high-anxiety subgroup, differently from other GPT subgroups. 

In alignment with the semantic frame of "math", also in the "anxiety" one, the local cluster coefficients indicate higher clustering in the human students, compared to the GPT ones. 
Always considering the high-anxiety subgroups, we report that the semantic frame of human students was more clustered ($C = .042$), and more than twice richer in semantic associations ($k = 81$), compared to its GPT-3.5 ($C = .014$; $k = 36$) and GPT-4o ($C = 0.024$, $k = 32$) counterparts. These differences indicate: (i) the persistence of structural differences also at the group level for "anxiety"; (ii) a simpler structure of semantic frames for "anxiety" in GPT students compared to humans, aligning with the finding for the concept "math".

\paragraph{Science}
Past studies showed that high school students’ BFMNs displayed traces of cognitive dissonance between "science" and its founding blocks, like math \citep{stella2019viability, stella2021mapping}. These traces consisted of students perceiving "science" as a positive entity, but linked in a contrastive way with "math", which was perceived as negative, despite its essential role in scientific inquiry. For these reasons, we also investigated the semantic frame of "science" in the current study, exploring how both humans and LLM students frame this concept.
\begin{figure}[!htbp]
\centering
\begin{subfigure}[p]{0.48\textwidth}
    \includegraphics[width=\textwidth]{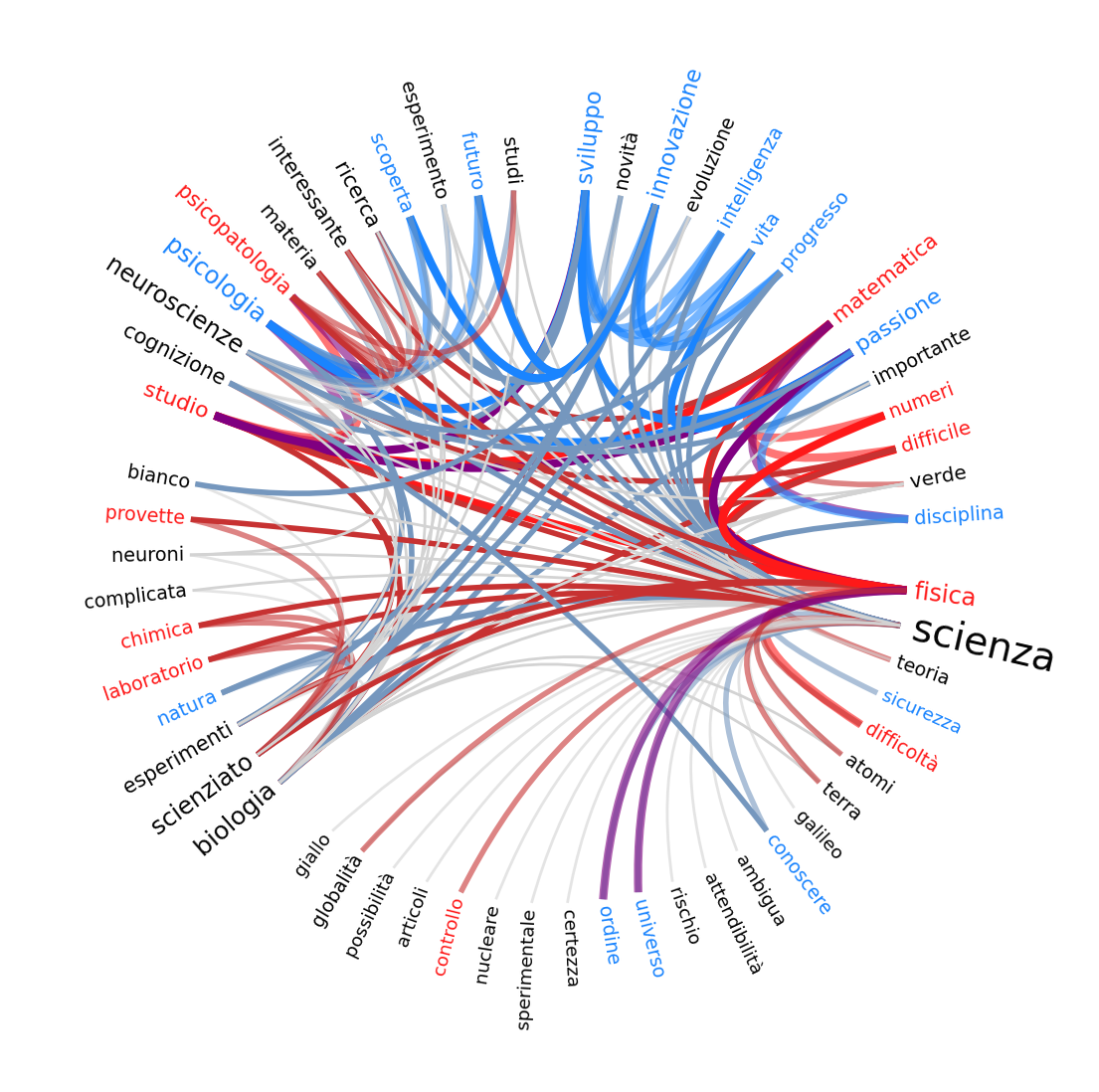}
    \caption{High anxiety students}
    \label{high_anx_students_sci}
\end{subfigure}
\begin{subfigure}[p]{0.48\textwidth}
    \includegraphics[width=\textwidth]{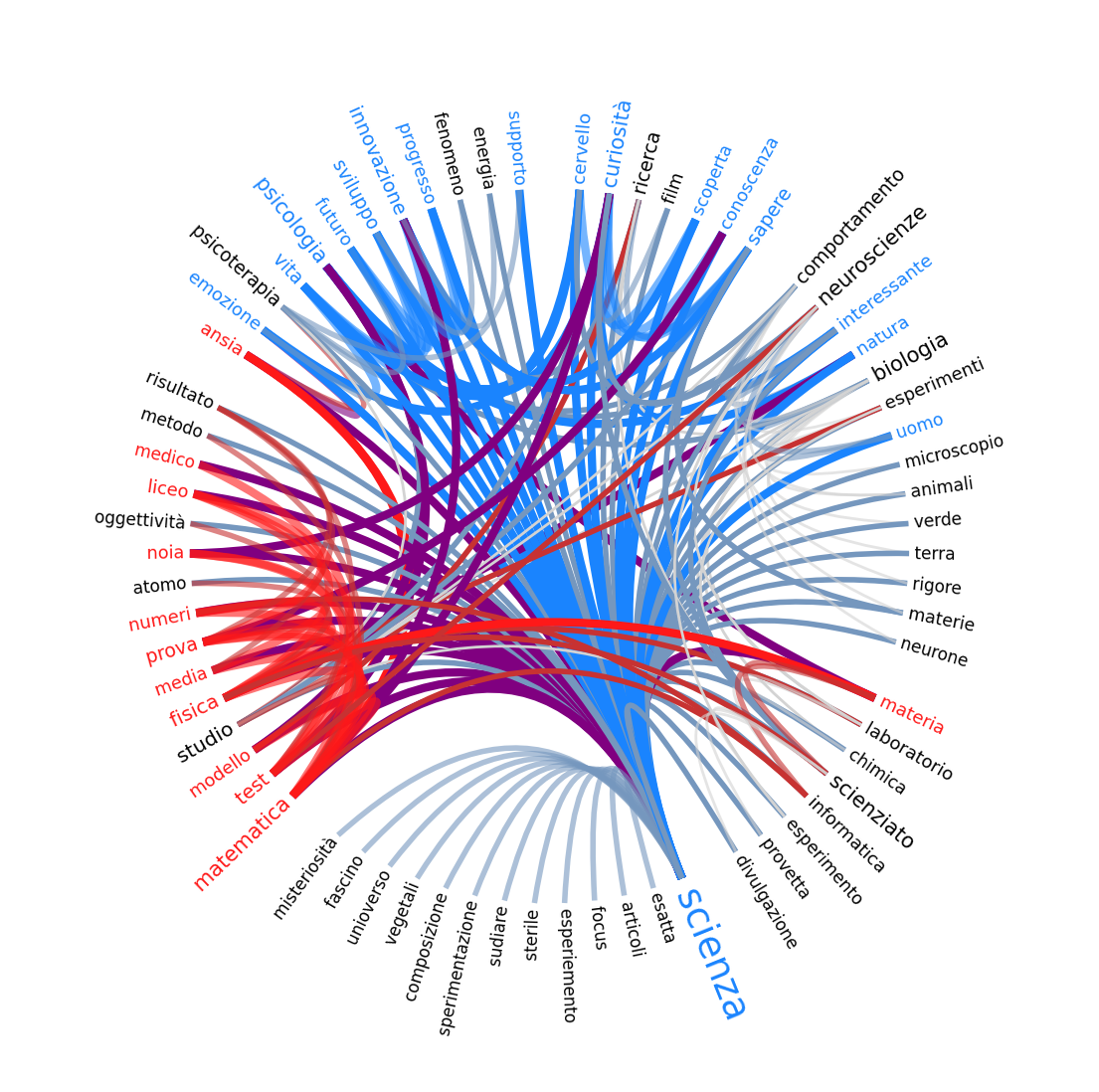}
    \caption{Low anxiety students}
    \label{low_anx_students_sci}
\end{subfigure}
\begin{subfigure}[p]{0.44\textwidth}
    \includegraphics[width=\textwidth]{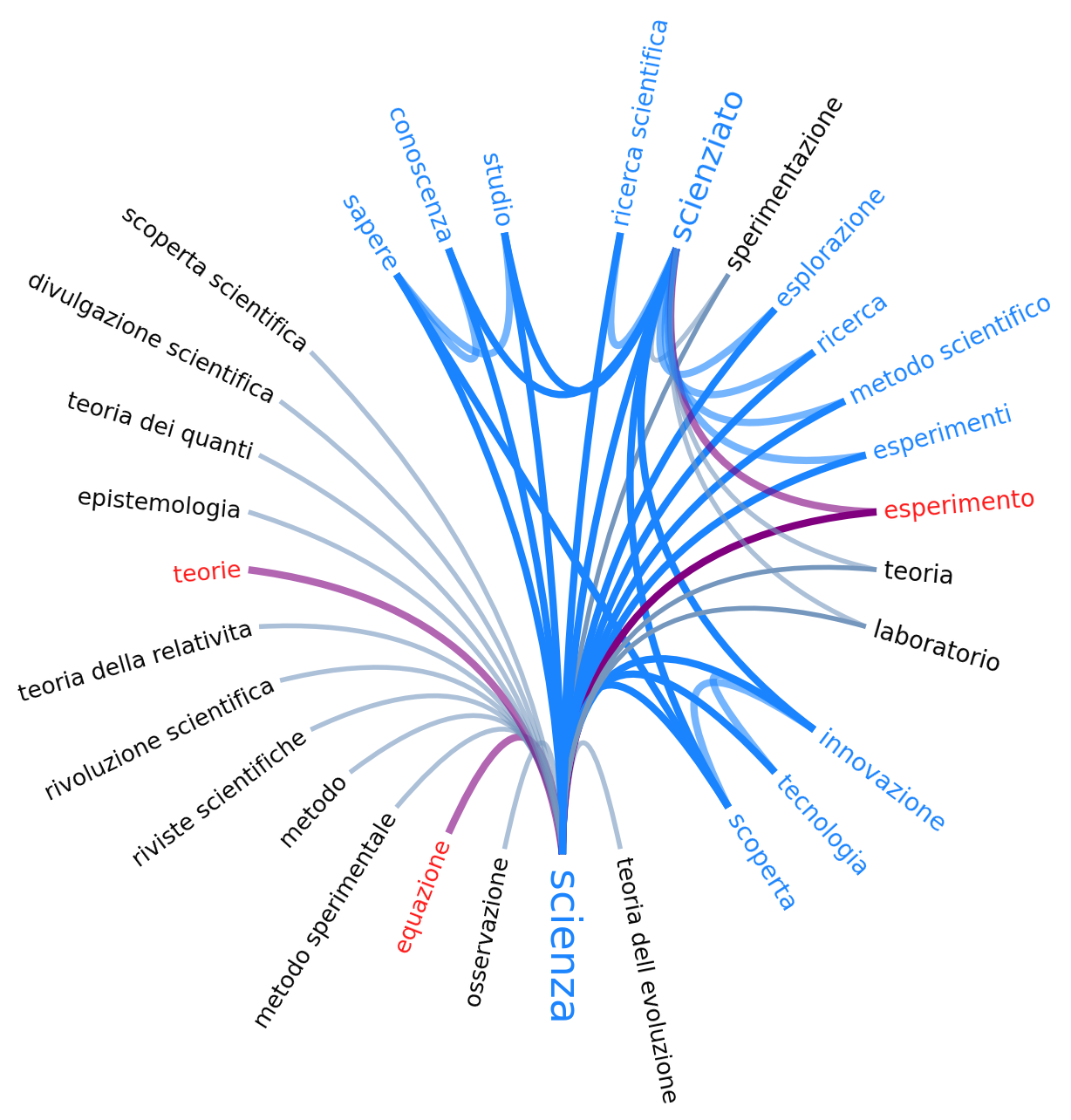}
    \caption{High anxiety GPT-3.5 students}
    \label{high_anx_GPT3.5_sci}
\end{subfigure}
\begin{subfigure}[p]{0.46\textwidth}
    \includegraphics[width=\textwidth]{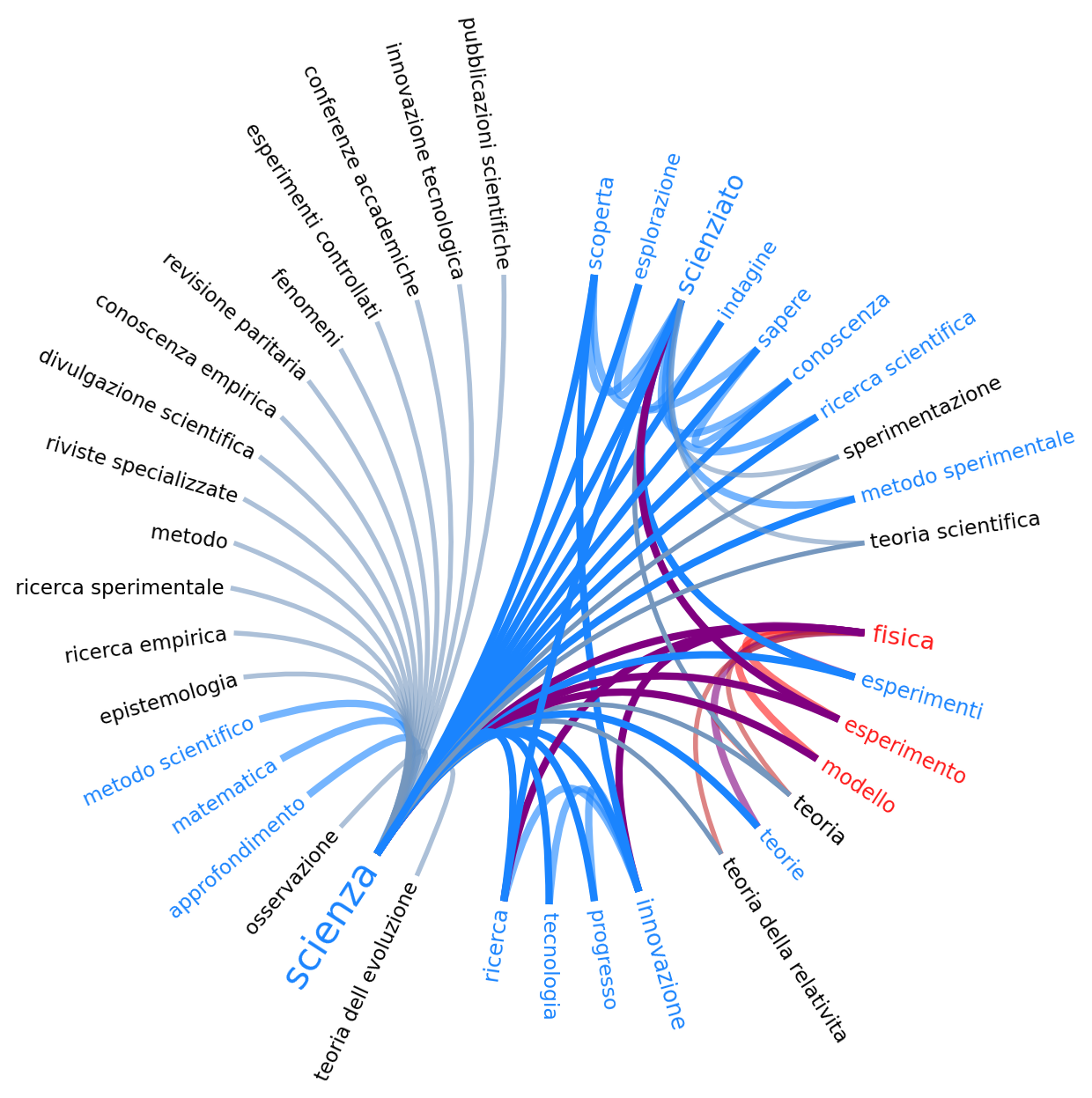}
    \caption{Low anxiety GPT-3.5 students}
    \label{low_anx_GPT3.5_sci}
\end{subfigure}
\begin{subfigure}[p]{0.43\textwidth}
    \includegraphics[width=\textwidth]{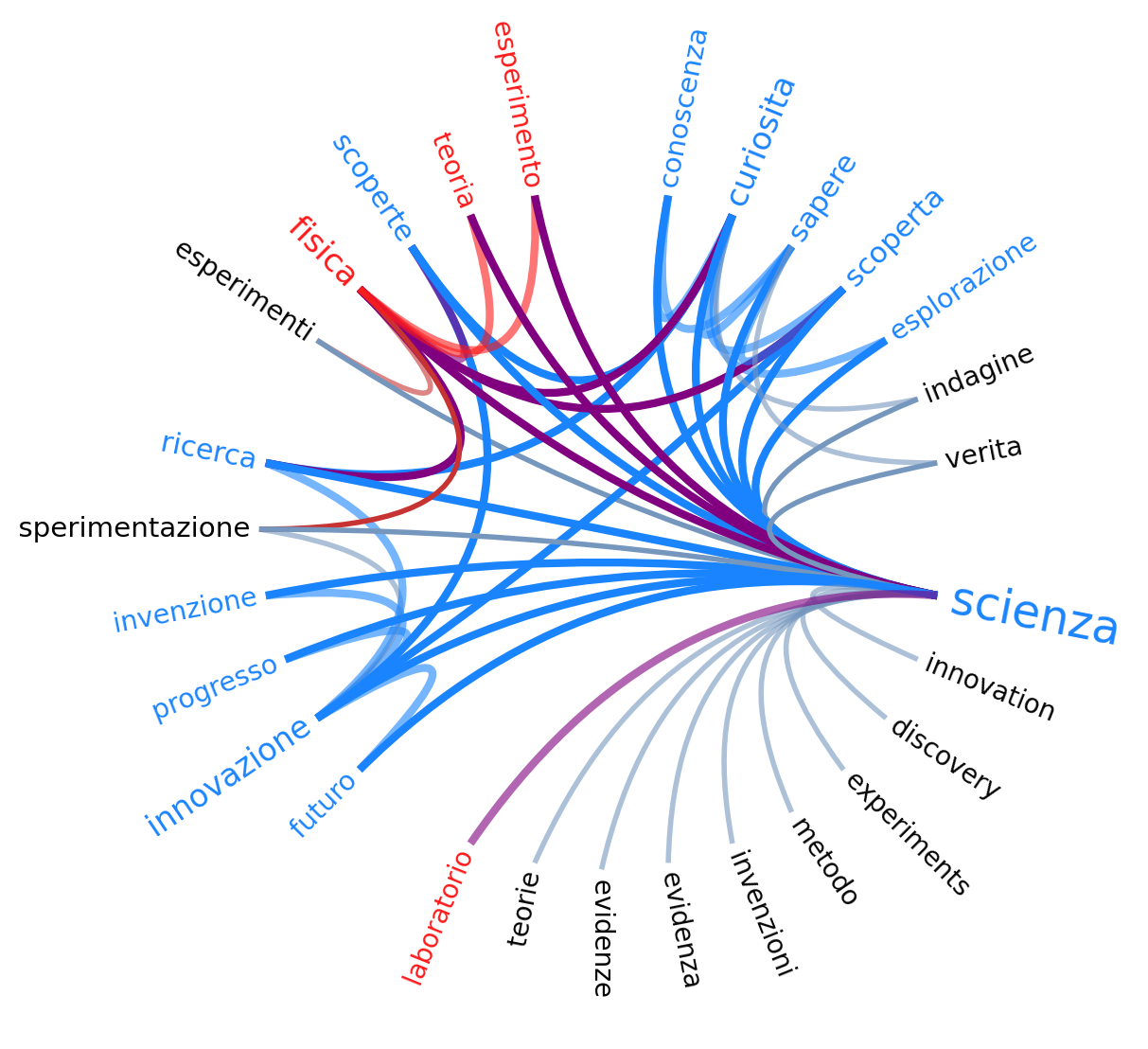}
    \caption{High anxiety GPT-4o students}
    \label{high_anx_gpt-4o_sci}
\end{subfigure}
\begin{subfigure}[p]{0.43\textwidth}
    \includegraphics[width=\textwidth]{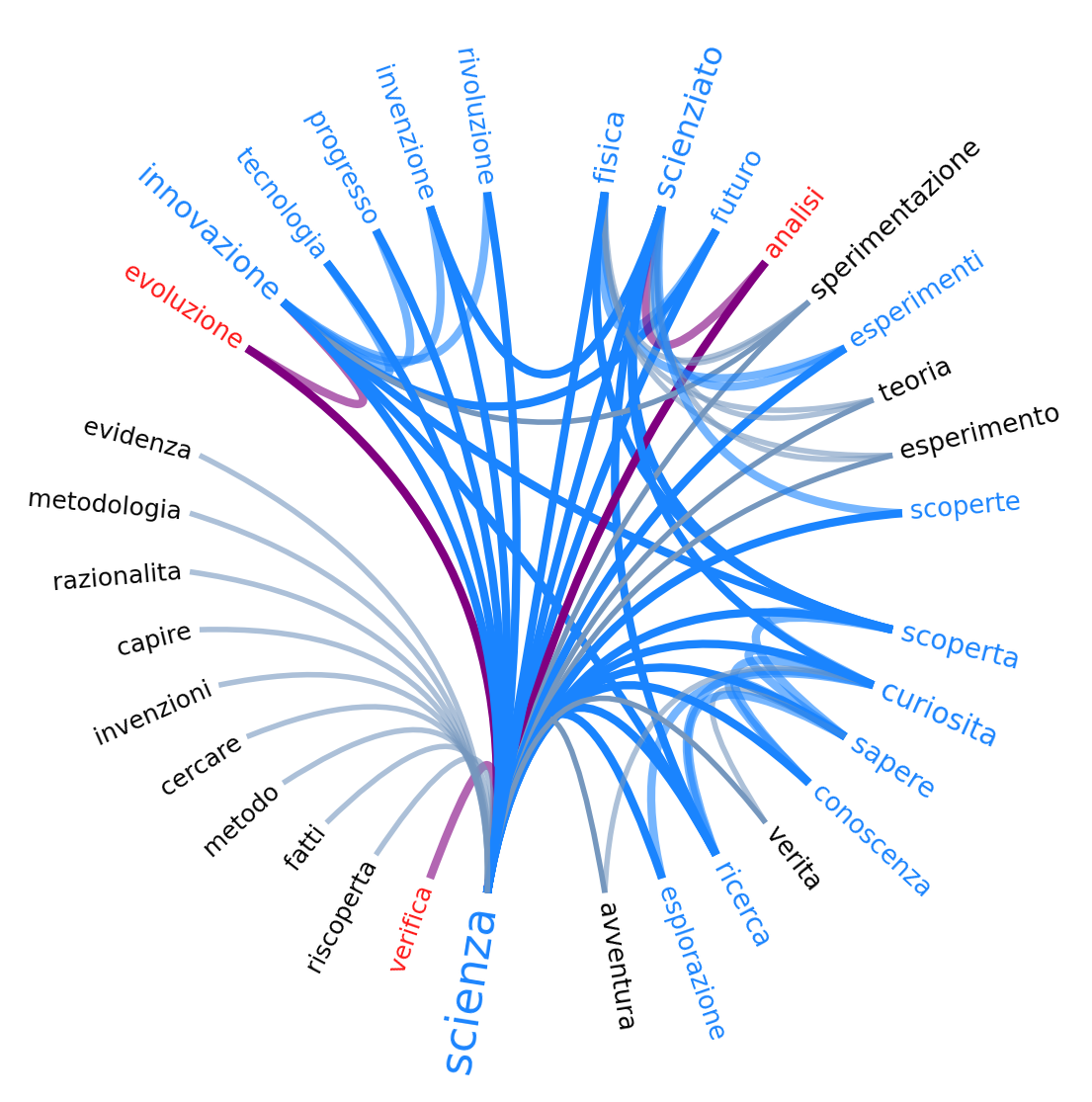}
    \caption{Low anxiety GPT-4o students}
    \label{low_anx_gpt-4o_sci}
\end{subfigure}
    \caption{Semantic frames for the node "science".}
    \label{fig:frame_science}
\end{figure}
Looking at the semantic frames of "science" (Figure \ref{fig:frame_science}), we observe that all GPT-simulated students, independently of their math anxiety levels, framed "science" as a positive concept and associated it mostly with other positively perceived ideas. Positive associations included words such as "research" and "knowledge", as well as "progress", "future", and "innovation", reflecting not only the inherently positive view of science but also its perceived potential and forward-looking impact. 

Importantly, GPT-3.5 low-anxiety students (\ref{low_anx_GPT3.5_sci}) managed to come up with the positive association between "science" and "math", which was absent in simulated GPT-3.5 high-anxiety students (\ref{high_anx_GPT3.5_sci}) and GPT-4o students (\ref{low_anx_gpt-4o_sci}; \ref{high_anx_gpt-4o_sci}). This association was also present in real-world students, but deformed: both high- and low-anxiety students managed to associate "math" with "science"; however, its valence was either contrastive or neutral. In fact, in the high-anxiety group (\ref{high_anx_students_math}), "science" was framed as a neutral idea, featuring also stereotypical associations (e.g. "science" and "hard"). 

On the other hand, low-anxiety students (\ref{low_anx_students_sci}) framed "science" in a more emotionally polarised way, showcasing a neutral, a positive and a negative cluster of associates. Here, positive associates were mostly related to discovery and innovation, while neutral associates were mostly linked to contextual ideas about science and its actors (e.g. "experiment" or "scientist"). Finally, negative associations referred to STEM subjects (e.g. "math", "physics" and "chemistry") and with negative jargon about exams and tests. These associations mirrored those present in GPT-3.5 low-anxiety students (e.g. "physics"; \ref{low_anx_GPT3.5_sci}) and both GPT-4o low- (e.g. "test"; \ref{low_anx_gpt-4o_sci}) and high- (e.g. "physics"; \ref{high_anx_gpt-4o_sci}) anxiety subgroups. Interestingly, these associations replicated results from a past study by \citet{abramski2023cognitive}, considering both humans and GPT students.

As for the local cluster coefficients, slightly different results were found compared to "math" and "anxiety". In the "science" semantic frame, GPT-4o and humans have almost identical clustering coefficients ($C = .0712$ and $C = .0714$, respectively), while GPT-3.5 has noticeably lower clustering ($ C = .0443$), indicating a less cohesive semantic frame for "science".

\paragraph{Statistics}
Since "statistics" was a recurring word in the semantic frames of both "math" and "science" for many subgroups of students, we decided to explore the semantic frame of this STEM-related concept.

\begin{figure}[!htbp]
\centering
\begin{subfigure}[p]{0.48\textwidth}
    \includegraphics[width=\textwidth]{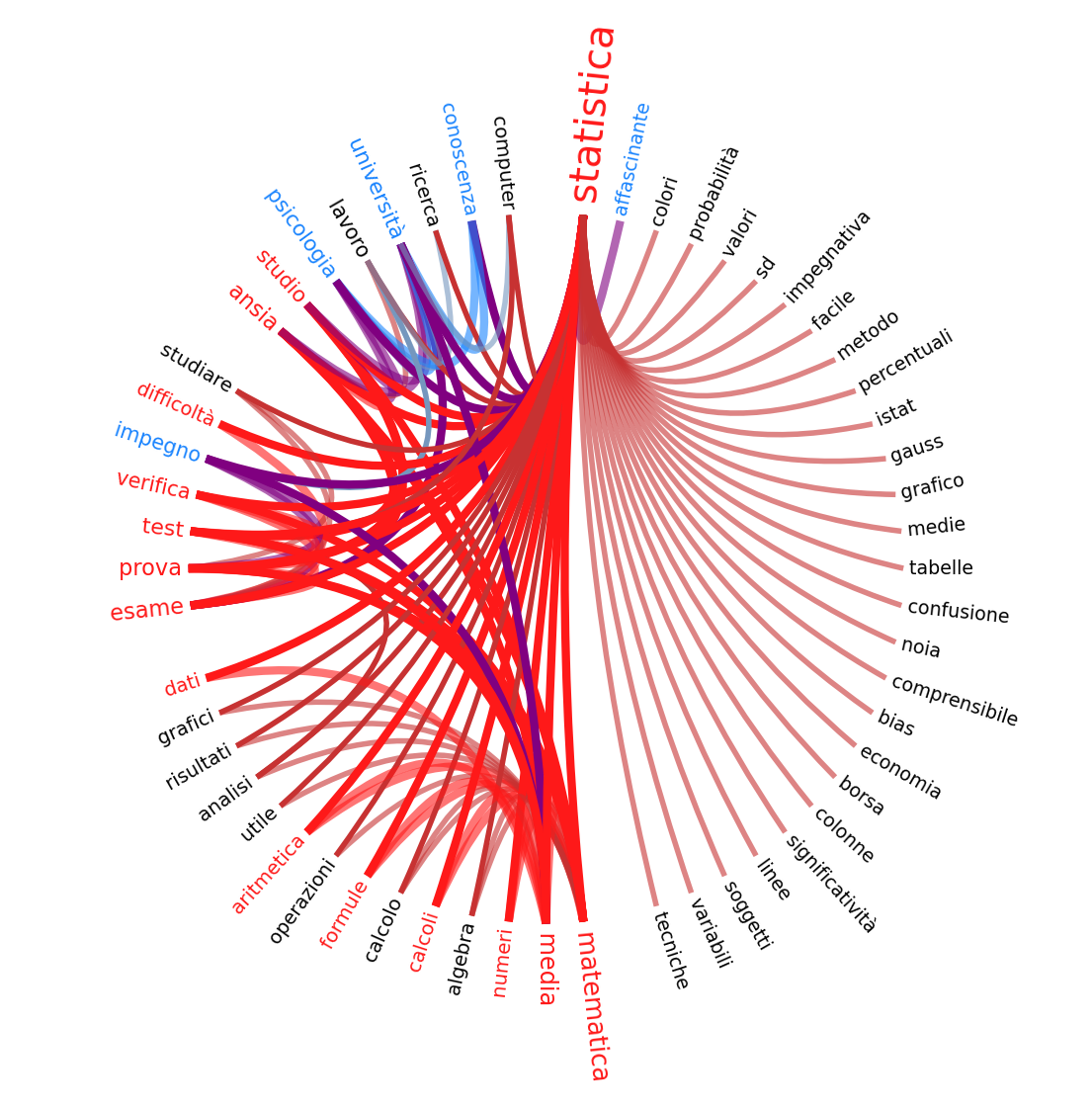}
    \caption{High anxiety students}
    \label{high_anx_students_stats}
\end{subfigure}
\begin{subfigure}[p]{0.48\textwidth}
    \includegraphics[width=\textwidth]{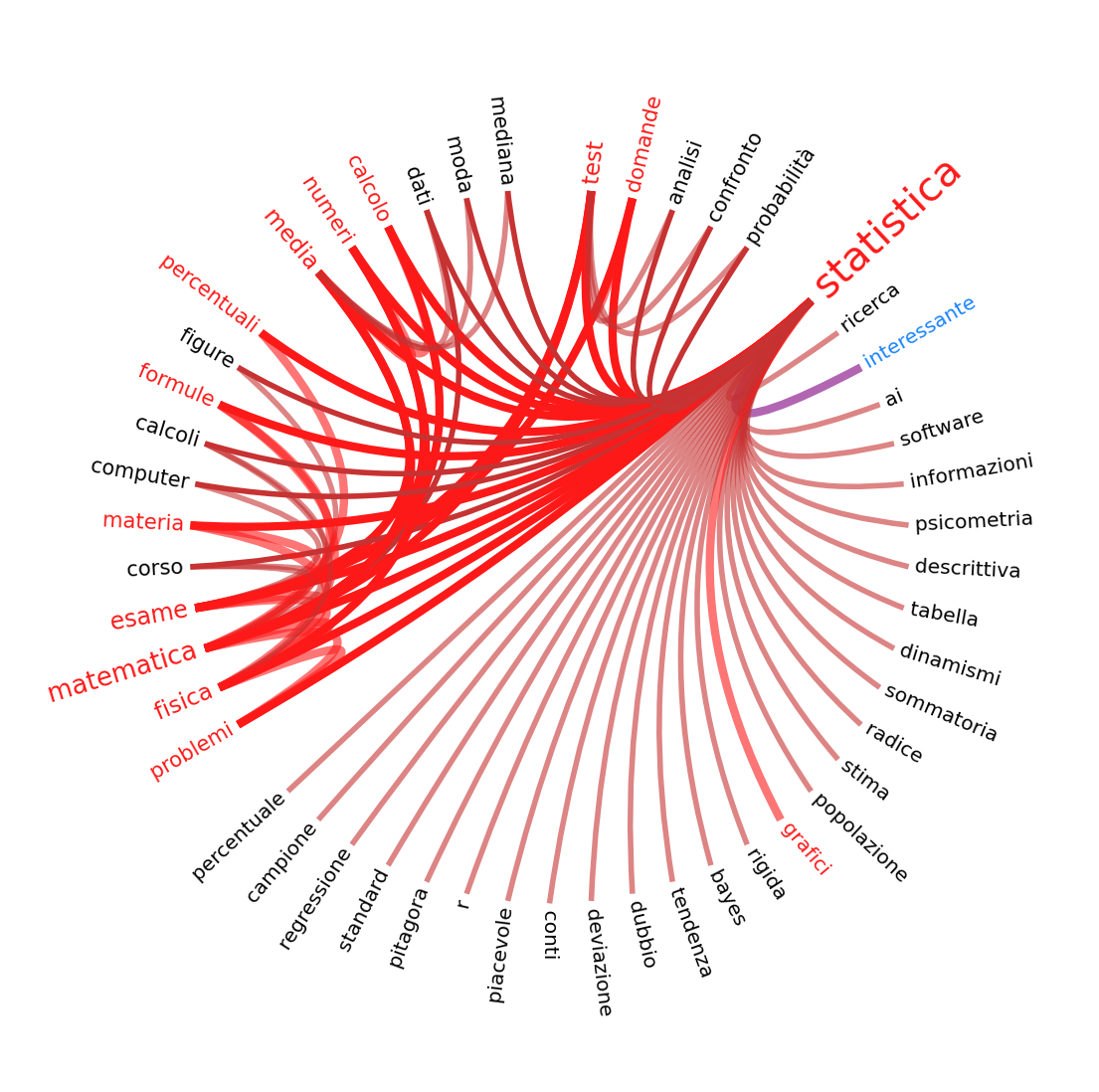}
    \caption{Low anxiety students}
    \label{low_anx_students_stats}
\end{subfigure}
\begin{subfigure}[p]{0.46\textwidth}
    \includegraphics[width=\textwidth]{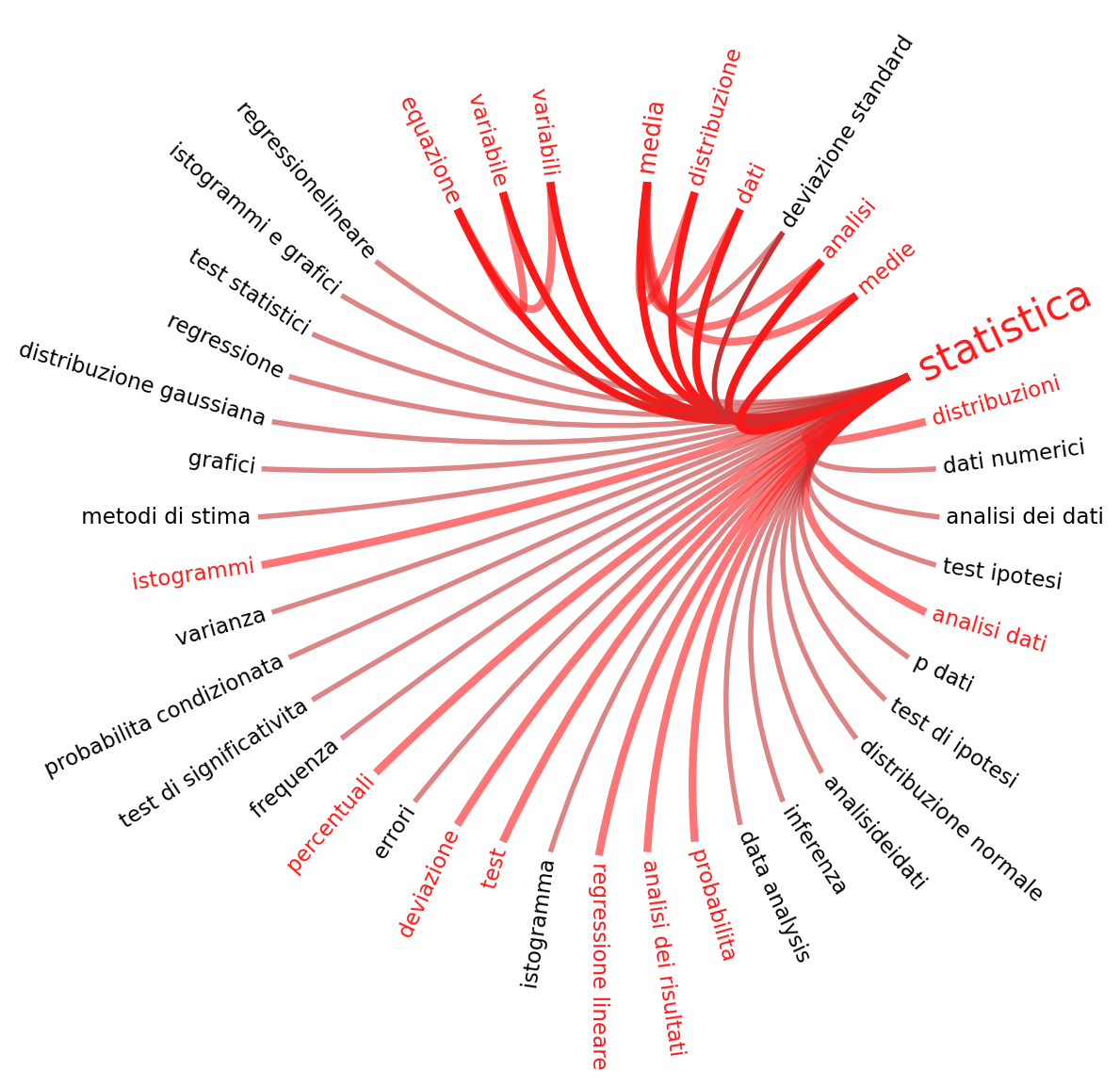}
    \caption{High anxiety GPT-3.5 students}
    \label{high_anx_GPT3.5_stats}
\end{subfigure}
\begin{subfigure}[p]{0.46\textwidth}
    \includegraphics[width=\textwidth]{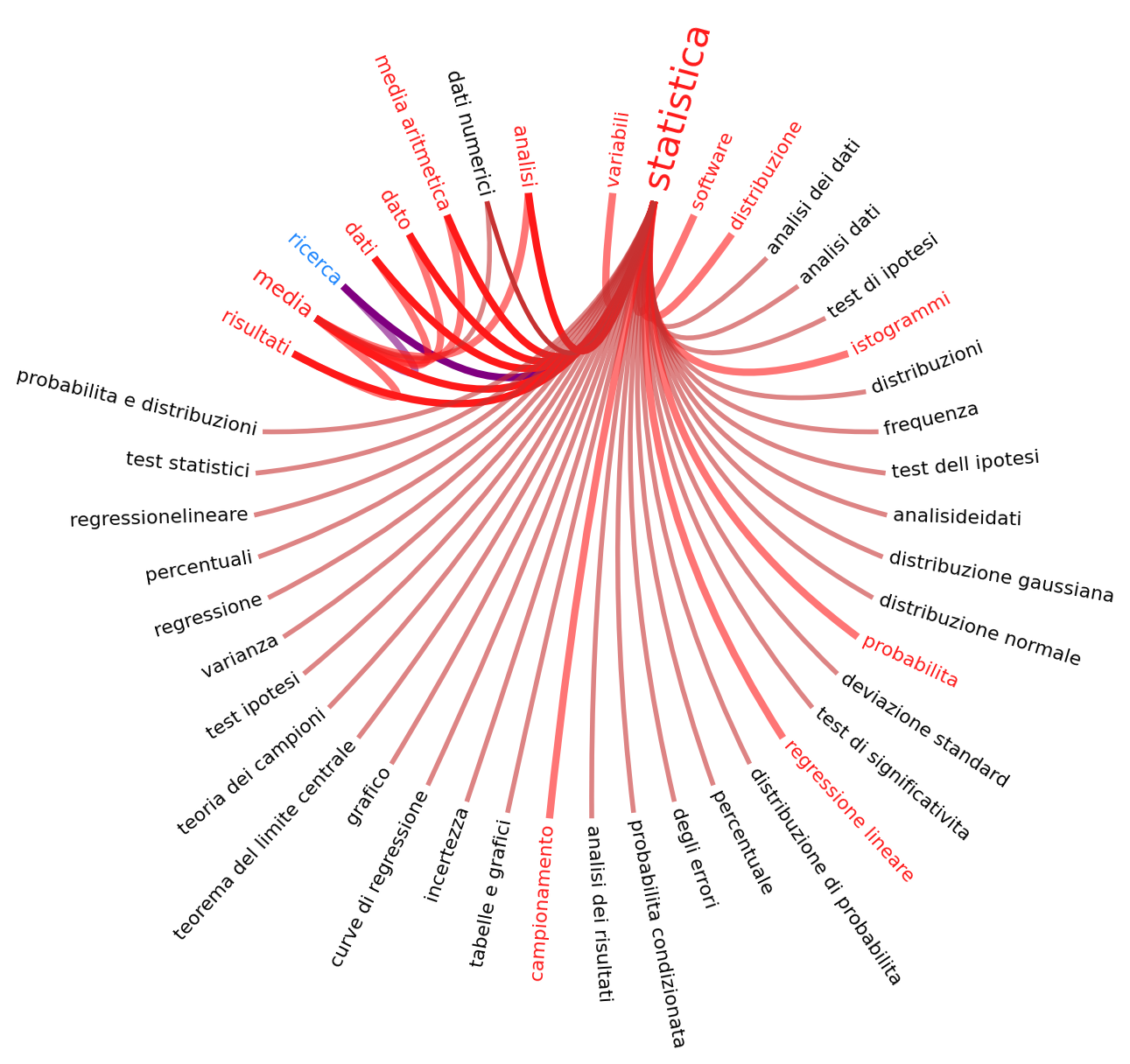}
    \caption{Low anxiety GPT-3.5 students}
    \label{low_anx_GPT3.5_stats}
\end{subfigure}
\begin{subfigure}[p]{0.43\textwidth}
    \includegraphics[width=\textwidth]{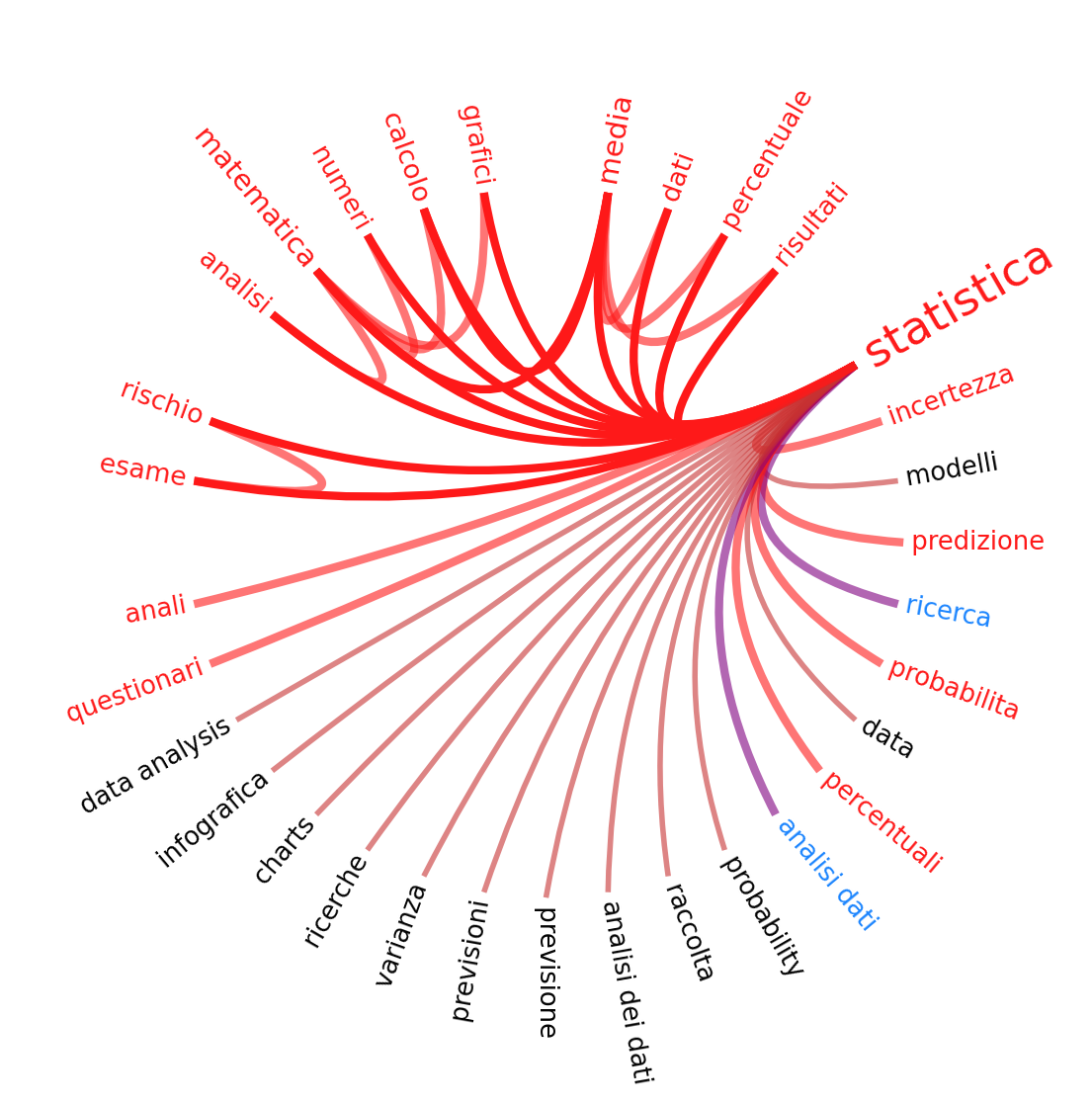}
    \caption{High anxiety GPT-4o students}
    \label{high_anx_gpt-4o_stats}
\end{subfigure}
\begin{subfigure}[p]{0.43\textwidth}
    \includegraphics[width=\textwidth]{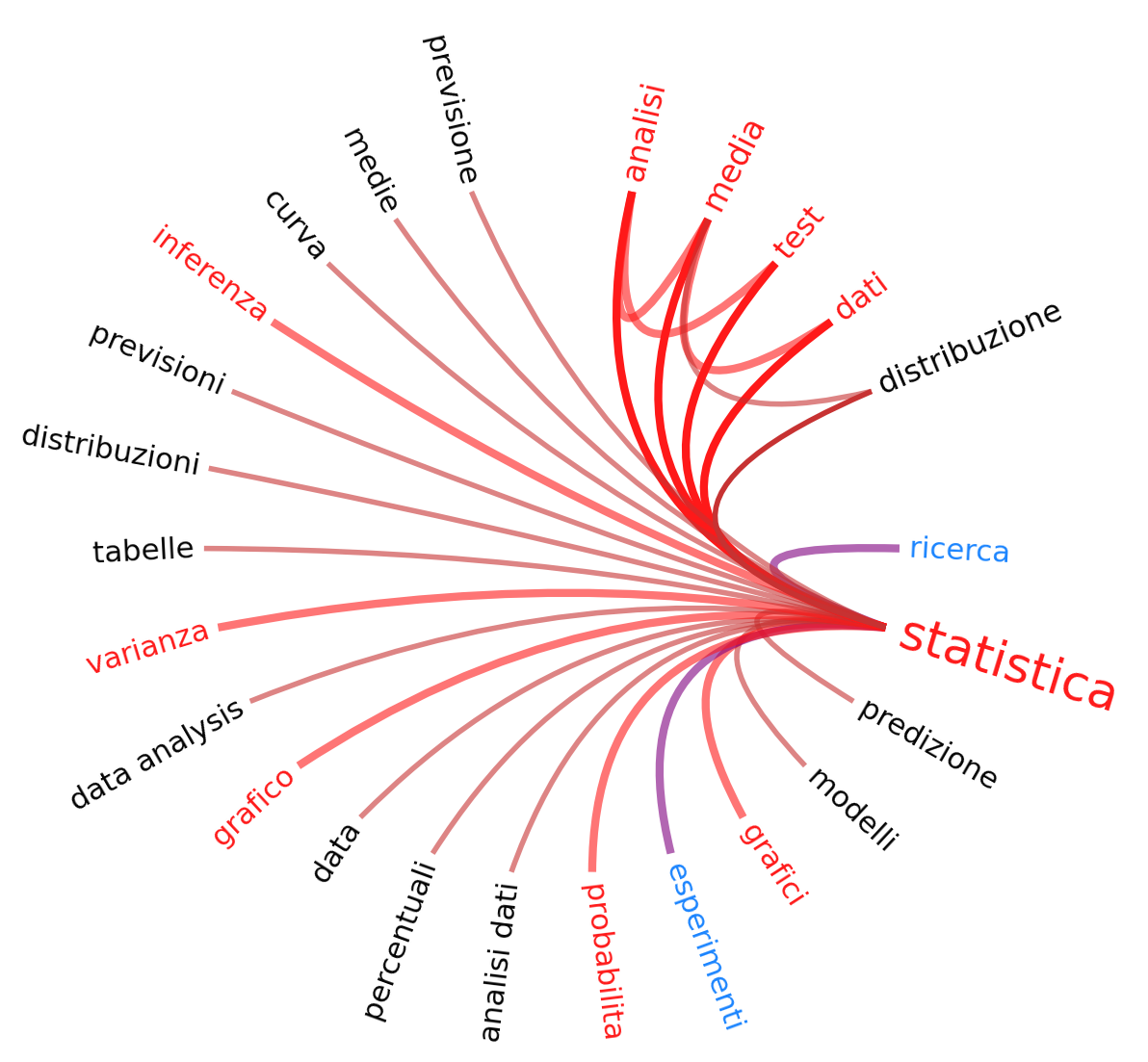}
    \caption{Low anxiety GPT-4o students}
    \label{low_anx_gpt-4o_stats}
\end{subfigure}
    \caption{Semantic frames for the node "statistics".}
    \label{fig:frame_statistics}
\end{figure}

Observing Figure \ref{fig:frame_statistics}, we can grasp that, independently of math anxiety levels, all simulated and real-world groups of students perceived "statistics" as a predominantly negative concept. Across both human and GPT-based data, "statistics" was surrounded by negatively or neutrally valenced associations such as "formulas", "percentage", "probability" and "variables". These associations highlight the technical, procedural, and often abstract nature of the concept, suggesting that it is primarily understood in terms of its cognitive demands rather than its practical or intellectual value.

Overall, these findings indicate an overwhelmingly negative or emotionally neutral framing of "statistics", consistent with prior evidence of widespread statistics anxiety among students \citep{siew2019anxiety}. This pattern underscores how both human and model-based associative networks reflect a shared cognitive bias towards viewing statistical concepts as effortful and anxiety-inducing, rather than as empowering or intellectually rewarding.

As with the previously analysed concept, also in the "statistics" semantic frame, its closeness centrality of both high-anxiety GPT-3.5 ($C = .0090$) and GPT-4o ($C = .0215$) groups was poorer than in the human counterpart ($C = .0418$). 

\paragraph{Therapist}
Figure \ref{fig:frame_science} importantly illustrates that the adopted methodology does not contain only negative semantic frames, but it can adapt and identify a variety of positive, negative and even polarised affective perceptions. This is even more evident when considering the overwhelmingly negative semantic frames of "statistics" (Figure \ref{fig:frame_statistics}) compared to positive semantic frames.

\begin{figure}[!htbp]
\centering
\begin{subfigure}[b]{0.44\textwidth}
    \includegraphics[width=\textwidth]{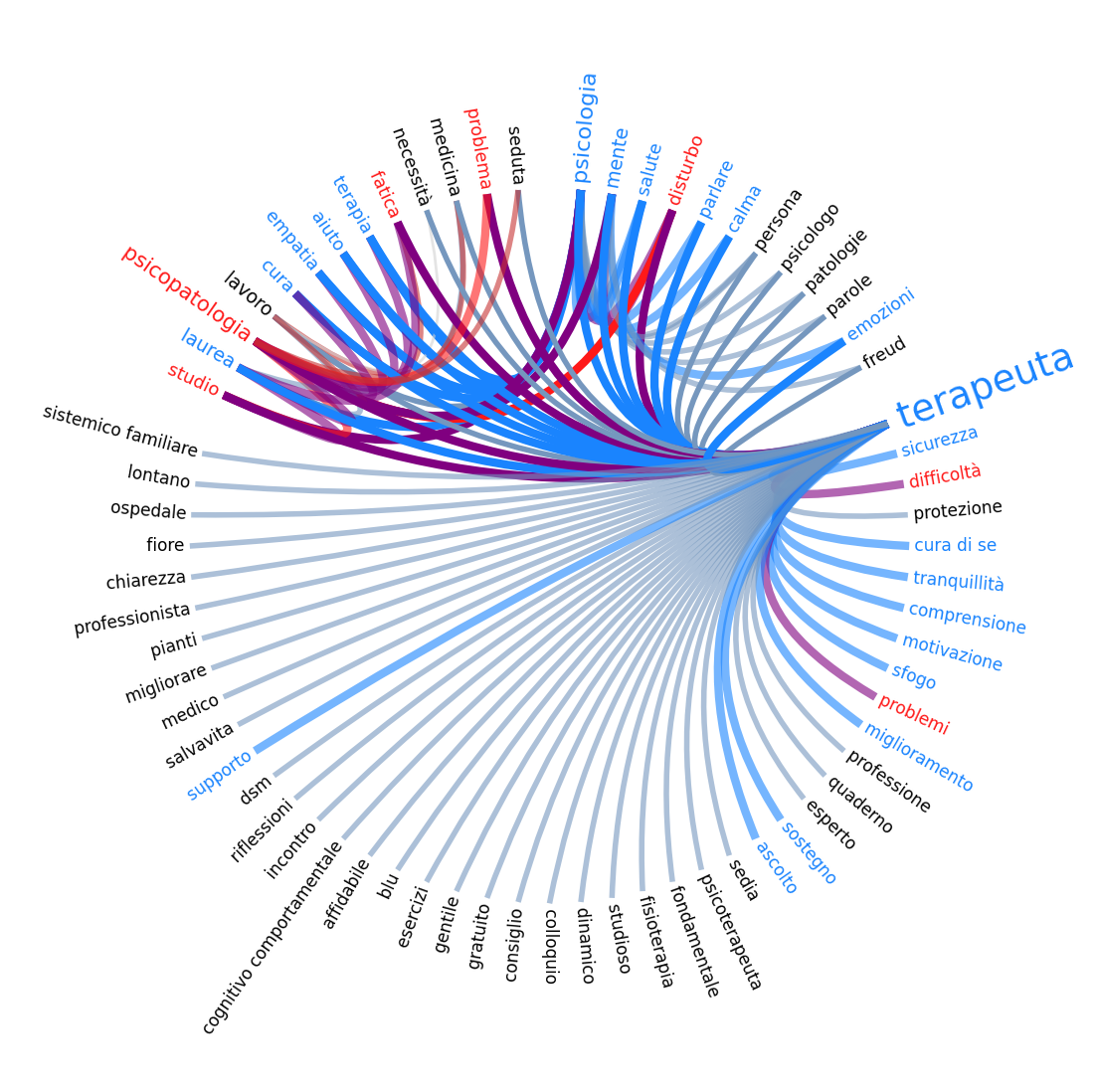}
    \caption{High anxiety students}
    \label{high_anx_students_ther}
\end{subfigure}
\begin{subfigure}[b]{0.44\textwidth}
    \includegraphics[width=\textwidth]{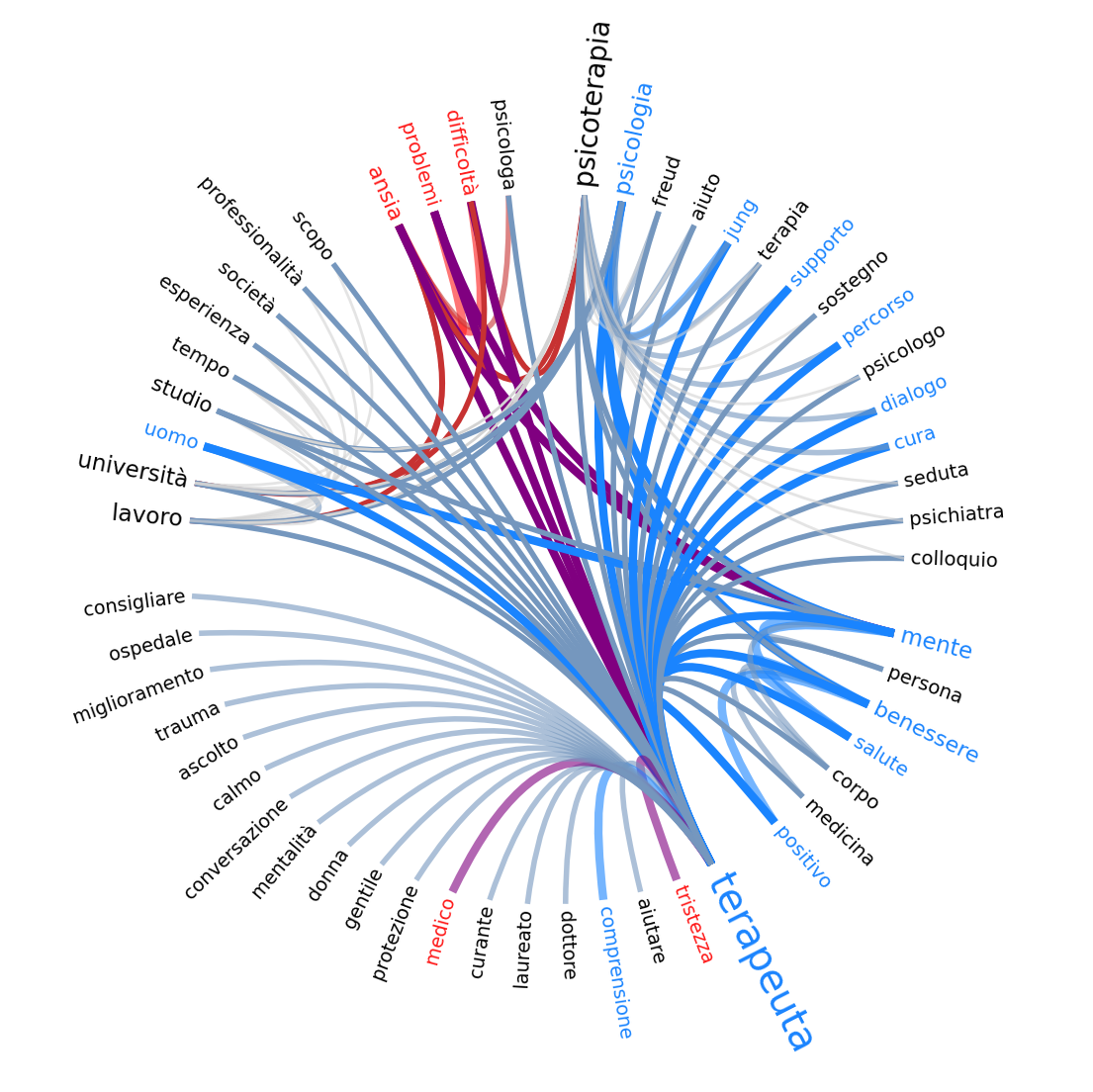}
    \caption{Low anxiety students}
    \label{low_anx_students_ther}
\end{subfigure}
\begin{subfigure}[b]{0.41\textwidth}
    \includegraphics[width=\textwidth]{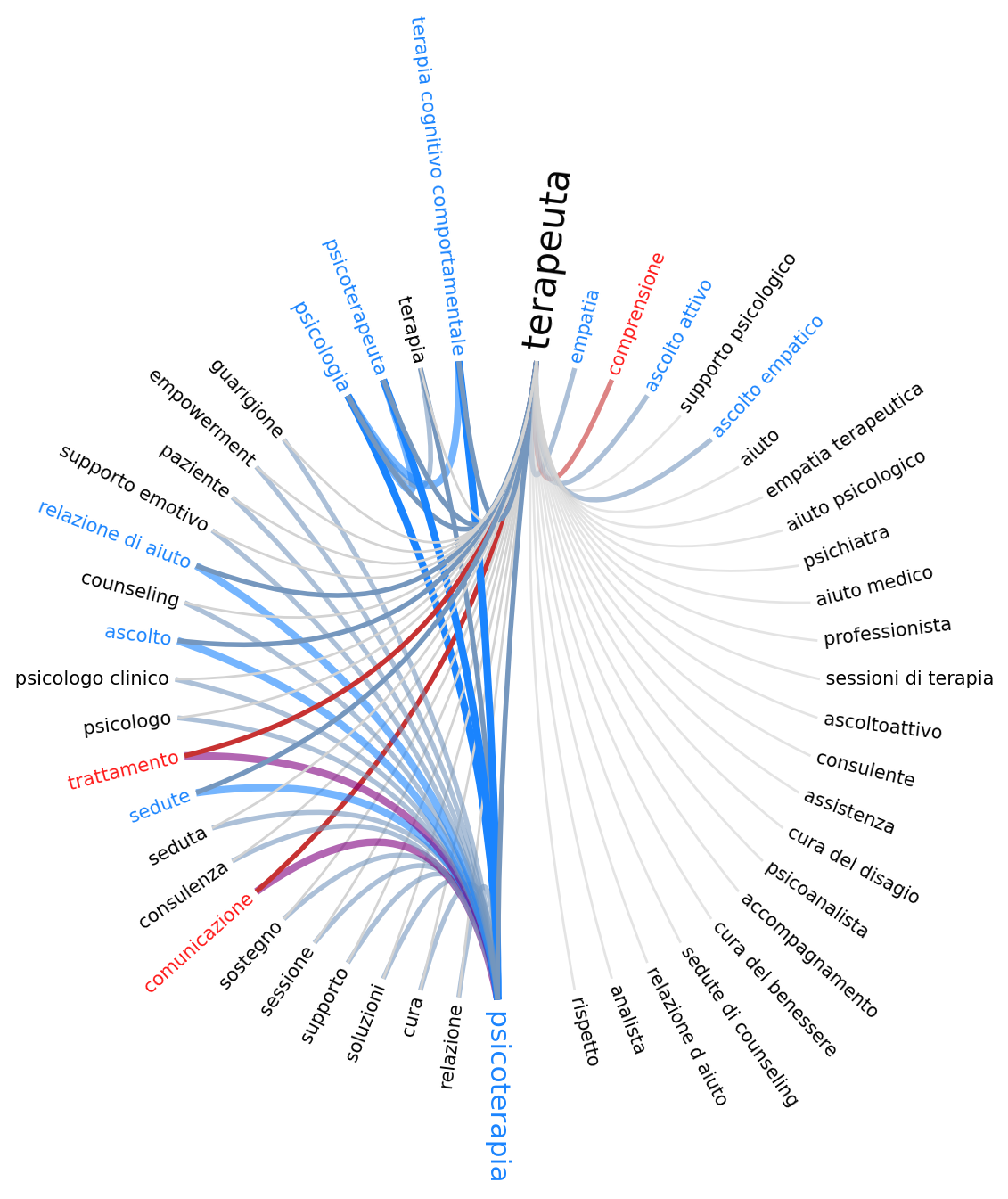}
    \caption{High anxiety GPT-3.5 students}
    \label{high_anx_GPT3.5_ther}
\end{subfigure}
\begin{subfigure}[b]{0.41\textwidth}
    \includegraphics[width=\textwidth]{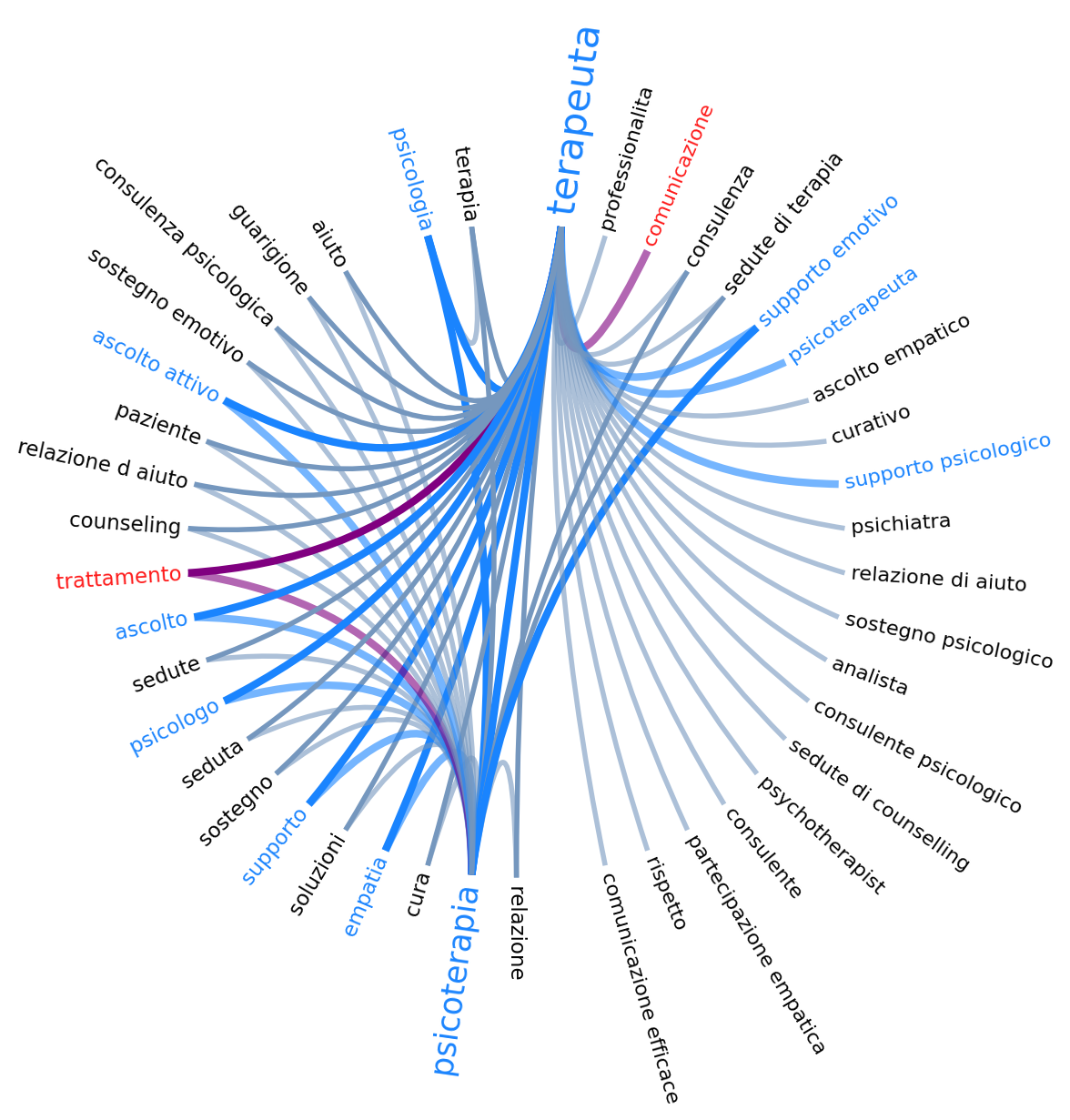}
    \caption{Low anxiety GPT-3.5 students}
    \label{low_anx_GPT3.5_ther}
\end{subfigure}
\begin{subfigure}[b]{0.4\textwidth}
    \includegraphics[width=\textwidth]{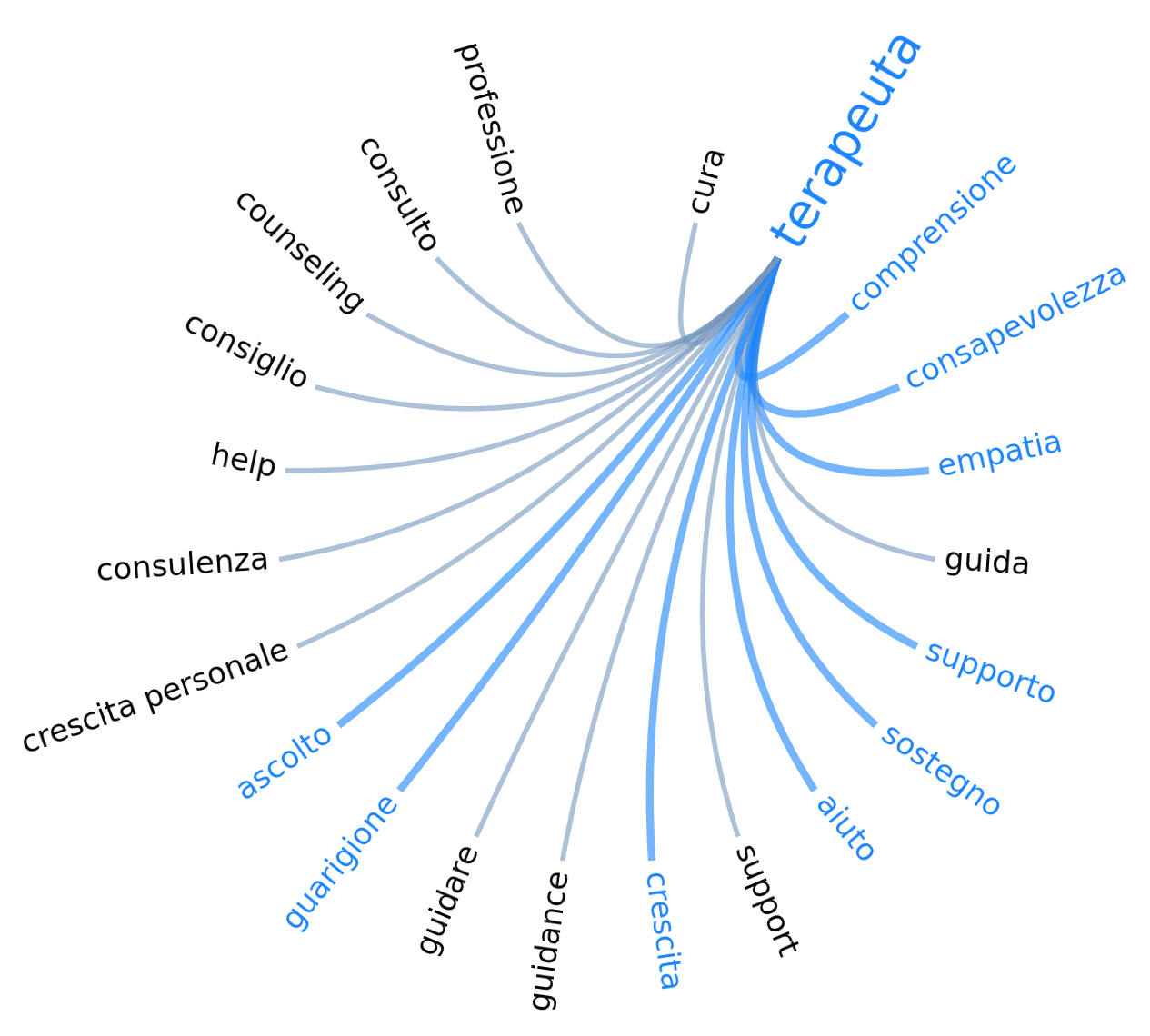}
    \caption{High anxiety GPT-4o students}
    \label{high_anx_gpt-4o_ther}
\end{subfigure}
\begin{subfigure}[b]{0.4\textwidth}
    \includegraphics[width=\textwidth]{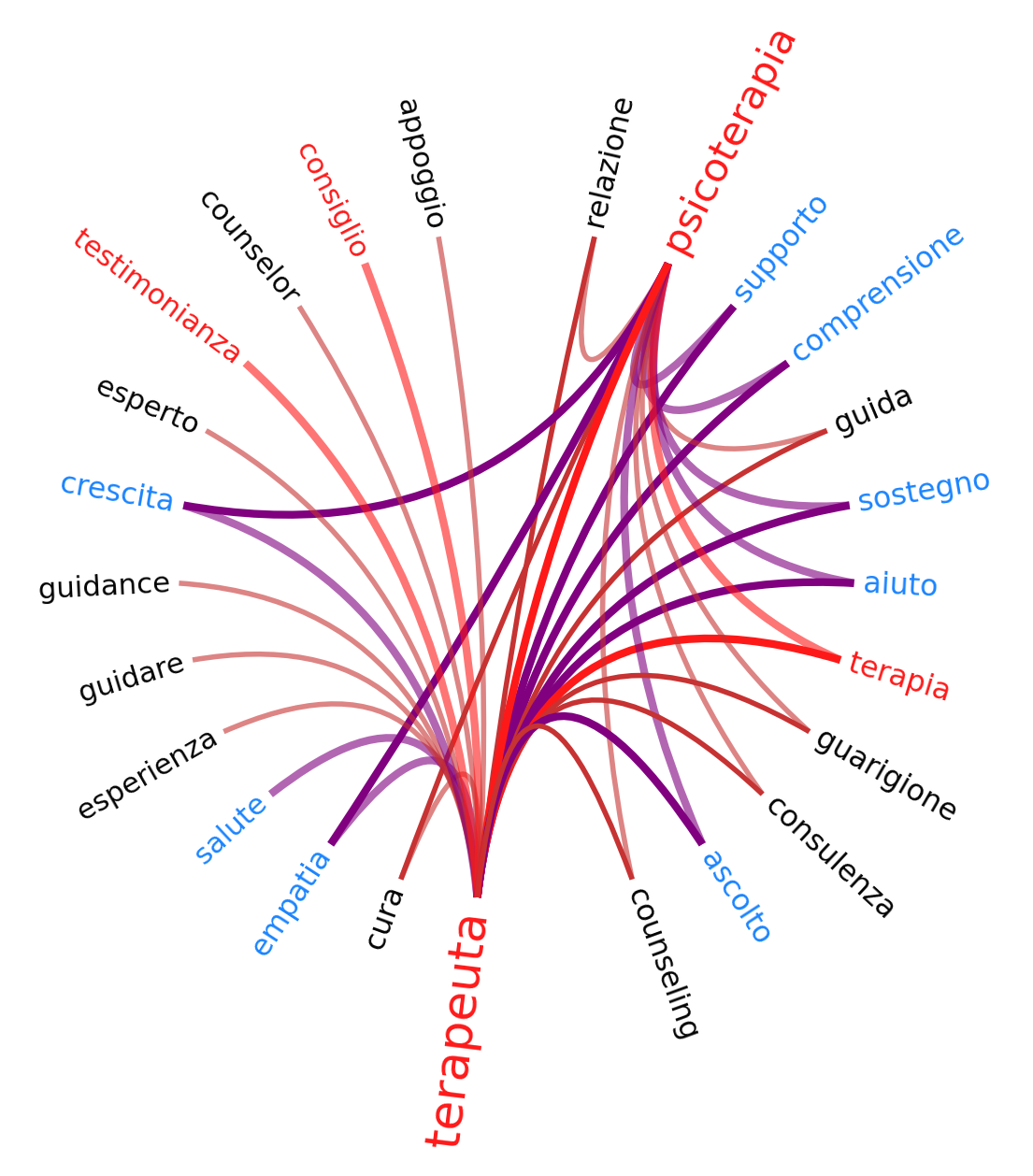}
    \caption{Low anxiety GPT-4o students}
    \label{low_anx_gpt-4o_ther}
\end{subfigure}
    \caption{Semantic frames for "therapist"}
    \label{fig:frame_therapist}
\end{figure}

To explore a concept likely to be perceived positively, we examined the semantic frame of "therapist", a figure often considered central in addressing math anxiety. As shown in Figure \ref{fig:frame_therapist}, "therapist" was generally framed as a positive concept, surrounded by mostly positive or neutral associations. However, two noteworthy deviations from this pattern emerged. Among high-anxiety GPT-3.5 simulated students, "therapist" was framed in a largely neutral way (\ref{high_anx_GPT3.5_ther}); while low-anxiety GPT-4o simulated students perceived "therapist" as a negative concept and associated it with negative or ambivalent terms (\ref{low_anx_gpt-4o_ther}), suggesting a subtle shift in connotation. One possible interpretation is that simulated students with low math anxiety—who would experience little perceived need for psychological support—may construe "therapist" as less relevant or even unnecessary, reflecting a mild negative bias in this context.

Taken together, these patterns indicate that while "therapist" is generally represented as a positive and emotionally supportive figure, GPT-based simulated populations can exhibit affective bias in framing this concept, which might be linked to their math anxiety levels. Nevertheless, these findings are noteworthy and warrant further investigation in future research.

\FloatBarrier

\subsection{Discussion}
This experiment investigated how math anxiety and related concepts are represented within the semantic mindsets of both human participants and simulated students generated by GPT-3.5 and GPT-4o. Our aims were (i) to clarify why the regression models in \nameref{Experiment 3} poorly fitted GPT data, and (ii) to understand structural and affective differences between human and LLM-based semantic networks.

Analyses of key concepts linked to math anxiety ("math", "anxiety", "science", "statistics", "therapist") revealed several divergences between humans and simulated students. Although GPT-3.5 and GPT-4o generally reproduced the same emotional polarity as humans, notable exceptions emerged. Both GPT-3.5 subgroups and low-anxiety GPT-4o students framed "math" positively, whereas humans perceived it as consistently negative. Similarly, while GPT-generated data depicted "science" as uniformly positive, human students included neutral and negative associations, particularly involving "math" and other STEM subjects. This contrast reflects a form of cognitive dissonance \citep{Festinger1957}, suggesting that math anxiety scores alone do not fully capture negative attitudes toward mathematics. Other cognitive-emotional mechanisms—such as emotional reasoning \citep{arntz1995if, Gangemi2012} or safety-seeking behaviour \citep{vandenhout2014behavior}—may underlie these perceptions.

The persistently negative framing of "statistics" across all groups pointed to widespread statistics anxiety \citep{siew2019anxiety}, particularly among psychology students. Both humans and GPTs emphasised the technical and abstract nature of statistics, highlighting a shared bias towards its cognitive difficulty rather than its practical value. In contrast, the concept of "anxiety" was consistently negative across all groups, even though humans displayed richer, more contextually grounded associations (e.g. "exam", "fear"), whilst GPTs’ associations remained broader and symptom-based (e.g. "tension", "nervousness").

Interestingly, "therapist" provided a counterpoint to these negative frames. It was largely positive across groups, except for low-anxiety GPT-4o students, who perceived it more negatively—perhaps reflecting a reduced perceived need for psychological support. This affective bias aligns with previous findings on LLM-generated affective distortions \citep{abramski2023cognitive}.

Another important difference emerged in the structure of their semantic frames: the semantic frames of GPT's students exhibited lower complexity and lower local clustering compared to the humans' ones, particularly for "math" and "anxiety". These simpler associative patterns indicate that GPTs construct more linear and less integrated conceptual networks, consistent with their reliance on co-occurrence patterns rather than experiential grounding.

Taken together, these findings suggest that GPT-3.5 and GPT-4o cannot yet reproduce the nuanced, context-sensitive structure of human semantic cognition. While capable of capturing general conceptual relationships, they lack the situational grounding and theory of mind \citep{binz2023using} that underpin human emotional reasoning. This likely explains the weaker regression model fits observed in \nameref{Experiment 3}.

Overall, \nameref{Experiment 4} highlights both the promise and current limitations of LLMs as cognitive simulators. Although they can approximate surface-level semantics, they still fail to mirror the affective richness and contextual depth of human conceptual organisation. Bridging this gap remains a key challenge for future research in both cognitive modelling and educational psychology.

\section{Discussion}
This study combines cognitive networks and psychometric scores related to math anxiety, as measured within two independent samples of undergraduate students and in their simulated counterparts (i.e. simulated personifications obtained with GPT-3.5 and GPT-4o). We use cognitive networks, in the form of behavioural forma mentis networks \citep{stella2019viability, stella2020forma, stella2021mapping}, to reconstruct individuals' mindsets around STEM and math anxiety \citep{stella2022network, franchino2025network}, i.e. how individuals perceive and link concepts with "math" and "anxiety" in their associative knowledge. We show that the structure and emotional valence of these conceptual representations are predictive of different dimensions of math anxiety in undergraduate psychology students, but not in their simulated counterparts generated by GPT models.

Different from textual forma mentis networks, which encode syntactic relationships between words in texts \citep{semeraro2024emoatlas}, this work uses behavioural forma mentis networks (BFMNs), i.e. networks of memory recall patterns enriched with valence data \citep{abramski2023cognitive}. Hence, this behavioural approach offers a fine-grained, interpretable access to individuals' semantic and autobiographic memories, linking together concepts - via knowledge available from the mental lexicon \citep{siew2019network} - and pleasantness ratings - coming from personal experience or expectations \citep{stella2020forma}.

Our quantitative findings are mainly two: (i) math anxiety is mirrored in the structure of individual students' mindsets, i.e. behavioural forma mentis networks capturing cognitive associations and affective ratings (Experiments 1, 2 and 4); (ii) this anxiety–structure correspondence is absent in state-of-the-art language models (GPT-4o, GPT-3.5), despite their fluent human-like output (Experiments 3 and 4).

Our first finding extends classic accounts in which math anxiety depletes working memory and biases attention \citep{ashcraft2007working}. The correspondence between cognitive network structure and psychometric levels of math anxiety indicates that anxiety is not only a consequence detectable in academic math performance; rather, math anxiety is encoded in semantic and affective structures relative to human memory. Higher anxiety co-occurs with a more negative framing of "math" and a more structurally prominent "anxiety". 

In cognitive network science, the number of direct connections a node has (its degree centrality) indicates how easily that concept can be reached or triggered within the network \citep{siew2019network}. Thus, a higher degree of "anxiety" means that anxious thoughts are more readily accessible and more likely to influence other ideas, such as "math". These results suggest that math-anxious students tend to form denser and more tightly linked associations around anxiety-related ideas. In network terms, the concept "anxiety" becomes a highly connected hub, indicating that it is frequently activated together with many other thoughts. Such structural prominence may reflect repetitive activation of anxiety-related concepts, a phenomenon known as rumination \citep{Dowker2016}. 

In simpler terms, our quantitative results indicate that behavioural forma mentis networks act as a map of how emotion shapes cognition: students with higher math anxiety show associative structures where "anxiety" dominates the map and "math" carries a more negative emotional tone. This further contributes to past scientific evidence showing how emotions can reorganise semantic connections in memory \citep{kenett2019semantic, stella2020forma, barrett2007experience}, producing measurable network patterns that align with psychometric indicators of anxiety. \citet{Dowker2016} emphasised the role of individual differences in math anxiety, highlighting how varying cognitive and emotional factors can influence anxiety levels. Our observation of less cohesive networks (around target concepts) in high-anxiety students aligns with these findings, suggesting that cognitive network structures bear crucial individual differences worth exploring. Indeed, our study reproduces the mindset (i.e. the ways of perceiving and associating ideas based on memory recall) for each individual of the dataset in \nameref{Experiment 1} and \nameref{Experiment 2}. If individual differences might be encoded as cognitive networks, how would one interpret and investigate their source, as in our case? Cognitive theories of anxiety, such as those discussed by \citet{Eysenck2007}, state that anxiety disorders are characterised by attentional and interpretive biases. Our study suggests that these biases might be reflected in the structural/affective properties of forma mentis networks, where math-anxious students attribute more negative valence to math concepts, while giving more importance (i.e. degree centrality) to anxiety itself. 

This structural–affective mirroring opens the way for practical interventions. Psychological interventions targeting bias mechanisms (e.g. behaviour-as-information; emotional reasoning) can help students identify and interrupt maladaptive appraisal loops before they consolidate in memory \citep{Gangemi2021, vandenhout2014behavior}. Integrating emotion-regulation skills with associative retraining may foster more flexible, less negatively valenced math frames \citep{Gross2002}. Because BFMNs quantify target representations, they also provide actionable outcome measures for intervention trials (e.g. shifts toward more diverse and less negatively valenced associations to "math") at the individual level. \citet{Dowker2016} emphasised the role of individual differences in math anxiety, highlighting how varying cognitive and emotional factors can influence anxiety levels. Our observation of less cohesive networks (around target concepts) in high-anxiety students aligns with these findings, suggesting that cognitive network structures bear crucial individual differences worth exploring.

Critically, the same correspondence between BFMN structure and math anxiety does not emerge in GPT-based simulations, including extensive details for personification. Both GPT-3.5 and GPT-4o produce coherent associations and plausible self-reports, yet LLMs' network associations fail to predict their own prompted "anxiety" levels. This dissociation indicates that current LLMs lack the affectively grounded associative organisation observed in humans. We find that Large Language Models, even those as complex and advanced as GPT-4o, can fluently describe anxiety, yet they do not show the structural footprints of anxiety within their generated cognitive networks. This gap is theoretically informative: our regression analyses and semantic frames all highlight affective grounding and embodied experience \citep{barrett2007experience} as necessary conditions for linking emotional states to the structure of associative knowledge \citep{kenett2019semantic, stella2024cognitive}. Our findings indicate that, without real emotional or physiological feedback loops, current large language models might lack the mechanisms through which emotions reorganise semantic networks in humans, delineating a fundamental boundary in using LLM personae as models of affective cognition within educational contexts \citep{Gross2002}.

\subsection{Consequences for Education and AI}
The absence of an anxiety/structure correspondence in GPT's cognitive networks has tangible educational consequences. It limits the fidelity with which language models can represent emotional influences on learning \citep{abramski2023cognitive}, constrains their capacity for adaptive, empathy-based feedback \citep{de2025introducing}, and restricts their utility as predictive tools for identifying students at risk of math anxiety \citep{franchino2025network, stella2022network}.

LLMs such as GPT-4o and GPT-3.5 lack affective grounding, i.e. the emotional learning history that shapes how humans learn and link concepts \citep{vilhunen2022clarifying}. GPT systems might also inherit human and non-human negative biases towards mathematics, further distorting their perceptions of STEM subjects \citep{abramski2023cognitive}. Hence, current LLMs cannot faithfully simulate how anxiety, motivation, or confidence influence learning behaviour, as also found in past relevant works \citep{de2025introducing}. In other words, educational AI systems may appear empathetic linguistically but remain "emotionally flat", limiting their ability to personalise instruction or provide effective anxiety-sensitive support \citep{turkkila2025combining}. 

In humans, this work found that the structure and valence of associative networks around "math" reflect real cognitive–emotional features that predict math anxiety and its facets. Because LLMs do not reproduce these structural signatures, they cannot yet serve as reliable diagnostic models for detecting early indicators of math anxiety or related affective learning barriers \citep{franchino2025network}. Consequently, relying on LLM-based "digital twins" to screen or model student anxiety could yield false negatives, obscuring genuine emotional distress. Given the promising results of LLMs in fostering creativity through question asking \citep{sasson2023mirror} and positive classroom experiences \citep{turkkila2025combining}, this calls for further research.

Cognitive network models of human learners integrate both knowledge structure and affect. LLM-simulated learners, however, encode only textual knowledge distributions. Without anxiety-linked restructuring of concept networks, these models cannot reproduce how negative emotions disrupt cognitive flexibility or working memory during learning \citep{ashcraft2007working, Eysenck2007}. 

The present findings also bear important implications for educational practice and teacher training. Given that current Large Language Models cannot genuinely perceive or model students’ anxiety, as found here, teachers remain irreplaceable agents in the personalisation of learning approaches. Teachers' sensitivity to students’ emotional states allows differentiated interventions that respond to individual needs and promote classroom practices of reflection and self-reflection on affective aspects of learning \citep{stella2022network}. Moreover, since the participants in this study were psychology students, our results highlight the importance of preparing future psychologists to critically understand both the potential and the limitations of AI tools. Developing methodological literacy and awareness of how AI can simulate but not genuinely reproduce human affective processes is essential for their future professional practice \citep{grion2024students}. Training programs should emphasise formative assessment and targeted feedback as strategies to detect and support students showing early signs of math anxiety, thereby preventing the consolidation of avoidance behaviours and negative self-perceptions toward mathematics.

The above four points all converge towards the same point: educational technologies that aim to simulate or support human learners must therefore integrate affective grounding—through associative networks or knowledge graphs \citep{stella2024cognitive}, while reinforcing the emotional context of concepts. Novel generations of hybrid cognitive/affective models could more accurately mirror the intertwined cognitive and emotional dynamics of real learning, as captured by recent complex systems approaches \citep{vilhunen2022clarifying}. Pursuing this direction can potentially fill a current gap in the available LLMs' technology.

\subsection{Limitations and Future Directions}
While this study demonstrates that math anxiety is embedded in the structure and affective valence of individual semantic networks, several limitations must be acknowledged.

Firstly, sample size and domain specificity constrain generalisation. The human data were collected from psychology undergraduates and may not reflect how math anxiety manifests in students from other STEM-oriented fields. With access to more resources, broader sampling, and longitudinal tracking can assess the stability of BFMN–anxiety relationships across learning contexts \citep{stella2020forma}. Secondly, our LLM simulations highlight conceptual rather than emotional modelling. GPT-based "students" lack genuine affective grounding or embodied experience \citep{barrett2007experience}. Their associative patterns depend entirely on linguistic co-occurrence, not lived emotion–cognition coupling. Consequently, comparisons between human and LLM-based BFMNs should be interpreted as contrasts between emotionally grounded and purely symbolic systems, highlighting LLMs' design boundaries. 

Finally, the cross-sectional design limits causal inference. Although network structure correlates with anxiety levels, it remains unclear whether anxiety reorganises associative structure or whether pre-existing associative biases increase vulnerability to anxiety. Longitudinal and intervention-based BFMN studies can disentangle these causal pathways as an exciting direction for future research.

\section{Conclusion}
In conclusion, our study provides novel insights into the cognitive underpinnings of math anxiety. By clarifying the relationships between associative knowledge structure, conceptual perceptions and math anxiety, we pave the way for new approaches to intervention and support. Integrating cognitive and emotional strategies could be beneficial in mitigating the adverse effects of math anxiety, ultimately enhancing educational outcomes and student well-being.

\newpage
\section*{Results Tables}
\begin{table}[!htbp]
\centering
\begin{tabular}{>{\raggedright\arraybackslash}p{3cm} >{\raggedright\arraybackslash}p{6cm} c c c}
\toprule
\textbf{Group} & \textbf{Variable} & \textbf{N} & \textbf{M} & \textbf{SD} \\
\midrule
\multirow{4}{=}{\textbf{Experiment 1 Students}} 
& "Anxiety" Degree & 70.00 & 6.29 & 1.91 \\
& "Anxiety" Closeness Centrality & 70.00 & 0.08 & 0.06 \\
& "Math" Degree & 70.00 & 6.53 & 1.21 \\
& "Math" Closeness Centrality & 70.00 & 0.09 & 0.05 \\
\midrule
\multirow{4}{=}{\textbf{Experiment 2 Students}} 
& "Anxiety" Degree & 57.00 & 7.19 & 2.89 \\
& "Anxiety" Closeness Centrality & 57.00 & 0.09 & 0.06 \\
& "Math" Degree & 57.00 & 7.53 & 2.05 \\
& "Math" Closeness Centrality & 57.00 & 0.09 & 0.05 \\
\midrule
\multirow{4}{=}{\textbf{GPT-3.5 Students}} 
& "Anxiety" Degree & 300.00 & 5.53 & 0.59 \\
& "Anxiety" Closeness Centrality & 300.00 & 0.03 & 0.02 \\
& "Math" Degree & 300.00 & 5.16 & 0.38 \\
& "Math" Closeness Centrality & 300.00 & 0.03 & 0.01 \\
\midrule
\multirow{4}{=}{\textbf{GPT-4o Students}} 
& "Anxiety" Degree & 300.00 & 5.63 & 0.54 \\
& "Anxiety" Closeness Centrality & 300.00 & 0.04 & 0.01 \\
& "Math" Degree & 300.00 & 5.35 & 0.50 \\
& "Math" Closeness Centrality & 300.00 & 0.03 & 0.01 \\
\bottomrule
\end{tabular}
\caption{Descriptive statistics of study variables.}
\label{tab:descriptive_stats_network}
\end{table}

\begin{table}[!htbp]
\centering
\begin{tabular}{>{\raggedright\arraybackslash}p{3cm} >{\raggedright\arraybackslash}p{4.5cm} c c c c c c}
\toprule
\textbf{Sample} & \textbf{Anxiety Factor} & \textbf{N} & \textbf{M} & \textbf{SD} & \textbf{Q1} & \textbf{Q3} & \textbf{SKP} \\
\midrule
\multirow{4}{=}{\textbf{Experiment 1 Students}}
& Evaluation MA & 70.00 & 25.84 & 6.86 & 21.25 & 30.00 & 0.22 \\
& Everyday/Social MA & 70.00 & 14.54 & 5.91 & 10.00 & 18.00 & 1.54 \\
& Passive Observation MA & 70.00 & 8.46 & 2.59 & 6.00 & 10.75 & 0.87 \\
& Total MA & 70.00 & 48.84 & 13.19 & 39.25 & 56.75 & 0.84 \\
\midrule
\multirow{4}{=}{\textbf{Experiment 2 Students}}
& Evaluation MA & 57.00 & 25.53 & 7.60 & 20.00 & 31.00 & 0.11 \\
& Everyday/Social MA & 57.00 & 15.63 & 6.68 & 10.00 & 22.00 & 0.69 \\
& Passive Observation MA & 57.00 & 10.67 & 5.16 & 7.00 & 13.00 & 1.49 \\
& Total MA & 57.00 & 51.82 & 16.79 & 39.00 & 64.00 & 0.66 \\
\midrule
\multirow{4}{=}{\textbf{GPT-3.5 Students}}
& Evaluation MA & 300.00 & 35.08 & 2.69 & 33.00 & 37.00 & -0.03 \\
& Everyday/Social MA & 300.00 & 22.54 & 2.44 & 21.00 & 24.00 & 0.23 \\
& Passive Observation MA & 300.00 & 16.73 & 1.84 & 16.00 & 18.00 & 0.39 \\
& Total MA & 300.00 & 74.35 & 5.49 & 71.00 & 77.00 & 0.15 \\
\midrule
\multirow{4}{=}{\textbf{GPT-4o Students}}
& Evaluation MA & 300.00 & 33.10 & 3.86 & 31.00 & 35.00 & $-$0.58 \\
& Everyday/Social MA & 300.00 & 17.56 & 3.05 & 15.00 & 20.00 & $-$0.12 \\
& Passive Observation MA & 300.00 & 11.50 & 2.96 & 9.00 & 13.00 & 0.16 \\
& Total MA & 300.00 & 62.16 & 8.90 & 56.00 & 68.00 & $-$0.29 \\
\bottomrule
\end{tabular}
\caption{Descriptive statistics of math anxiety factors for each participant sample.}
\label{tab:descriptive_stats_MA}
\end{table}

\begin{table}[!htbp]
\centering
\begin{tabular}{>{\raggedright\arraybackslash}p{3cm} >{\raggedright\arraybackslash}p{4.3cm} c c c c}
\toprule
\textbf{Group} & \textbf{Predictors} & $\beta$ & $S.E.$ & $t$ & $p$ \\
\midrule
\multirow{4}{=}{\textbf{Experiment 1 Students}} 
& $R^2$ = .238 & & & & \\
& Intercept & .249 & .060 & 4.169 & $<$ .001*** \\
& "Anxiety" Degree & .542 & .130 & 4.158 & $<$ .001*** \\
& "Math" Degree & $-$0.177 & .127 & $-$1.389 & .170 \\
& "Anxiety" Valence & .104 & .124 & .834 & .407 \\
& "Math" Valence & $-$0.106 & .067 & $-$1.582 & .119 \\
\midrule
\multirow{4}{=}{\textbf{Experiment 2 Students}} 
& $R^2$ = .347 & & & & \\
& Intercept & .268 & .045 & 6.005 & $<$ .001*** \\
& "Anxiety" Degree & .642 & .127 & 5.065 & $<$ .001*** \\
& "Anxiety" Valence & $-$0.040 & .197 & $-$0.206 & .838 \\
& "Math" Valence & $-$0.052 & .067 & $-$0.767 & .446 \\
\midrule
\multirow{4}{=}{\textbf{GPT-3.5 Students}}  
& $R^2$ = .027 & & & & \\
& Intercept & .565 & .015 & 37.632 & $<$ .001*** \\
& "Anxiety" Degree & .012 & .026 & .454 & .651 \\
& "Math" Degree & $-$0.104 & .040 & $-$2.603 & .010** \\
& "Anxiety" Valence & .046 & .054 & .848 & .397 \\
& "Math" Valence & $-$0.008 & .016 & $-$0.485 & .628 \\
\midrule
\multirow{4}{=}{\textbf{GPT-4o Students}} 
& $R^2$ = .072 & & & \\
& Intercept & .526 & .020 & 26.239 & $<$ .001*** \\
& "Anxiety" Degree & .096 & .038 & 2.506 & .013* \\
& "Anxiety" Valence & $-$0.053 & .180 & -0.293 & .770 \\
& "Math" Valence & -0.099 & .024 & -4.189 & $<$ .001*** \\
\bottomrule
\end{tabular}
\caption{Results of Linear Regression Analysis for Total Math Anxiety.}
\label{tab:lin_reg_total_MA}
\end{table}

\begin{table}[!htbp]
\centering
\begin{tabular}{>{\raggedright\arraybackslash}p{3cm} >{\raggedright\arraybackslash}p{4.3cm} c c c c}
\toprule
\textbf{Group} & \textbf{Predictors} & $\beta$ & $S.E.$ & $t$ & $p$ \\
\midrule
\multirow{4}{=}{\textbf{Experiment 1 Students}}  
& $R^2$ = .199 & & & & \\
& Intercept & .362 & .066 & 5.515 & $<$ .001*** \\
& "Anxiety" Degree & .524 & .143 & 3.657 & $<$ .001*** \\
& "Math" Degree & $-$0.254 & .140 & $-$1.812 & .075 \\
& "Anxiety" Valence & $-$0.010 & .137 & $-$0.076 & .939 \\
& "Math" Valence & $-$0.136 & .074 & $-$1.840 & .070 \\
\midrule
\multirow{4}{=}{\textbf{Experiment 2 Students}}  
& $R^2$ = .405 & & & & \\
& Intercept & .399 & .042 & 9.470 & $<$ .001*** \\
& "Anxiety" Degree & .617 & .120 & 5.160 & $<$ .001*** \\
& "Anxiety" Valence & $-$0.032 & .185 & $-$0.174 & .863 \\
& "Math" Valence & $-$0.144 & .063 & $-$2.278 & .027* \\
\midrule
\multirow{4}{=}{\textbf{GPT-3.5 Students}}  
& $R^2$ = .029 & & & & \\
& Intercept & .608 & .018 & 33.470 & $<$ .001*** \\
& "Anxiety" Degree & $-$0.002 & .031 & $-$0.049 & .961 \\
& "Math" Degree & $-$0.142 & .048 & $-$2.932 & .004** \\
& "Anxiety" Valence & .024 & .066 & .370 & .712 \\
& "Math" Valence & $-$0.006 & .019 & $-$0.325 & .745 \\
\midrule
\multirow{4}{=}{\textbf{GPT-4o Students}} 
& $R^2$ = .054 & & & & \\
& Intercept & .591 & .020 & 29.446 & $<$ .001*** \\
& "Anxiety" Degree & .091 & .038 & 2.370 & .018* \\
& "Anxiety" Valence & $-$0.065 & .181 & $-$0.360 & .719 \\
& "Math" Valence & $-$0.082 & .024 & $-$3.465 & $<$ .001*** \\
\bottomrule
\end{tabular}
\caption{Results of Linear Regression Analysis for Evaluation MA Factor.}
\label{tab:lin_reg_eval_MA}
\end{table}

\begin{table}[!htbp]
\centering
\begin{tabular}{>{\raggedright\arraybackslash}p{3cm} >{\raggedright\arraybackslash}p{4.3cm} c c c c}
\toprule
\textbf{Group} & \textbf{Predictors} & $\beta$ & $S.E.$ & $t$ & $p$ \\
\midrule
\multirow{4}{=}{\textbf{Experiment 1 Students}}  
& $R^2$ = .192 & & & & \\
& Intercept & .072 & .041 & 1.772 & .081 \\
& "Anxiety" Degree & .455 & .122 & 3.738 & $<$ .001*** \\
& "Anxiety" Valence & .098 & .116 & .845 & .401 \\
& "Math" Valence & $-$0.037 & .063 & $-$0.588 & .559 \\
\midrule
\multirow{4}{=}{\textbf{Experiment 2 Students}}
& $R^2$ = .272 & & & & \\
& Intercept & .150 & .059 & 2.563 & .013* \\
& "Anxiety" Degree & .740 & .167 & 4.441 & $<$ .001*** \\
& "Anxiety" Valence & .005 & .258 & .018 & .985 \\
& "Math" Valence & .048 & .088 & .539 & .592 \\
\midrule
\multirow{4}{=}{\textbf{GPT-3.5 Students}} 
& $R^2$ = .010 & & & & \\
& Intercept & .473 & .017 & 27.616 & $<$ .001*** \\
& "Anxiety" Degree & .020 & .030 & .668 & .504 \\
& "Anxiety" Valence & .082 & .064 & 1.284 & .200 \\
& "Math" Valence & $-$0.013 & .019 & $-$0.678 & .498 \\
\midrule
\multirow{4}{=}{\textbf{GPT-4o Students}} 
& $R^2$ = .020 & & & & \\
& Intercept & .491 & .023 & 21.704 & $<$ .001*** \\
& "Anxiety" Degree & .096 & .043 & 2.209 & .028* \\
& "Anxiety" Valence & $-$0.072 & .204 & $-$0.353 & .725 \\
& "Math" Valence & $-$0.032 & .027 & $-$1.191 & .235 \\
\bottomrule
\end{tabular}
\caption{Results of Linear Regression Analysis for Everyday/Social MA Factor.}
\label{tab:lin_reg_social_MA}
\end{table}

\begin{table}[!htbp]
\centering
\begin{tabular}{>{\raggedright\arraybackslash}p{3cm} >{\raggedright\arraybackslash}p{4.3cm} c c c c}
\toprule
\textbf{Group} & \textbf{Predictors} & $\beta$ & $S.E.$ & $t$ & $p$ \\
\midrule
\multirow{4}{=}{\textbf{Experiment 1 Students}} 
& $R^2$ = .185 & & & & \\
& Intercept & .139 & .057 & 2.423 & .018* \\
& "Anxiety" Degree & .390 & .171 & 2.278 & .026* \\
& "Anxiety" Valence & .452 & .164 & 2.760 & .007** \\
& "Math" Valence & $-$0.137 & .089 & $-$1.533 & .130 \\
\midrule
\multirow{4}{=}{\textbf{Experiment 2 Students}} 
& $R^2$ = .131 & & & & \\
& Intercept & .122 & .052 & 2.362 & .022* \\
& "Anxiety" Degree & .403 & .147 & 2.742 & .008** \\
& "Anxiety" Valence & $-$0.089 & .228 & $-$0.390 & .698 \\
& "Math" Valence & $-$0.002 & .078 & $-$0.026 & .979 \\
\midrule
\multirow{4}{=}{\textbf{GPT-3.5 Students}} 
 & $R^2$ = .017 & & & & \\
& Intercept & .481 & .018 & 27.167 & $<$ .001*** \\
& "Anxiety" Degree & .017 & .030 & .554 & .580 \\
& "Math" Degree & $-$0.103 & .047 & $-$2.182 & .030* \\
& "Anxiety" Valence & .017 & .064 & .272 & .786 \\
& "Math" Valence & $-$0.000 & .019 & $-$0.019 & .985 \\
\midrule
\multirow{4}{=}{\textbf{GPT-4o Students}} 
& $R^2$ = .154 & & & & \\
& Intercept & .431 & .020 & 21.168 & $<$ .001*** \\
& "Anxiety" Degree & .085 & .039 & 2.166 & .031* \\
& "Anxiety" Valence & $-$0.006 & .183 & $-$0.034 & .973 \\
& "Math" Valence & $-$0.171 & .024 & $-$7.091 & $<$ .001*** \\
\bottomrule
\end{tabular}
\caption{Results of Linear Regression Analysis for Passive Observation MA Factor.}
\label{tab:lin_reg_obs_MA}
\end{table}

\FloatBarrier

\section*{Appendix}
\renewcommand{\thetable}{A.\arabic{table}}
\setcounter{table}{0}

\begin{table}[!htbp]
\centering
\begin{tabular}{>{\raggedright\arraybackslash}p{3.5cm} >{\raggedright\arraybackslash}p{3.5cm} >{\raggedright\arraybackslash}p{3.5cm} >{\raggedright\arraybackslash}p{3.5cm}}
\toprule
\textbf{STEM and Academic Disciplines} & \textbf{Psychology and Mental Health} & \textbf{Academic Evaluation} & \textbf{Personal Attitude} \\
\midrule
Physics        & Relationship     & Exam        & Fun \\
Mathematics    & Psychopathology  & Anxiety     & Creativity \\
Statistics     & Emotion          & Test        & Passion \\
Cognition      & Psychotherapist  & Assessment  & Work \\
Computer science & Mind           & Degree      & Model \\
Psychology     & Behaviour         & Average     & Knowledge \\
Biology        & Personality      & Trial       & Innovation \\
Equation       & Psychopathology  & Professor   & Adventure \\
Science        & Attachment       & Teacher     & Curiosity \\
Neuroscience   & Wellbeing        & University  & Challenge \\
\bottomrule
\end{tabular}
\caption{Cue words selected as stimuli in Experiment 1 Associations Task (translated from Italian to English).}
\label{tab:cue_words_exp_1}
\end{table}

\begin{table}[!htbp]
    \centering
    \begin{tabular}{>{\raggedright\arraybackslash}p{3.5cm} >{\raggedright\arraybackslash}p{3.5cm} >{\raggedright\arraybackslash}p{3.5cm} >{\raggedright\arraybackslash}p{3.5cm}}
\toprule
\textbf{STEM Disciplines} & \textbf{Academic Actors and Places} & \textbf{Learning and Motivation} & \textbf{Math Anxiety triggers} \\
\midrule
Technology               & Student         & To study                 & Change of plans \\
Mathematics              & Parent          & To plan                  & Confusion \\
Engineering              & Teacher         & To set a goal           & Problem \\
Biology                  & Class           & To memorize              & Anxiety \\
Physics                  & Exam            & Concept map              & Distraction \\
Statistics               & Notes           & To review                & Math test \\
Numbers                  & Blackboard      & Importance               & Errors \\
Calculation              & Assignment      & Commitment               & Failure \\
Algorithm                & Program         & Motivation               & Usefulness of math \\
Equation                 & Math Performance  & Concentration   & \\
Mathematician            &                 & Effort                   & \\
Solution                 &                 & To keep track           & \\
Proof                    &                 &                          & \\
\bottomrule
\end{tabular}
\caption{Cue words selected as stimuli in Experiment 2 Associations and Valence Task (translated from Italian to English).}
\label{tab:cue_words_exp_2}
\end{table}

\newcolumntype{C}[1]{>{\centering\arraybackslash}m{#1}}

\begin{table}[!htbp]
\renewcommand{\arraystretch}{1.3} 
  \centering
  \footnotesize
  \begin{tabular}{C{1cm} m{7cm} m{7cm}}
    \toprule 
    \textbf{Factor} & \textbf{MAS-UK item} & \textbf{MAS-IT item} \\ 
    \midrule
    \multirow{9}{*}{\rotatebox{90}{\parbox{8cm}{\centering Everyday/Social MA}}}
        & Having someone watch you multiply 12 x 23 on paper & Avere qualcuno che mi guarda moltiplicare 12 x 23 su carta \\ 
        & Being asked to write an answer on the board at the front of a maths class & Se mi viene chiesto di scrivere una risposta alla lavagna all'inizio di una lezione di matematica \\ 
        & Taking a maths exam & Sostenere un esame di matematica \\ 
        & Being asked to calculate £9.36 divided by four in front of several people & Se mi viene chiesto davanti a molte altre persone di calcolare EUR 9,36 diviso per 4\\ 
        & Calculating a series of multiplication problems on paper & Calcolare una serie di moltiplicazioni su carta\\ 
        & Being given a surprise maths test in a class  & Dover affrontare un test di matematica a sorpresa in una classe\\ 
        & Being asked to memorize a multiplication table & Dover memorizzare una tabellina \\ 
        & Being asked to calculate three fifths as a percentage & Se mi viene chiesto di calcolare i 3/5 di una percentuale \\ 
        & Being asked a maths question by a teacher in front of a class & Se mi viene chiesta una domanda di matematica da un/una insegnante di fronte alla classe \\ 
    \midrule
    \multirow{8}{*}{\rotatebox{90}{\parbox{8cm}{\centering Everyday/Social MA}}}
        & Adding up a pile of change & Calcolare la somma degli spiccioli di un resto\\ 
        & Being asked to add up the number of people in a room & Se mi viene chiesto di sommare il numero di persone in una stanza \\ 
        & Calculating how many days until a person’s birthday & Calcolare quanti giorni mancano al compleanno di una persona \\ 
        & Being given a telephone number and having to remember it & Ricevere un numero di telefono e doverlo ricordare \\ 
        & Working out how much time you have left before you set off to work or place of study & Calcolare quanto tempo mi rimane prima di partire per il lavoro o il luogo di studio \\ 
        & Working out how much change a cashier should have given you in a shop after buying several items & Calcolare quanto resto dovrebbe avermi dato un cassiere in un negozio dopo aver acquistato diversi articoli \\ 
        & Deciding how much each person should give you after you buy an object that you are all sharing the cost of & Decidere quanto ogni persona dovrebbe darmi dopo aver acquistato un oggetto di cui condividete il costo \\ 
        & Working out how much your shopping bill comes to & Calcolare quanto sia il conto di uno scontrino \\ 
    \midrule
    \multirow{6}{*}{\rotatebox{90}{\parbox{4.2cm}{\centering Passive Observation MA}}}
        & Reading the word "algebra" & Leggere la parola "algebra"\\ 
        & Listening to someone talk about maths & Ascoltare qualcuno che parla di matematica \\ 
        & Reading a maths textbook & Leggere un testo di matematica\\ 
        & Watching someone work out an algebra problem & Guardare qualcuno risolvere un problema di algebra\\ 
        & Sitting in a maths class & Frequentare una lezione di matematica \\ 
        & Watching a teacher/lecturer write equations on the board & Guardare un/una insegnante scrivere equazioni alla lavagna\\ 
    \bottomrule
  \end{tabular}
  \caption{MAS-UK items and corresponding Italian translated MAS-IT items \citep{franchino2025network}.}
  \label{tab:mas_it_uk_items}
\end{table}

\clearpage
\bibliographystyle{elsarticle-harv}
\bibliography{bibliography.bib}

\end{document}